\documentclass[letterpaper, 10pt, conference]{ieeeconf}
\IEEEoverridecommandlockouts
\usepackage[utf8]{inputenc}
\usepackage[T1]{fontenc}
\usepackage[hidelinks]{hyperref}
\usepackage{url}
\usepackage{booktabs}
\usepackage{amsfonts}
\usepackage{nicefrac}
\usepackage{microtype}
\usepackage{graphicx}
\usepackage{multirow}
\let\labelindent\relax
\usepackage{enumitem}
\usepackage{xcolor}
\usepackage{amsmath}
\usepackage{subcaption}
\usepackage{tabularx}
\usepackage{makecell}
\usepackage{float}
\usepackage{longtable}
\usepackage{colortbl}

\graphicspath{{media/}}

\definecolor{catgen}{HTML}{1A5276}
\definecolor{catif}{HTML}{196F3D}
\definecolor{catld}{HTML}{B9770E}
\definecolor{catlh}{HTML}{C0392B}
\definecolor{catmem}{HTML}{8E44AD}

\def\keywordname{{\bfseries\emph Keywords}}%
\def\keywords#1{\par\addvspace\medskipamount{\rightskip=0pt plus1cm
\def\and{\ifhmode\unskip\nobreak\fi\ $\cdot$
}\noindent\keywordname\enspace\ignorespaces#1\par}}

\newcommand{\methodname}{BiDex-100}

\makeatletter
\def\@IEEEdynamictitlevspace{\vspace{6pt}}%
\makeatother

\title{\LARGE\bf BiDex-100: A Comprehensive Benchmark for Bimanual Dexterous Manipulation Across 100+ Simulation Tasks}

\author{Yong~Wang$^{1,\ast}$, Kecheng~Zheng$^{2}$, Xing Zhu$^{2}$, Chao~Xu$^{1}$, Yujun~Shen$^{2}$, and Shengze~Cai$^{1,\ast}$
\thanks{$^{\ast}$ indicates the corresponding authors. Emails: \{\tt yongw@zju.edu.cn, shengze\_cai@zju.edu.cn\}}\\
{\normalsize $^{1}$Zhejiang University \qquad $^{2}$Ant Group}%
}

\begin{document}

\IEEEaftertitletext{%
  \begingroup\centering
    \textcolor{blue}{\url{https://robbyant-research.github.io/BiDex-100/}}\\[\baselineskip]%
    \includegraphics[width=\textwidth]{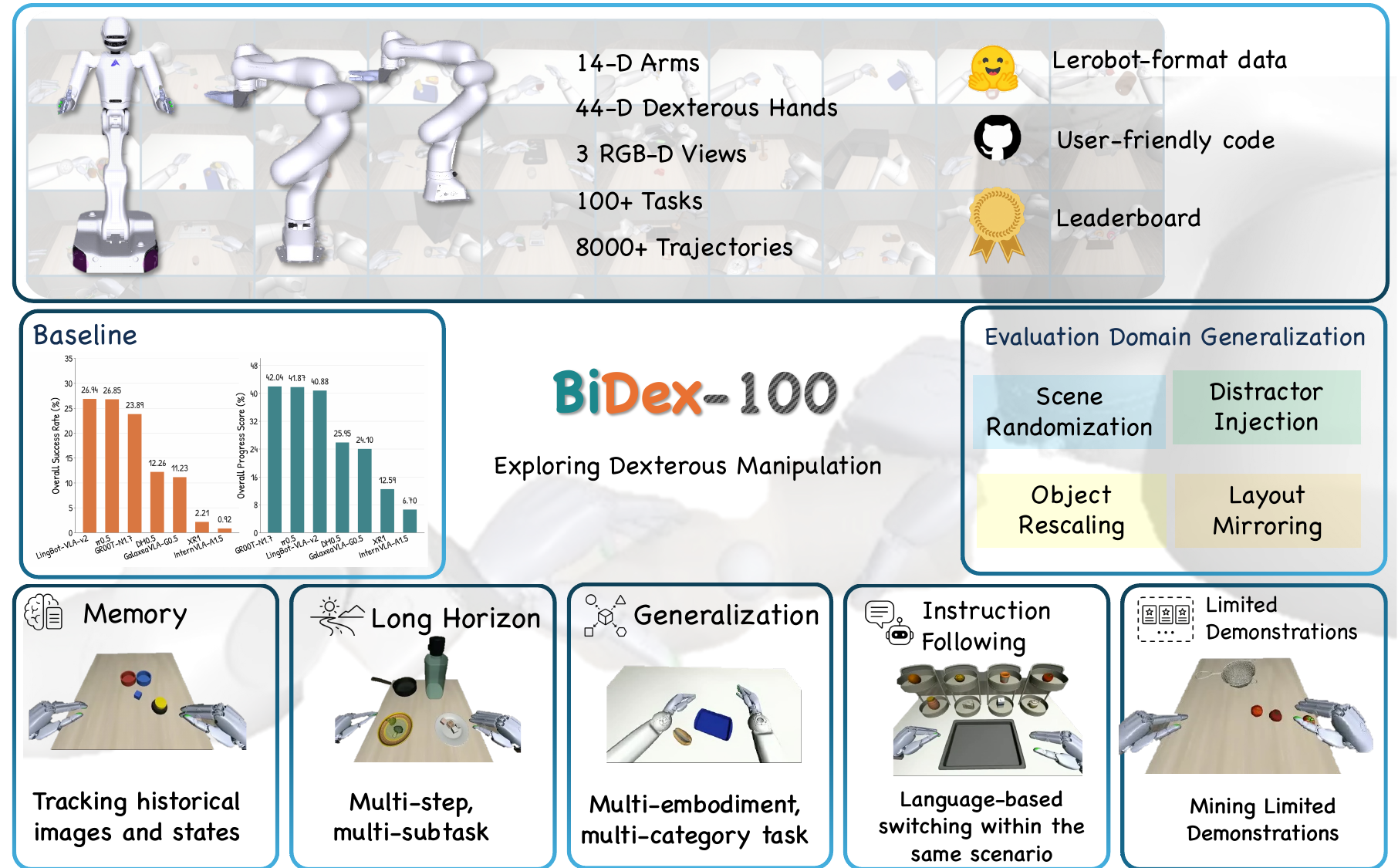}%
    \captionof{figure}{\textbf{Overview of \methodname.} The benchmark provides 106 declarative task specifications across five capability axes, 8,080 success-validated teleoperated demonstrations on two bimanual embodiments equipped with dexterous hands, and employs progress-based scoring under matched In-Domain and Out-of-Domain protocols. Five representative tasks, one per capability axis, each with a real demonstration and its language instruction, cover Memory, Long-Horizon, Instruction-Following, Limited-Demonstrations, and Generalization.}~\label{fig:overview}%
  \endgroup
  \vspace{\baselineskip}%
}

\maketitle
\thispagestyle{empty}
\pagestyle{empty}

\begin{abstract}

Bimanual manipulation with dexterous hands has attracted growing attention in research on robotic action models. However, few studies have developed benchmarks specifically tailored for dexterous manipulation, leaving the dexterous manipulation capabilities of existing models insufficiently characterized. In this paper, we present BiDex-100, a simulation benchmark for bimanual dexterous manipulation that combines broad task coverage, multi-dimensional evaluation, and high-DoF hand embodiments. Concretely, the benchmark comprises 106 tasks organized along five capability axes: generalization, instruction following, limited demonstrations, long horizon, and memory. Each task is paired with an out-of-domain variant featuring controlled visual and spatial changes, allowing systematic assessment of robustness. In addition, the benchmark supports two embodiments and provides 8,080 success-validated teleoperated demonstrations, totaling 6.25 million frames. We evaluate seven recent Vision-Language-Action models under a unified protocol, establishing baselines for capability-specific performance, robustness, and data efficiency. The website is available at \url{https://robbyant-research.github.io/BiDex-100/}.

\end{abstract}


\section{INTRODUCTION}
\label{sec:intro}

Bimanual dexterous manipulation, the coordinated control of two multi-fingered hands acting on a shared workspace, is widely regarded as a prerequisite for general-purpose household and industrial robots. The past two years have brought this capability within reach: VLA models~\cite{brohan2023rt2,kim2024openvla,black2024pi0,intelligence2025pi05,bjorck2025gr00tn1} now absorb large-scale demonstrations and generalize across scenes, objects, and instructions. What has \emph{not} kept pace is our ability to measure where such policies actually fail. As training corpora grow, average success on narrow pick-and-place suites saturates, and the discriminating power of a benchmark shifts away from task coverage toward \emph{capability edges}: remembering scene state, sustaining long horizons, obeying instructions that change mid-episode, learning from few demonstrations, and degrading gracefully under visual and geometric shift. A 58-DoF bimanual dexterous system amplifies every one of these edges relative to the 7-DoF gripper platforms that dominate current evaluation. Its action space is high-dimensional and redundant, its success hinges on contact-rich finger coordination, and its failure modes are invisible to a binary success rate.

Existing manipulation benchmarks leave this gap open in three distinct ways. First, the widely used suites, among them Meta-World~\cite{yu2020meta}, RLBench~\cite{james2020rlbench}, LIBERO~\cite{liu2023libero}, CALVIN~\cite{mees2022calvin}, RoboCasa~\cite{nasiriany2024robocasa}, and ManiSkill~\cite{gu2023maniskill2}, are built around single arms terminated by parallel-jaw grippers, so the entire dexterity dimension is abstracted away. Second, dexterous-hand benchmarks such as Bi-DexHands~\cite{chen2022bidexhands} and DexMV~\cite{qin2022dexmv} target reinforcement learning from scratch on small task sets, without language grounding, large-scale demonstrations, or automated compositional scoring. Third, even recent large-scale efforts, such as RoboTwin~\cite{mu2025robotwin,chen2025robotwin} and RoboDojo~\cite{chen2026robodojo}, evaluate bimanual gripper platforms and report largely binary success, while the community has converged on the need for capability-targeted diagnosis of memory, long-horizon competence, and instruction fidelity that binary scores cannot provide. In short, the field lacks a benchmark that simultaneously offers bimanual \emph{dexterous} hands, a capability-organized task suite with progress-based predicate scoring, matched In-Domain and Out-of-Domain evaluation, and a large teleoperated demonstration dataset released alongside the benchmark.

To address this gap, we introduce BiDex-100, a MuJoCo benchmark of 106 bimanual dexterous manipulation tasks spanning five capability axes. BiDex-100 is built on a simple premise: as average task success improves, evaluation should reveal not only whether a policy succeeds, but also what it can do reliably and where it fails. Our contributions are:

\begin{enumerate}[leftmargin=*]
    \item A capability-based benchmark for bimanual dexterity. The 106 tasks measure Generalization, Instruction-Following, Limited Demonstrations, Long Horizon, and Memory. These axes expose distinct limitations of VLA policies. Results therefore form a capability profile rather than a single aggregate score.

    \item Matched In-Domain (ID) and Out-of-Domain (OOD) evaluation. Each task has a reproducible OOD counterpart with controlled changes to appearance, scene layout, distractors, object scale, and left-right configuration. This design makes in-domain performance and robustness directly comparable.

    \item A large-scale dataset and reference study. We provide 8,080 success-validated teleoperation episodes, covering 6.25M frames of data across the 106 released tasks. We also evaluate 7 recent VLA models under a unified protocol, establishing baselines for capability-specific performance, robustness, and data efficiency.
\end{enumerate}

\section{RELATED WORK}
\label{sec:related}

\paragraph{Manipulation benchmarks.}
The field's standard suites measure gripper-based manipulation. Meta-World~\cite{yu2020meta} and RLBench~\cite{james2020rlbench} offer 50--100 tasks for meta-RL and few-shot learning. LIBERO~\cite{liu2023libero} studies lifelong transfer across 130 tasks. CALVIN~\cite{mees2022calvin} targets language-conditioned long-horizon chains. RoboCasa~\cite{nasiriany2024robocasa} and BEHAVIOR-1K~\cite{li2023behavior} scale scene and activity diversity. ManiSkill2~\cite{gu2023maniskill2} and VLABench~\cite{zhang2025vlabench} emphasize generalizable skills and language grounding. SimplerEnv~\cite{li2024evaluating} aligns simulation with real-world policy rankings. All of these terminate the arm in a parallel-jaw gripper and score success binarily. RoboTwin~\cite{mu2025robotwin} and RoboTwin~2.0~\cite{chen2025robotwin} move to bimanual grippers with generative digital twins and strong domain randomization, and RoboDojo~\cite{chen2026robodojo} organizes 60 gripper tasks along capability dimensions with a unified sim-and-real protocol, but neither exercises multi-fingered dexterity, and their scoring remains predominantly binary. DexJoCo~\cite{wang2026dexjoco} is a bimanual dexterous benchmark with teleoperated demonstrations and matched ID/OOD variants, but only 11 tasks, binary scoring. BiDex-100 extends the design to 106 capability-organized tasks with progress-based predicate scoring. BiDex-100 occupies the vacant corner of this design space (Table~\ref{tab:benchmarks}): bimanual 22-DoF dexterous hands, five capability axes, progress-based predicate scoring, and matched ID/OOD evaluation.

\paragraph{Dexterous manipulation and teleoperation.}
Dexterous-hand benchmarks historically trade breadth for fidelity: early Shadow-Hand suites~\cite{rajeswaran2017learning} cover a handful of canonical tasks; Bi-DexHands~\cite{chen2022bidexhands} provides RL environments for two Shadow hands without demonstrations or language; DexMV~\cite{qin2022dexmv} retargets human videos but offers no standardized compositional scoring. Demonstrations for such systems depend on teleoperation: DexPilot~\cite{handa2020dexpilot} and AnyTeleop~\cite{qin2023anyteleop} retarget visual hand tracking, TeleMoMa~\cite{dass2024telemoma} composes input modalities, Open-TeleVision~\cite{cheng2024opentelevision} streams immersive head-mounted video, and FACTR~\cite{liu2025factr} adds low-cost force-aware exoskeleton. BiDex-100 adopts the last two modalities and unifies them through a single optimization-based retargeting stack (Section~\ref{sec:teleop}).

\paragraph{Vision-language-action models.}
VLAs fine-tune web-scale vision-language backbones into action policies~\cite{brohan2023rt2,kim2024openvla,octo2024}, with recent designs adding flow-matching or diffusion action experts for high-frequency dexterous control~\cite{black2024pi0,intelligence2025pi05,bjorck2025gr00tn1}. Their published evaluations concentrate on gripper embodiments, and cross-model comparisons are complicated by per-paper protocols, action spaces, and data mixtures. BiDex-100 supplies the missing controlled substrate for this model class on dexterous bimanual tasks: a fixed 58-DoF joint-space interface, a shared demonstration corpus, and a stratified, automated scoring protocol, which we use to benchmark seven contemporary VLAs.

\begin{table}[t]
\caption{Benchmark comparison. ``DoF'' = degrees of freedom of the single dexterous hand (-- = parallel-jaw gripper, no multi-fingered dexterity); ``Progress'' = non-binary partial-credit scoring; ``OOD'' = built-in OOD protocol; ``Demos'' = released demonstrations (\emph{teleop.} = human-teleoperated, \emph{synth.} = scripted/synthetic, ``teleop+synth.'' = both sources, ``N/task'' = $N$ demonstrations per task, a bare count = total, \texttimes = none).}
\label{tab:benchmarks}
\centering
\scriptsize
\setlength{\tabcolsep}{3.2pt}
\begin{tabular}{lcccccc}
\toprule
\textbf{Benchmark} & \textbf{Biman.} & \textbf{DoF} & \textbf{Tasks} & \textbf{Progress} & \textbf{OOD} & \textbf{Demos} \\
\midrule
Meta-World~\cite{yu2020meta}        & \texttimes & -- & 50  & \checkmark & \checkmark & \texttimes \\
RLBench~\cite{james2020rlbench}     & \texttimes & -- & 100 & \texttimes & \texttimes & synth. \\
LIBERO~\cite{liu2023libero}         & \texttimes & -- & 130 & \texttimes & \texttimes & teleop. \\
CALVIN~\cite{mees2022calvin}        & \texttimes & -- & 34  & partial & \checkmark & teleop \\
RoboCasa~\cite{nasiriany2024robocasa} & \texttimes & -- & 100 & \texttimes & \checkmark & teleop+synth. \\
ManiSkill2~\cite{gu2023maniskill2}  & \checkmark & -- & 20 & \texttimes & \checkmark & synth. \\
RoboTwin 2.0~\cite{chen2025robotwin} & \checkmark & -- & 50 & \texttimes & \checkmark & synth. \\
RoboDojo~\cite{chen2026robodojo}    & \checkmark & -- & 60  & \checkmark & \checkmark & teleop+synth. \\
DexJoCo~\cite{wang2026dexjoco}    & \checkmark & 16 & 11  & \texttimes & \checkmark & teleop. \\
\midrule
\textbf{BiDex-100 (ours)}         & \checkmark & \textbf{22} & \textbf{106} & \checkmark & \checkmark & \textbf{8{,}080 teleop} \\
\bottomrule
\end{tabular}
\end{table}

\section{THE BiDex-100 BENCHMARK}
\label{sec:benchmark}

\subsection{Design Principles}
\label{sec:principles}

Three principles guide BiDex-100. (1) Dexterity is the object of study. Tasks require dexterous hands for skills beyond conventional grippers, such as hammering nails and watering. (2) Diagnose capabilities, not tasks. Tasks are organized by the capability they stress, namely visual and geometric generalization, instruction fidelity, data efficiency, horizon depth, and latent-state memory, rather than by scene or object type. A leaderboard position therefore decomposes into an interpretable capability profile. (3) Scores must be reproducible and comparable. Machine-checkable predicate conjunctions, deterministic seeded evaluation, progress-based subtask credit, and matched ID/OOD variants replace human raters and per-paper protocols.

\subsection{Embodiments and Interfaces}
\label{sec:embodiments}
BiDex-100 supports two robot embodiments with the same dexterous hand design (Fig.~\ref{fig:overview}). The dual Franka platform consists of two 7-DoF Franka Research 3 arms~\cite{franka2023fr3} mounted on a shared base, each equipped with a 22-DoF Dexterous hand~\cite{sharpa2024wave}. It has a 58-DoF action space and executes 96 of the 106 task specifications. The remaining 10 Generalization tasks execute on the Astribot S1 humanoid~\cite{astribot2024s1}, which shares the same hands on two 7-DoF arms. Every task is evaluated on its designated embodiment under the unified protocol; consequently, the Generalization column in Tables~\ref{tab:results_id}--\ref{tab:results_ood} aggregates 30 dual-Franka and 10 Astribot tasks.
Observations include synchronized RGB-D images from head, left-wrist, and right-wrist cameras at $480 \times 360$ resolution and 25\,Hz, together with proprioceptive states. The latter include arm and hand joint positions and velocities, as well as absolute end-effector poses. Actions are 58-dimensional joint targets,
$\mathbf{u}_t = [\mathbf{q}^{\mathrm{arm}}_{\mathrm{tgt}}, \mathbf{q}^{\mathrm{hand}}_{\mathrm{tgt}}]$,
executed by position-controlled actuators. End-effector control is also available and is converted to joint targets through inverse kinematics. Simulation runs in MuJoCo~\cite{todorov2012mujoco} at 500\,Hz, with control at 25\,Hz.

\begin{figure}[t]
  \centering
  \includegraphics[width=\linewidth]{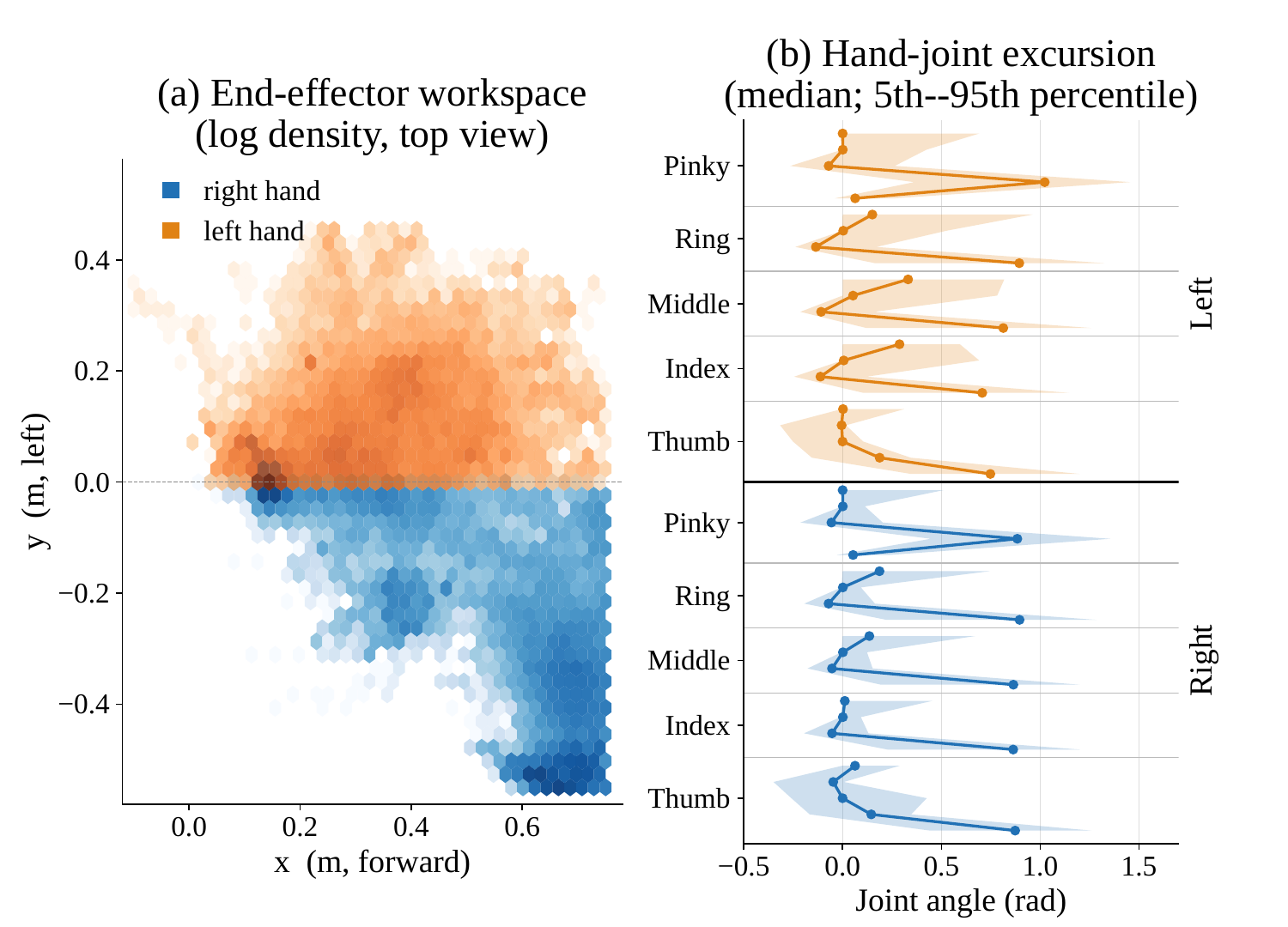}
  \caption{\textbf{Action Space Visualization.} (a) Top-view log-density of end-effector positions across all 8{,}080 demonstrations; the two hands partition but substantially overlap the shared workspace, forcing bimanual coordination. (b) Per-joint usage of the bimanual 44 hand DoF, shown as median and 5--95\% band; all five fingers engage, and thumb CMC and finger MCP joints carry the largest excursions.}
  \label{fig:dexterity}
\end{figure}

\begin{figure}[t]
  \centering
  \includegraphics[width=\linewidth]{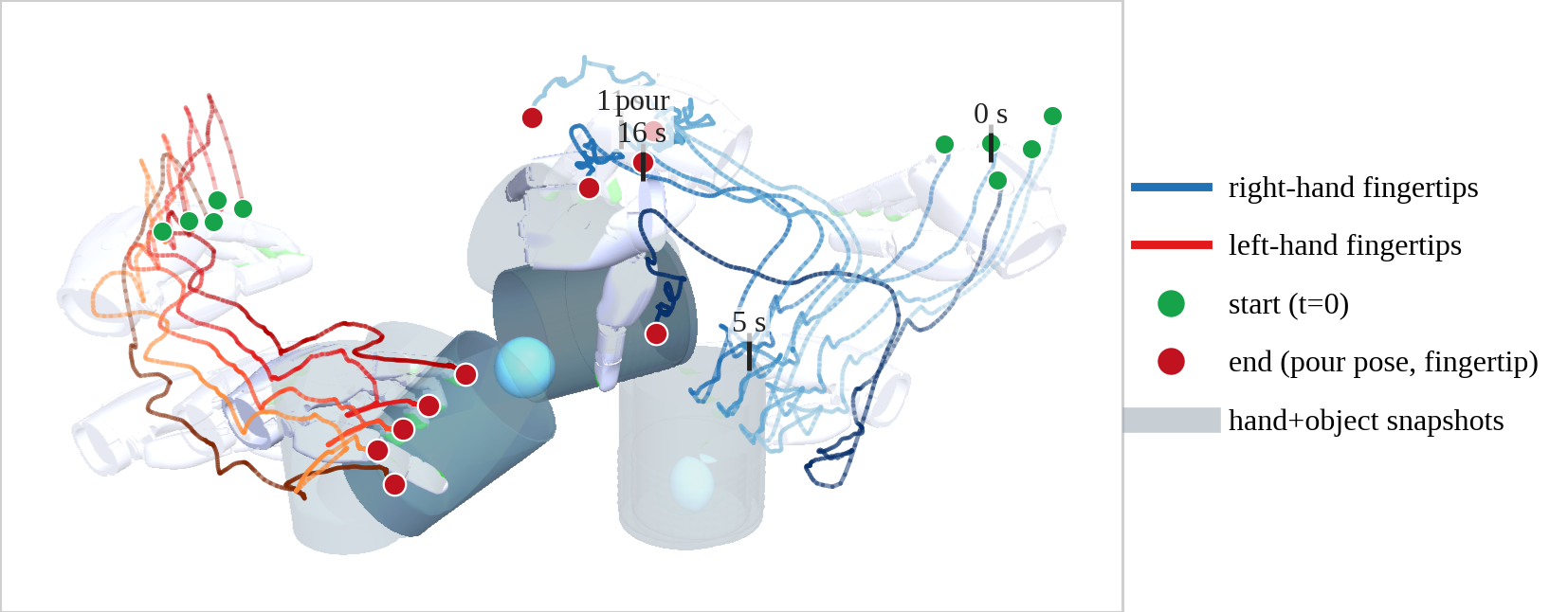}
  \caption{\textbf{Bimanual Collaborative Task.} Fingertip trajectories of all ten fingers over the full motion reveal how contact-rich manipulation unfolds within a single episode. The right hand lifts the source cup above the recipient and tilts it to transfer the ice ball between the two cups, operating within centimeters while the fingers reconfigure among open-palm transport, precision pinch, and power grasp.}
  \label{fig:replay}
\end{figure}

\begin{figure}[t]
  \centering
  \includegraphics[width=\linewidth]{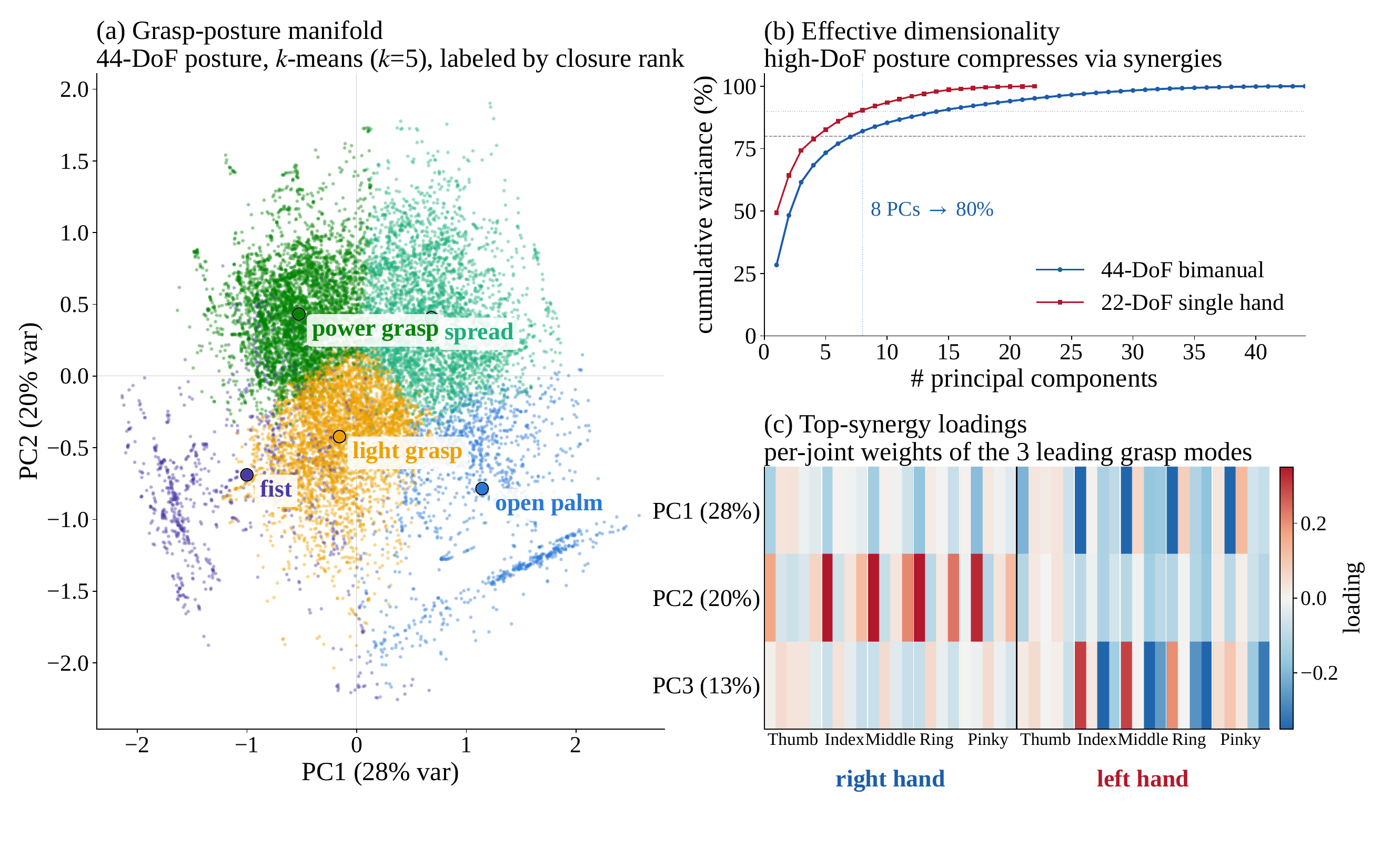}
  \caption{\textbf{Dexterity Analysis.} (a) Two-dimensional PCA of the bimanual 44-DoF posture over all 8{,}080 demonstrations; $k$-means ($k{=}5$) clusters, labeled by mean grasp-closure rank, span a continuous manifold from open palm to fist. (b) Cumulative PCA variance of the bimanual 44-DoF and single-hand 22-DoF posture: only eight components capture 80\% of motion, so the high-DoF hand is governed by a compact synergy basis. (c) Per-joint loadings of the three leading grasp modes, grouped by finger for the right and left hands.}
  \label{fig:dexsynergy}
\end{figure}

The dexterous hands are central to BiDex-100. Fig.~\ref{fig:replay} visualizes the Memory task \textit{pour ice into cups} by overlaying solid hand and object snapshots with time-gradient fingertip trajectories and markers that terminate at the pour instant. The right hand lifts and tilts the source cup above the recipient to transfer the ice ball while both hands coordinate within centimeters and reconfigure among open-palm transport, precision pinch, and power grasp. Fig.~\ref{fig:dexterity}(a) shows that the end effectors partition but substantially overlap a shared workspace, and Fig.~\ref{fig:dexterity}(b) shows substantial motion across every finger. Fig.~\ref{fig:dexsynergy} reveals that the bimanual posture follows a low-dimensional manifold: eight principal components capture 80\% of postural variance, indicating a compact set of reusable grasp synergies.

\subsection{Declarative Task Formalism and Scoring}
\label{sec:formalism}

Each task is a declarative specification
\begin{equation}
\label{eq:task}
\mathcal{T} \;=\; \big(\mathcal{R},\; \mathcal{O},\; \mathcal{A},\; \mathcal{G},\; \mathcal{L}\big),
\end{equation}
comprising named placement \emph{regions} $\mathcal{R}$, each a center with an $xy$- and yaw-range from which object poses are sampled uniformly per episode, movable \emph{objects} $\mathcal{O}$, \emph{articulation} initial states $\mathcal{A}$, a compositional \emph{goal} $\mathcal{G}$, and \emph{language} $\mathcal{L}$, possibly a \emph{schedule} of instructions that changes mid-episode. The goal is a conjunction of $K$ evaluation subtasks,
\begin{equation}
\label{eq:goal}
\mathcal{G} \;=\; \bigwedge_{k=1}^{K} S_k,
\qquad
S_k \;=\; \bigwedge_{p \in \mathcal{P}_k} \phi_p(\mathbf{s}_t),
\end{equation}
where each predicate $\phi_p$ is drawn from a typed, threshold-parameterized library spanning geometric checks for containment, support, and pose, event and memory predicates that count or recall discrete state, and negation predicates that police forbidden outcomes.

Subtask completion follows \emph{latching semantics}: with $c_k(t)$ the instantaneous truth of $S_k$, the latched state $\ell_k(t)=\max_{\tau\le t} c_k(\tau)$ is monotone, so progress cannot be undone by later disturbances. Three mechanisms harden this scheme against degenerate solutions. \emph{Must-hold} predicates must additionally be true at episode end, defeating touch-and-go strategies; \emph{latch-violation} predicates permanently invalidate a subtask if a forbidden state ever occurs, such as watering the wrong plant; and \emph{ordered} subtasks gate later credit behind earlier completion. Let $m_k$ indicate that subtask $k$ contains a must-hold predicate, let $h_k(T)$ denote its terminal truth, and let $v_k(T)$ indicate whether one of its latch-violation predicates has ever been violated. The effective completion state is $e_k(T)=\ell_k(T)(1-m_k+m_kh_k(T))(1-v_k(T))$. Each subtask carries a nonnegative integer weight $w_k$ with $\sum_k w_k = 100$, defining strict success and the progress score as
\begin{equation}
\label{eq:score}
\text{success} \;=\; \prod_{k=1}^{K}e_k(T),
\qquad
\text{score} \;=\; \sum_{k=1}^{K} w_k\,e_k(T),
\end{equation}
which preserves a strict success criterion while rewarding partial progress. Strict success is evaluated independently of the numeric score because a small number of safety or terminal constraints carry zero weight. Across the 106 specifications, tasks carry 1 to 10 subtasks with mean 3.9 and compose 470 predicate instances; Fig.~\ref{fig:dataset_stats}(c,d) profiles this composition.

\subsection{Five Capability Axes}
\label{sec:axes}

\begin{figure}[t]
  \centering
  \includegraphics[width=\linewidth]{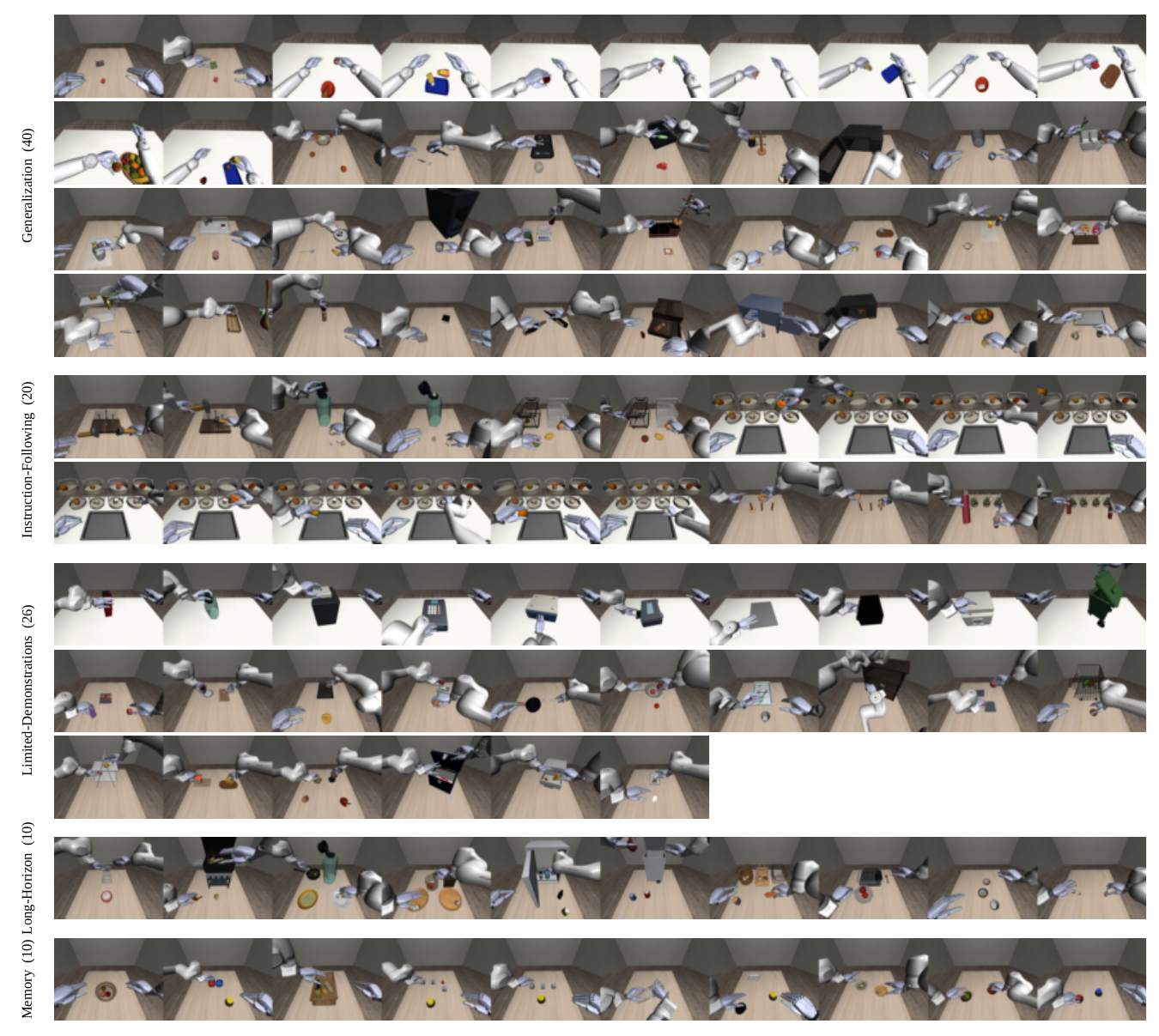}
  \caption{\textbf{Overview of All Tasks.} Full details on the tasks and data are available in Section~\ref{app:tasks}.}
  \label{fig:106}
\end{figure}

The suite design turns ``is the policy good?'' into ``\emph{what} is the policy good at?'' by evaluating it across the following dimensions:

\textbf{Generalization (40 tasks).} Single- and few-stage skills, including lifting, placement, insertion, threading, and cooking, over a diverse object library spanning food, tableware, containers, and tools (Fig.~\ref{fig:106}). Ten \texttt{kitchen\_*} tasks execute on Astribot S1 and the remaining 30 on dual Franka (Sec.~\ref{sec:embodiments}); all 40 are included in the main evaluation. In-domain episodes randomize object pose per reset, so the axis measures transfer across object geometry and embodiment.

\textbf{Instruction-Following (20 tasks).} Verbal-command fidelity tested three ways: \emph{count variants} share one scene but demand different object counts, for example watering \emph{one} vs.\ \emph{two} plants or hammering \emph{one} vs.\ \emph{two} nails per hand; \emph{combo} tasks conjoin two instructions; and \emph{switch} tasks replace the instruction mid-episode via a language schedule, so the policy must serve the \emph{current} command rather than a remembered one. 

\textbf{Limited-Demonstrations (26 tasks).} Low-data skill acquisition: each task ships with only 30 demonstrations, versus 100 elsewhere, defining a fixed data-scarce regime. The axis comprises 10 articulated-mechanism tasks, including microwaves, cash registers, and rice cookers, needing force-aware actuation of hinged, sliding, and rotary joints, where bimanual 44-DoF hand coordination is decisive.

\textbf{Long-Horizon (10 tasks).} Multi-stage chores, such as barbecue grilling, fridge restocking, dessert plating, and recycling sorting, with 4 to 9 subtasks, 6.5 objects per scene, and demonstration durations averaging 48.3\,s in the sampled statistics. Success requires sustaining a plan far beyond the policy's action chunk.

\textbf{Memory (10 tasks).} Latent-state tracking under perceptual aliasing: in \emph{color-swap cups}, a cube's color is visually removed after a button sequence and must be recalled to choose the matching cup; in \emph{picnic-basket FIFO}, removal order must match insertion order; in \emph{memory cups}, a reveal phase shows target contents that are later hidden. These tasks are impossible from a single frame and probe genuine temporal state tracking rather than reactive control.

\subsection{Matched In-Domain / Out-of-Domain Evaluation}
\label{sec:ood}

\begin{figure}[t]
  \centering
  \includegraphics[width=\linewidth]{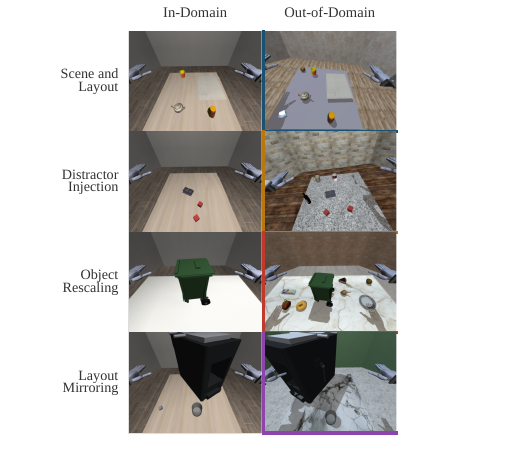}
  \caption{\textbf{Matched ID / OOD Pairs.} Rows show scene and layout changes, distractor injection, articulated-object rescaling, and left-right layout mirroring. The first row combines scene randomization with the task-specific OOD layout. The remaining rows illustrate optional transformations composed with the same OOD protocol. The task goal is unchanged, and each task, seed, and episode key is reproduced identically across models.}
  \label{fig:ood}
\end{figure}

All 106 tasks are accompanied by OOD variants generated by six deterministic transformation families (Fig.~\ref{fig:ood}): (i)~\emph{scene randomization} redraws floor, wall, and tabletop textures from a held-out pool with lighting position and intensity jitter; (ii)~\emph{camera variation} selects validated head and wrist camera extrinsics; (iii)~\emph{layout variation} applies task-specific placement and geometry changes recorded in the OOD specification; (iv)~\emph{distractor injection} adds 3 to 9 objects to 45 compatible tasks, where grasping a distractor immediately terminates the episode at its current score; (v)~\emph{object rescaling} assigns 10 articulated Limited-Demonstrations tasks a smaller or larger scale regime anchored by a factor in $\{0.85,0.9,1.1,1.15\}$ and samples a deterministic factor within that regime for each episode; and (vi)~\emph{left-right mirroring} uses an arm-mirrored layout in 9 tasks. Every OOD task uses scene and camera randomization together with a committed OOD layout. Additional transformations are enabled only when compatible with the intended task. The committed YAML fixes the geometry and applicable transformations, while appearance, camera, articulated scale, and distractor placement are derived reproducibly from the task, seed, and episode. Thus every model receives the same conditions for a given evaluation key. The variants are evaluation-only: training and in-domain rollouts use the base specification, so no OOD episode enters the training set.

\section{THE BiDex-100 DATASET}
\label{sec:dataset}

\subsection{Demonstration Collection}
\label{sec:teleop}

\begin{figure}[t]
  \centering
  \begin{subfigure}[b]{0.48\linewidth}
    \centering
    \includegraphics[width=\linewidth]{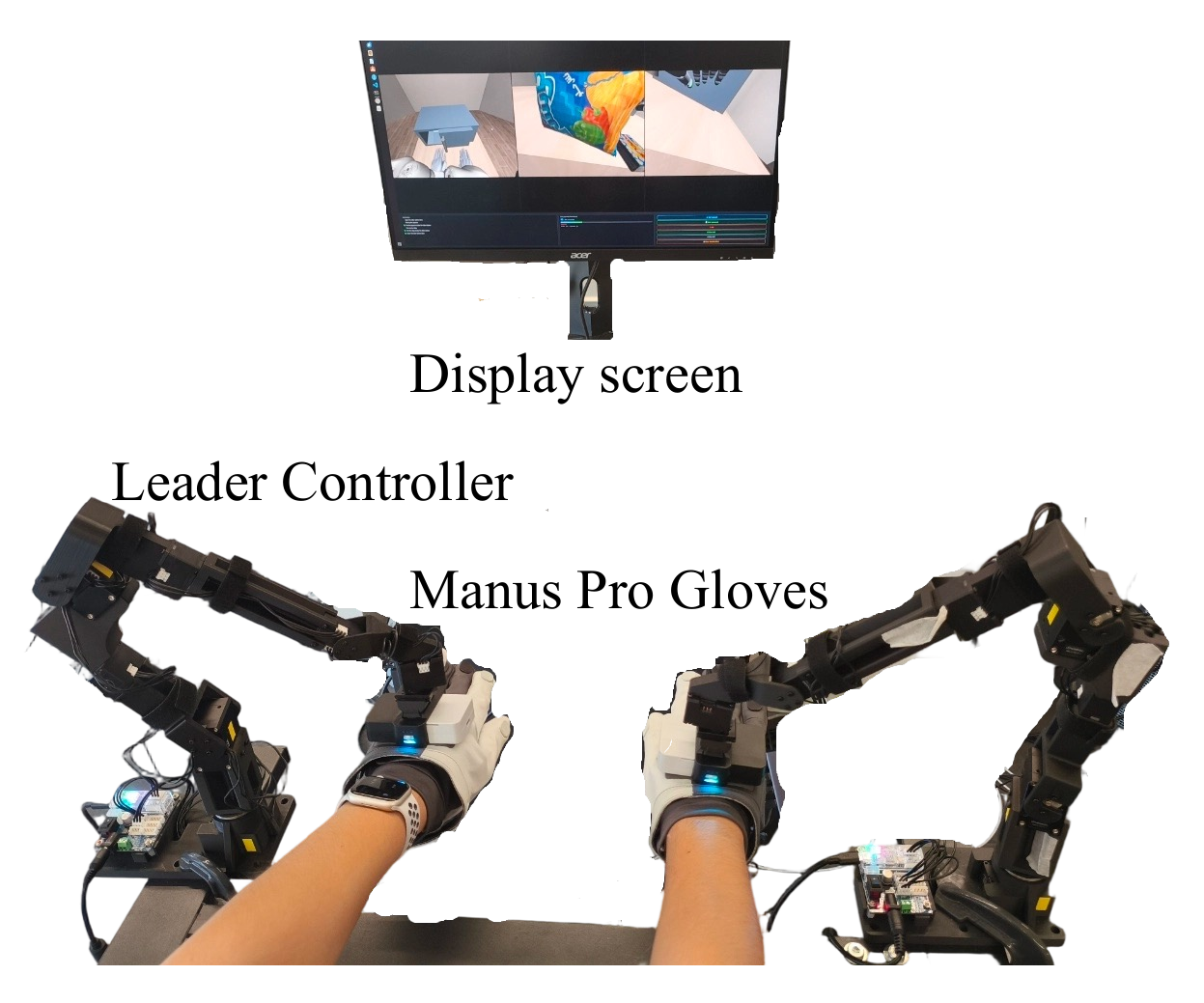}
    \caption{Exoskeleton-style glove}
  \end{subfigure}
  \hfill
  \begin{subfigure}[b]{0.48\linewidth}
    \centering
    \includegraphics[width=\linewidth]{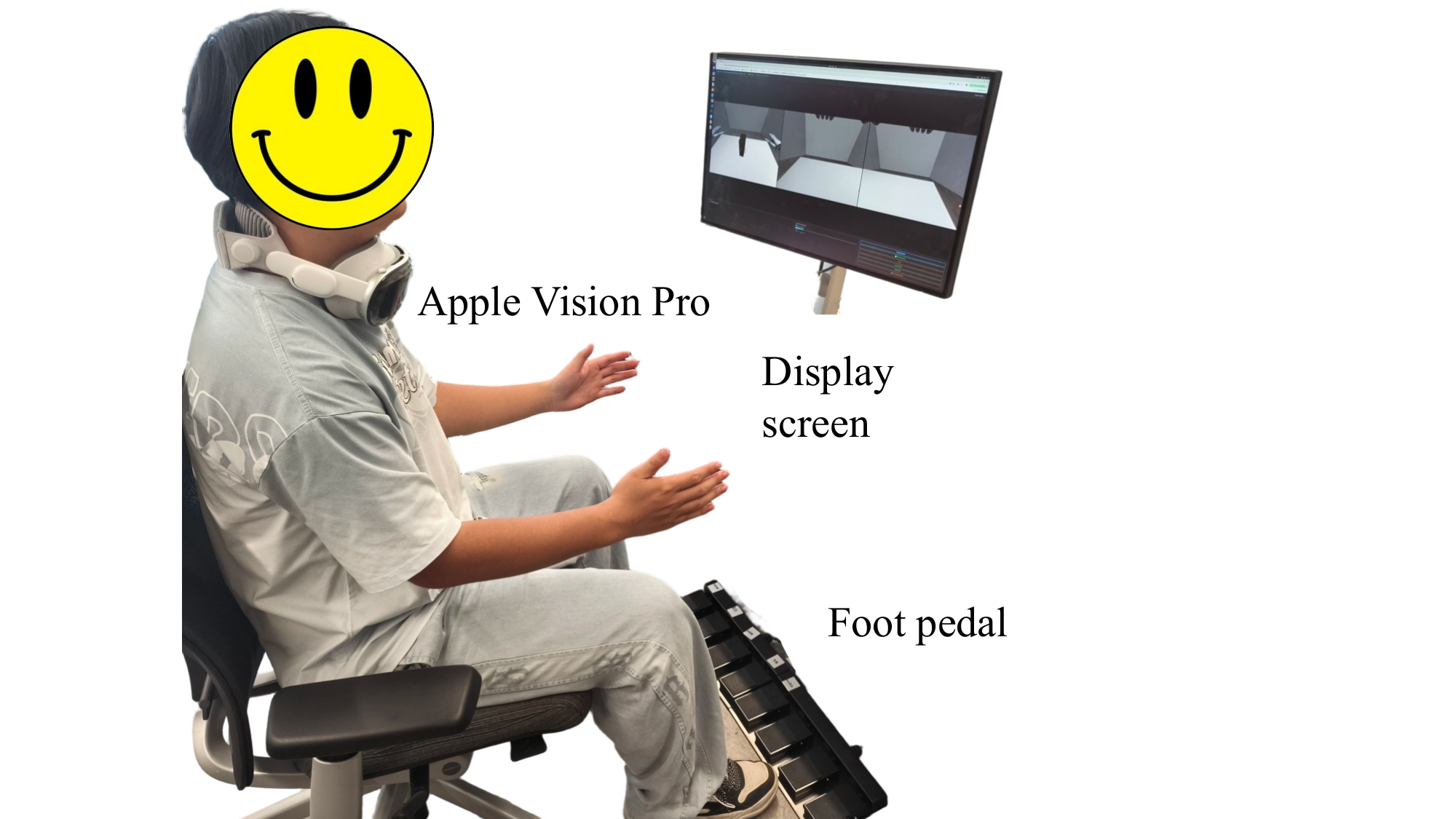}
    \caption{Apple Vision Pro headset}
  \end{subfigure}
  \caption{\textbf{Teleoperation hardware} used for demonstration collection. (a) Joint-level tracking via an exoskeleton glove; (b) wrist-pose and hand-keypoint streaming from a head-mounted device.}
  \label{fig:teleop}
\end{figure}

Demonstrations are teleoperated on the simulated platforms through two complementary modalities (Fig.~\ref{fig:teleop}): a FACTR-style exoskeleton glove~\cite{liu2025factr} streaming joint-level finger tracking for high-dexterity work, and an Apple Vision Pro~\cite{apple2024visionpro} streaming wrist poses and 25 hand keypoints over WebXR~\cite{cheng2024opentelevision} for untethered operation. Offering both lets operators match the device to the task while producing a single, homogeneous data stream.

Both modalities feed one retargeting stack, producing a unified 58-DoF schema across platforms and devices. At the control rate, a constrained nonlinear program maps arm targets over the first five joints of each 7-DoF arm, minimizing weighted position, orientation, and temporal-smoothness costs under joint limits with SLSQP~\cite{kraft1988slsqp}. The two remaining wrist DoF are recovered in closed form from forearm-wrist geometry, and shoulder-anchored Procrustes scaling handles arm-length mismatch. A temporally regularized weighted least-squares fit maps robot hand keypoints, obtained from forward kinematics, to tracked human keypoints. A fixed linear node map converts headset keypoints to the glove representation, allowing both modalities to share the solver. The mapping is parameterized only by embodiment-specific kinematics, joint bounds, and scale factors.

\subsection{Dataset Contents and Statistics}
\label{sec:datastats}

The release contains 8{,}080 \emph{success-validated} episodes over 106 tasks: 100 per task on Memory, Long-Horizon, Generalization, and the 10 core Instruction-Following tasks, and 30 per task on Limited-Demonstrations and the 10 shelf-transfer Instruction-Following variants. The 6.25\,M frames of 25\,Hz teleoperation occupy 94\,GB. Each episode stores six synchronized $480{\times}360$ color and depth streams from the head and wrist cameras, 58-DoF actions, 130-dimensional observations, and subtask progress. LeRobot v2.1 and v3.0 layouts~\cite{cadene2024lerobot} are provided as parquet and H.264 mp4.

\begin{figure}[t]
  \centering
  \includegraphics[width=\linewidth]{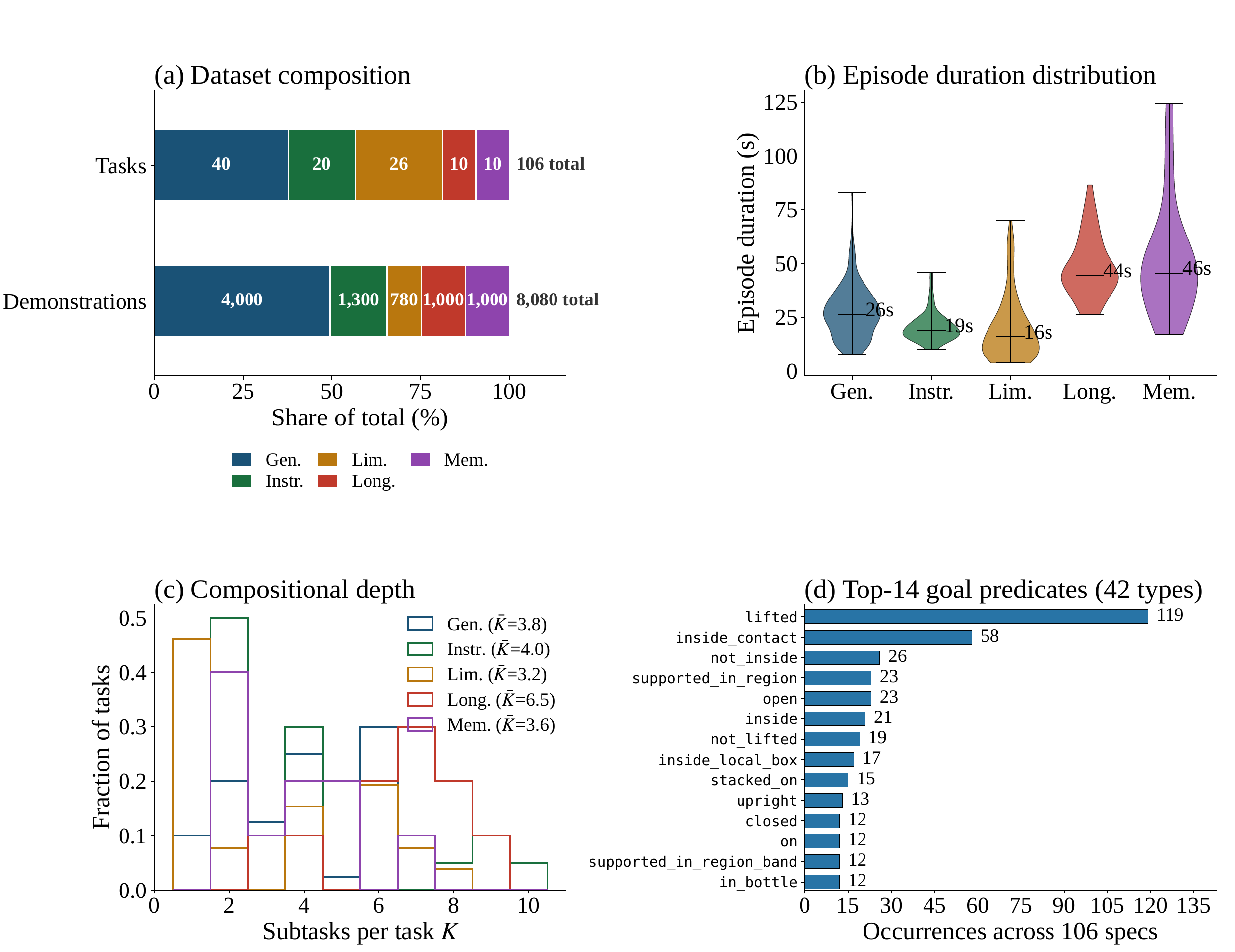}
  \caption{\textbf{Dataset Statistics.} (a) Distribution of the 106 tasks and 8{,}080 demonstrations across the five capability axes. (b) Episode duration distributions, with the median marked for each axis. (c) Distribution of compositional depth measured by the number of subtasks per task. (d) The 14 most frequently used goal predicates among 42 predicate types across the 106 task specifications.}
  \label{fig:dataset_stats}
\end{figure}

\begin{figure}[t]
  \centering
  \includegraphics[width=\linewidth]{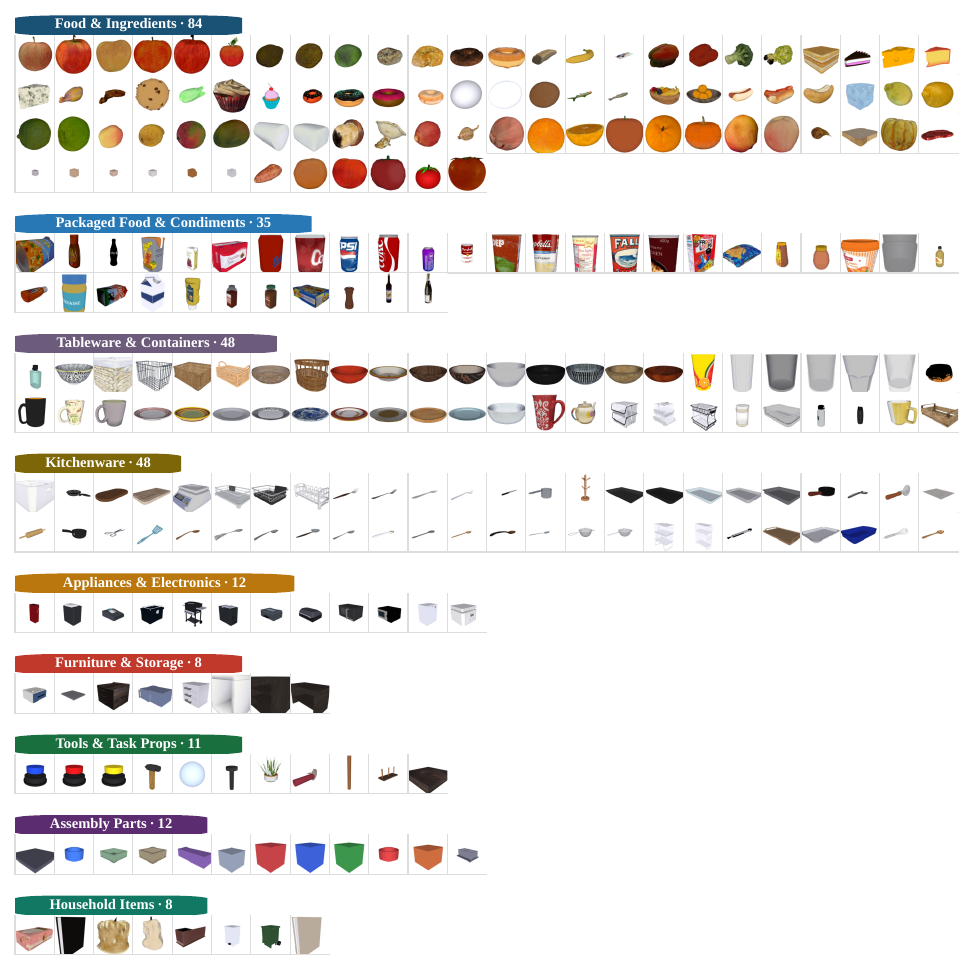}
  \caption{\textbf{Object Asset Library.} The 266 unique visible assets used by the benchmark, grouped semantically with the count shown for each category. Every asset is simulated with its mesh, mass, and friction properties.}
  \label{fig:objects}
\end{figure}

Task scenes draw on 266 unique visible assets (Fig.~\ref{fig:objects}) across nine semantic categories. Sources include RoboCasa-Objaverse~\cite{nasiriany2024robocasa}, Lightwheel, HOPE, scanned items, Turbosquid, Articraft~\cite{zhou2026articraft}, and purpose-built assembly and control primitives. Fig.~\ref{fig:dataset_stats} summarizes the corpus. Long-Horizon and Memory episodes average 48.3\,s and 50.9\,s, versus 19.8 to 27.5\,s for the other axes. Their compositional-depth and predicate-usage profiles also differ, confirming distinct task structures across the five suites.

\section{EXPERIMENTS}
\label{sec:experiments}

\subsection{Baselines and Training Protocol}
\label{sec:baselines}

We benchmark seven recent VLA models spanning the major action-expert designs (Table~\ref{tab:models}): $\pi_{0.5}$~\cite{intelligence2025pi05}, GR00T-N1.7~\cite{bjorck2025gr00tn1}, InternVLA-A1.5~\cite{ma2026internvla}, Galaxea~G0.5~\cite{liu2026g0}, OpenDM-DM05~\cite{dexmal2025dm05}, LingBot-VLA-v2~\cite{wu2026foundation}, and Xiaomi~XR1~\cite{team2026xiaomi}. Their official checkpoints are co-trained by jointly sampling one mixture of demonstrations from all 106 tasks and five axes, using the same 58-DoF joint-space interface. Optimizer, schedule, freezing strategy, and training steps follow each model's official configuration. When an action head mismatches the interface, its projection layers are resized and re-initialized by the model's standard procedure.

\begin{table}[t]
\caption{Evaluated VLA baselines. ``Chunk'' = action chunk length; ``Train'' = fine-tuning optimizer steps, kept at each model's \emph{official} configuration. All models use 58-DoF joint control and are co-trained on one mixture containing demonstrations from all 106 tasks.}
\label{tab:models}
\centering
\scriptsize
\setlength{\tabcolsep}{3pt}
\begin{tabular}{llccc}
\toprule
\textbf{Model} & \textbf{VLM backbone} & \textbf{Action expert} & \textbf{Chunk} & \textbf{Train} \\
\midrule
$\pi_{0.5}$~\cite{intelligence2025pi05} & PaliGemma-2B & 300M flow-matching & 50 & 60k \\
GR00T-N1.7~\cite{bjorck2025gr00tn1} & Cosmos-R2-2B & DiT flow-matching & 16 & 400k \\
InternVLA-A1.5~\cite{ma2026internvla} & Qwen3.5-2B & 1B flow-matching & 50 & 400k \\
Galaxea G0.5~\cite{liu2026g0} & Qwen3.5-2B & flow-matching & 32 & 400k \\
OpenDM-DM05~\cite{dexmal2025dm05} & Gemma3-4B & rectified flow & 50 & 120k \\
LingBot-VLA-v2~\cite{wu2026foundation} & Qwen3-VL-4B & MoE flow (L1) & 50 & 60k \\
Xiaomi XR1~\cite{team2026xiaomi} & Qwen3-VL-4B & 60-D DiT flow & 30 & 400k \\
\bottomrule
\end{tabular}
\end{table}

\subsection{Evaluation Protocol}
\label{sec:protocol}

Each model is evaluated on all 106 tasks in both domains. For every task, ID evaluation uses 2 fixed seeds with 25 episodes per seed, giving 50 ID rollouts. OOD evaluation independently uses the same count, giving another 50 rollouts. Thus each model completes 100 rollouts per task and 10{,}600 rollouts in total. Episode horizons use the task's explicit \texttt{max\_steps} value when provided and otherwise use $1.2\times$ the P90~\cite{chen2026robodojo} demonstration length, so timeouts reflect task duration rather than a single global cap. We report the strict \emph{success rate} first, followed by the \emph{progress score}, the mean over rollouts in the range $[0,100]$ defined by Eq.~\eqref{eq:score}; grabbing a distractor in OOD terminates the episode at its current score. All episodes execute in the benchmark's vectorized evaluation harness with one renderer per worker. For each domain, every model receives the same task, seed, episode index, reset procedure, and OOD randomization key.

\subsection{Main Results}
\label{sec:results}

\begin{table*}[t]
\caption{In-Domain results. Success rate (\%) / progress score per capability axis, averaged over 2 seeds $\times$ 25 episodes per task.}
\label{tab:results_id}
\centering
\scriptsize
\setlength{\tabcolsep}{6pt}
\begin{tabular}{lcccccc}
\toprule
\textbf{Model} & \textbf{Generalization} & \textbf{Instruction-Foll.} & \textbf{Limited-Demo.} & \textbf{Long-Horizon} & \textbf{Memory} & \textbf{Overall} \\
\midrule
LingBot-VLA-v2~\cite{wu2026foundation}  & 61.35 / 75.80 & 35.90 / 66.57 & 66.15 / 76.14 & 38.20 / 66.78 & 26.00 / 43.20 & 52.21 / 70.22 \\
$\pi_{0.5}$~\cite{intelligence2025pi05}  & 55.45 / 72.70 & 36.60 / 64.48 & 66.23 / 75.63 & 34.00 / 64.51 & 33.00 / 50.16 & 50.40 / 68.97 \\
GR00T-N1.7~\cite{bjorck2025gr00tn1}  & 46.55 / 65.75 & 24.60 / 59.28 & 61.38 / 73.97 & 32.80 / 62.32 & 20.80 / 36.10 & 42.32 / 63.43 \\
OpenDM-DM05~\cite{dexmal2025dm05}  & 33.05 / 53.18 & 1.80 / 23.53 & 35.69 / 49.11 & 3.20 / 17.45 & 12.60 / 26.13 & 23.06 / 40.66 \\
Galaxea G0.5~\cite{liu2026g0}  & 24.90 / 44.51 & 5.70 / 36.33 & 39.62 / 53.20 & 5.20 / 25.10 & 10.60 / 28.22 & 21.68 / 41.73 \\
Xiaomi XR1~\cite{team2026xiaomi}  & 6.40 / 24.48 & 0.90 / 21.06 & 7.15 / 21.71 & 0.00 / 15.63 & 0.60 / 7.00 & 4.40 / 20.67 \\
InternVLA-A1.5~\cite{ma2026internvla}  & 0.80 / 7.49 & 0.90 / 16.66 & 4.23 / 10.99 & 0.00 / 2.34 & 0.00 / 0.96 & 1.51 / 8.98 \\
\bottomrule
\end{tabular}
\end{table*}

\begin{table*}[t]
\caption{Out-of-Domain results. Success rate (\%) / progress score per capability axis, averaged over 2 seeds $\times$ 25 episodes per task.}
\label{tab:results_ood}
\centering
\scriptsize
\setlength{\tabcolsep}{6pt}
\begin{tabular}{lcccccc}
\toprule
\textbf{Model} & \textbf{Generalization} & \textbf{Instruction-Foll.} & \textbf{Limited-Demo.} & \textbf{Long-Horizon} & \textbf{Memory} & \textbf{Overall} \\
\midrule
LingBot-VLA-v2~\cite{wu2026foundation}  & 1.60 / 10.76 & 0.80 / 19.90 & 3.46 / 10.06 & 0.00 / 7.15 & 0.80 / 6.18 & 1.68 / 11.54 \\
$\pi_{0.5}$~\cite{intelligence2025pi05}  & 2.40 / 13.19 & 2.40 / 23.96 & 6.31 / 13.54 & 0.40 / 10.02 & 3.80 / 10.70 & 3.30 / 14.77 \\
GR00T-N1.7~\cite{bjorck2025gr00tn1}  & 4.65 / 17.59 & 3.10 / 30.25 & 10.23 / 21.34 & 2.60 / 19.96 & 3.80 / 12.59 & 5.45 / 20.65 \\
OpenDM-DM05~\cite{dexmal2025dm05}  & 0.85 / 10.89 & 0.10 / 17.04 & 3.92 / 10.97 & 0.00 / 5.72 & 1.80 / 7.28 & 1.47 / 11.24 \\
Galaxea G0.5~\cite{liu2026g0}  & 0.20 / 4.40 & 0.20 / 15.41 & 2.62 / 5.75 & 0.00 / 3.66 & 0.20 / 1.57 & 0.77 / 6.47 \\
Xiaomi XR1~\cite{team2026xiaomi}  & 0.05 / 2.26 & 0.00 / 14.95 & 0.00 / 1.47 & 0.00 / 2.89 & 0.00 / 2.18 & 0.02 / 4.51 \\
InternVLA-A1.5~\cite{ma2026internvla}  & 0.05 / 1.96 & 0.00 / 13.91 & 1.23 / 3.33 & 0.00 / 1.17 & 0.00 / 1.38 & 0.32 / 4.42 \\
\bottomrule
\end{tabular}
\end{table*}

Tables~\ref{tab:results_id} and~\ref{tab:results_ood} report the capability-stratified results. Per-axis success-rate and progress-score vectors form capability profiles that distinguish short-horizon strengths from memory and long-horizon competence. ID/OOD gaps measure aggregate robustness to appearance, layout, distractor, scale, and mirroring shifts, while Limited-Demonstrations isolates acquisition from 30 demonstrations. Progress scores distinguish partial completion from failure. On ID Long-Horizon tasks, every model's success rate remains below its progress score. The highest success rate is 38.20\%, while the highest progress score is 66.78, revealing near-complete attempts hidden by binary success.

Model rankings change between ID and OOD. LingBot-VLA-v2 leads ID success at 52.21\%, versus 42.32\% for GR00T-N1.7. Under OOD shifts, GR00T-N1.7 reaches 5.45\%, versus LingBot-VLA-v2's 1.68\%, and leads all five axes, tying with $\pi_{0.5}$ on Memory. Thus, stronger in-domain performance does not necessarily imply greater OOD robustness.

Across both domains, metric choice also changes rankings: LingBot-VLA-v2 leads success rate at 26.94\%, whereas GR00T-N1.7 leads progress score at 42.04, versus LingBot-VLA-v2's 40.88. Reporting both distinguishes complete task execution from partial subtask completion.

\begin{figure}[t]
  \centering
  \includegraphics[width=\linewidth]{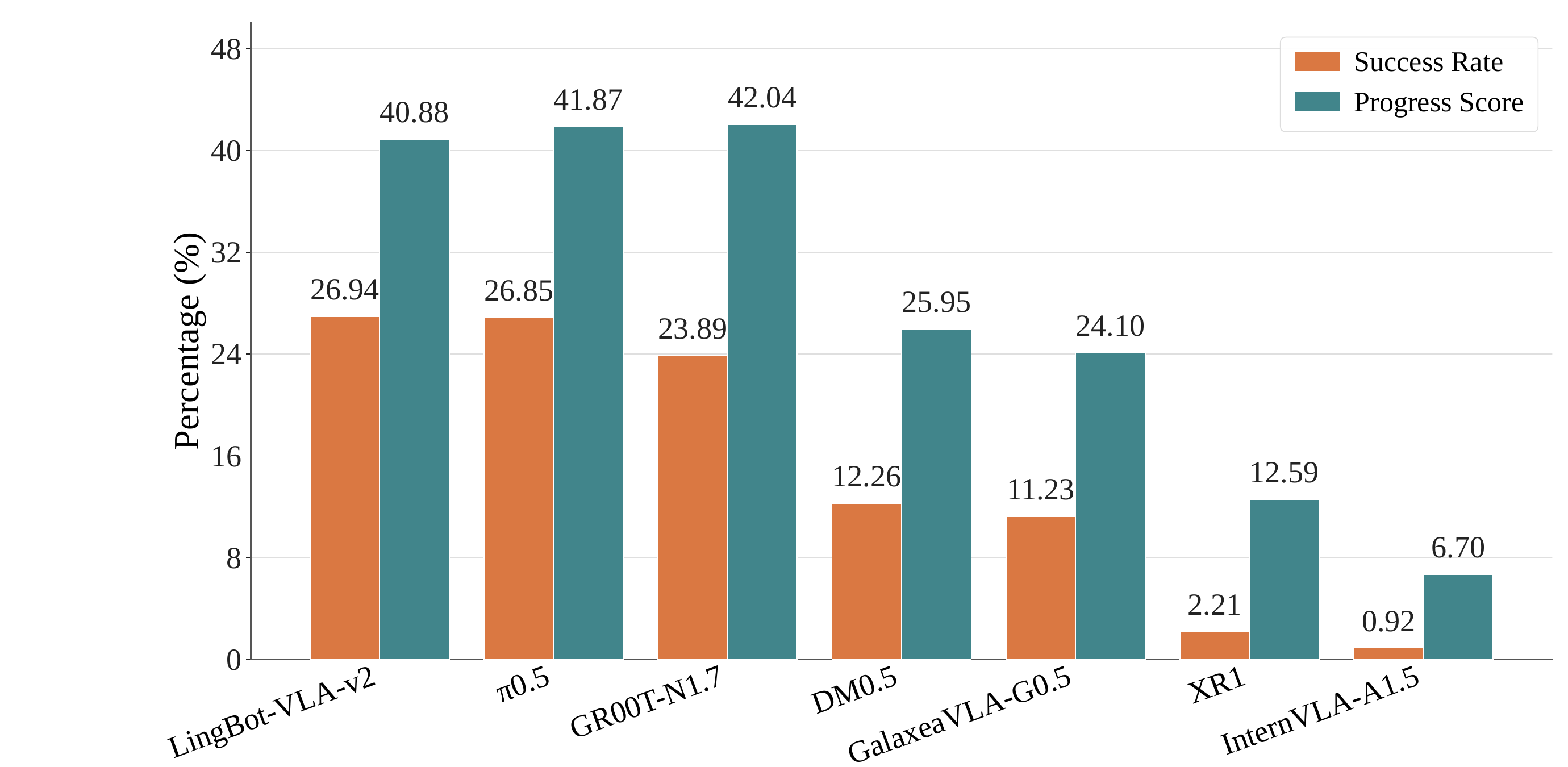}
  \caption{\textbf{Overall performance of the evaluated VLA models.} Strict success rate (\%) followed by progress score in $[0,100]$, averaged over all 106 tasks and both evaluation domains. For every task, each model is evaluated with 2 fixed seeds and 25 episodes per seed in ID, and the same protocol in OOD. This gives 50 rollouts per domain, 100 rollouts per task, and 10{,}600 rollouts per model. The progress score distinguishes partial completion from complete failure.}
  \label{fig:results_overall}
\end{figure}

Fig.~\ref{fig:results_overall} summarizes each model's overall success rate followed by its progress score. The per-domain breakdowns in Tables~\ref{tab:results_id} and~\ref{tab:results_ood} expose the corresponding ID$\to$OOD degradation across capability axes.

\section{CONCLUSION}
\label{sec:conclusion}

We presented BiDex-100, a simulation benchmark for bimanual dexterous manipulation with 106 declaratively specified tasks across five capability axes, two 22-DoF embodiments, progress-based predicate scoring, matched ID/OOD evaluation, and 8{,}080 success-validated teleoperated demonstrations. The five-axis design emphasizes capability edges as average-case skill improves, with progress-based scoring revealing differences hidden by binary success. Evaluating seven contemporary VLA models under one protocol establishes reference baselines and diagnostic capability profiles, robustness gaps, and low-data regimes. The rigid-body tabletop setting leaves sim-to-real transfer and richer physical phenomena to future work. We release all code, tasks, data, and evaluation tooling.


\bibliographystyle{IEEEtran}
\bibliography{references}


\onecolumn
\begin{center}
{\Large \bf BiDex-100: Supplementary Material}
\end{center}
\vspace{0.5em}

This supplement documents (A)~the goal-predicate catalog used by the
evaluator, (B)~the task-specification and scoring semantics,
(C)~the out-of-domain transformation rules, (D)~the dataset structure
and quality-control procedure, (E)~per-model training configurations
for the seven evaluated VLAs, (F)~the evaluation protocol, and
(G)~the complete task catalog with per-task instructions, image, and
scoring criteria, and (H)~discussion and limitations.

\section{Goal-Predicate Catalog}
\label{app:predicates}

Table~\ref{tab:predicate_catalog} lists the 42 canonical predicate types
used by the 106 released base task specifications. The evaluator additionally
accepts backward-compatible aliases for several of these types. A predicate
$\phi(\mathbf{s}_t;\,\text{args},\epsilon)$
maps the simulator state to $\{0,1\}$; \textit{args} name objects,
regions, articulated joints, or event counts, and $\epsilon$ is a per-task
tolerance. Selected negated guard predicates may adopt
\emph{latch-violation} semantics (Appendix~\ref{app:formalism}), under which
a forbidden state permanently invalidates the corresponding subtask.

\begingroup
\footnotesize
\setlength{\LTleft}{0pt}
\setlength{\LTright}{0pt}
\begin{center}
\captionof{table}{Goal-predicate catalog used by the 106 released base task
specifications (42 canonical types).}
\label{tab:predicate_catalog}
\end{center}
\begin{longtable}{@{}p{0.32\linewidth}p{0.65\linewidth}@{}}
\toprule
\textbf{Predicate} & \textbf{Semantics (tolerance $\epsilon$ unless noted)} \\
\midrule
\endfirsthead
\multicolumn{2}{c}{\tablename~\ref{tab:predicate_catalog} (continued)}\\
\toprule
\textbf{Predicate} & \textbf{Semantics (tolerance $\epsilon$ unless noted)} \\
\midrule
\endhead
\midrule
\multicolumn{2}{r}{\footnotesize Continued on the next page}\\
\endfoot
\bottomrule
\endlastfoot
\multicolumn{2}{l}{\textit{Spatial placement}} \\
\quad \texttt{on} & object center within a target's horizontal radius and above it in height \\
\quad \texttt{stacked\_on} & object one layer above another within horizontal and vertical windows \\
\quad \texttt{on\_table} & object directly supported from below by the scene table \\
\quad \texttt{inside} & object center within a circular horizontal radius and a vertical window \\
\quad \texttt{inside\_contact} & upward support contact from a container plus a minimum height \\
\quad \texttt{inside\_box} & object center inside a container's world-axis box footprint and height window \\
\quad \texttt{inside\_site} & object inside a live named site volume with contact gating \\
\quad \texttt{inside\_site\_loose} & live named-site containment without the contact requirement \\
\quad \texttt{inside\_site\_any} & contact-gated containment in any listed named site \\
\quad \texttt{inside\_site\_any\_loose} & containment in any listed named site without contact gating \\
\quad \texttt{inside\_region} & object center inside a live named three-dimensional region \\
\quad \texttt{not\_inside\_region} & exact negation of \texttt{inside\_region} \\
\quad \texttt{keypoint\_in\_site} & a selected endpoint of an elongated object inside a named site \\
\quad \texttt{supported\_in\_region} & live-region containment together with upward support contact \\
\quad \texttt{axis\_in\_region} & any sampled point on an object's long axis inside a live three-dimensional region \\
\quad \texttt{supported\_in\_region\_band} & region-band containment together with upward support contact \\
\quad \texttt{supported\_geom\_in\_region\_band} & named object-geometry center in a region band with upward support \\
\quad \texttt{supported\_in\_local\_box} & explicit container-local box containment with upward support \\
\quad \texttt{inside\_local\_box} & explicit container-local box containment without contact gating \\
\quad \texttt{inside\_any} & \texttt{inside} holds for at least one listed container \\
\quad \texttt{not\_inside} & exact negation of \texttt{inside} \\
\quad \texttt{hung\_on} & raised upward-support contact on a holder, excluding its base \\
\quad \texttt{lifted} & object height above an absolute or initialization-relative threshold \\
\quad \texttt{not\_lifted} & exact negation of \texttt{lifted} \\
\quad \texttt{upright} & object's local $z$-axis sufficiently aligned with world vertical \\
\quad \texttt{long\_axis\_vertical} & an elongated object's longest geometry axis sufficiently aligned with vertical \\
\quad \texttt{on\_edge} & object's local $z$-axis approximately horizontal \\
\midrule
\multicolumn{2}{l}{\textit{Articulation}} \\
\quad \texttt{open} & joint past a threshold toward the range endpoint farther from zero \\
\quad \texttt{open\_any} & \texttt{open} holds for any listed articulated joint \\
\quad \texttt{closed} & exact negation of \texttt{open} \\
\midrule
\multicolumn{2}{l}{\textit{State-change and event predicates}} \\
\quad \texttt{sunk} & nail displaced downward from its initial height by a required depth \\
\quad \texttt{not\_sunk} & exact negation of \texttt{sunk} \\
\quad \texttt{watered} & actuated spray nozzle currently within the plant's spatial window \\
\quad \texttt{not\_watered} & exact negation of \texttt{watered} \\
\quad \texttt{pressed\_exact} & environment-recorded button count exactly $n$ \\
\quad \texttt{pressed\_after} & exact press count whose recorded events all follow a prerequisite count \\
\quad \texttt{in\_bottle} & object inside a live bottle-relative radial and height window \\
\quad \texttt{not\_in\_bottle} & exact negation of \texttt{in\_bottle} \\
\midrule
\multicolumn{2}{l}{\textit{Counting and memory predicates}} \\
\quad \texttt{swaps\_n} & environment-recorded color swaps at least $n$ \\
\quad \texttt{color\_match\_place} & object placed in a container whose current color matches its remembered color \\
\quad \texttt{placed\_n} & at least $n$ first-entry placement events recorded for a target \\
\quad \texttt{order\_match} & all orderable objects removed in their recorded placement order \\
\end{longtable}
\endgroup

\section{Task Specification and Scoring Semantics}
\label{app:formalism}

\paragraph{Task structure.} Every task specifies a robot embodiment and
scene, a set of placed objects with their spawn regions, a natural-language
instruction, a goal composed of typed predicates, and a list of scored
subtasks. Each spawn region defines a center, an $xy$-extent, and a
yaw range; at reset, every assigned object is placed at a pose sampled
uniformly from its region, so initial conditions differ across episodes
even within one seed family. A surface-contact flag places objects
resting on the support surface rather than dropping them.

\paragraph{Subtask latching and scoring.} At each control step all
predicates are evaluated, and subtask $k$ latches the first time all of
its predicates hold simultaneously. Let $\ell_k(T)$ denote this latched
state, $m_k$ indicate that the subtask has a must-hold predicate,
$h_k(T)$ denote its terminal truth, and $v_k(T)$ indicate a latched
violation. Its effective completion is
$e_k(T)=\ell_k(T)(1-m_k+m_kh_k(T))(1-v_k(T))$. Strict success is
$\prod_k e_k(T)$ and the progress score is $\sum_k w_k e_k(T)$, where
$w_k\geq 0$ and $\sum_k w_k=100$. Zero-weight safety or terminal
constraints still participate in strict success. For ordered tasks, the
evaluator permits a subtask to latch only after the required earlier
subtasks or groups have latched, enforcing the intended staging.

\paragraph{Language schedules.} Instruction-Following \emph{switch}
tasks declare an ordered schedule of instruction segments. Each nonfinal
segment specifies an \texttt{until\_step} control-step boundary, while the
last segment supplies the instruction for the remainder of the episode.
The policy observation receives the active segment text. Scoring continues
to use the task's declared predicate set, with ordered subtasks enforcing the
intended sequence.

\paragraph{Episode horizon.} A task-level \texttt{max\_steps} value takes
precedence when present. Otherwise, the evaluation horizon is
$1.2\times$ the 90th-percentile demonstration length in control steps,
with a 30\,s fallback when no demonstration lengths are available. Timeouts
retain the latched partial score.

\section{Out-of-Domain Transformation Rules}
\label{app:ood}

Each in-domain task is paired with a single out-of-domain (OOD) variant
that perturbs the scene but keeps the goal unchanged, so the OOD score
isolates robustness rather than task mastery. Transformations are
deterministic and reproducible across runs and machines through stable,
component-scoped keys. Six transformation families are used:

\begin{enumerate}[leftmargin=*]
    \item \textbf{Scene randomization} (all tasks): floor, wall, and
    tabletop textures resampled from a held-out pool; light position and
    diffuse intensity jittered; shadow casting toggled.
    \item \textbf{Camera variation} (all tasks): head and wrist camera
    extrinsics selected from a validated finite pose bank for each evaluation
    key.
    \item \textbf{Layout variation} (all tasks): task-specific placement
    and geometry changes are recorded in the committed OOD YAML.
    \item \textbf{Distractor injection} (45 compatible tasks): three to
    nine objects are placed in collision-safe tabletop regions; grasping
    a distractor terminates the episode at its current score, so OOD
    scores penalize both distraction and mis-grasping.
    \item \textbf{Object rescaling} (ten articulated
    Limited-Demonstrations tasks): each OOD YAML selects a smaller or larger
    regime anchored by a factor in $\{0.85,0.9,1.1,1.15\}$. A deterministic
    episode factor is sampled within $\pm0.18$ of the anchor and clipped to
    $[0.65,0.96]$ for smaller regimes or $[1.04,1.35]$ for larger regimes.
    \item \textbf{Left-right mirroring} (nine tasks): task geometry is
    mirrored about the workspace midline to exchange arm-side demands.
\end{enumerate}

At evaluation time the OOD variant is loaded in place of the base task.
The committed YAML fixes the layout and applicable transformations, while
appearance, camera, articulated scale, and distractor placement are derived
reproducibly from the task, seed, and episode. The goal, episode count,
horizon, and scoring rules are shared across domains. In-domain evaluation
samples a recorded working robot pose and restores only the robot joints,
while retaining fresh base-YAML object placement. OOD evaluation uses the
default robot reset pose with the committed OOD placement. Each domain is
deterministic and identical across models.

\section{Dataset Structure and Quality Control}
\label{app:dataset}

\paragraph{Layout.} The dataset is released in LeRobot v2.1 and v3.0
layouts~\cite{cadene2024lerobot}, one directory per task, storing
per-episode state/action trajectories and six synchronized camera
videos. Table~\ref{tab:schema} summarizes the observation and action
modalities available to the training and evaluation interfaces.

\begin{table}[h]
\caption{Observation and action modalities per frame (25\,Hz). The six
video streams are RGB and depth from the head and the two wrist cameras.}
\label{tab:schema}
\centering
\footnotesize
\begin{tabular}{llc}
\toprule
\textbf{Modality} & \textbf{Content} & \textbf{Dim} \\
\midrule
\multicolumn{3}{l}{\textit{Observations}} \\
\quad Arm end-effector pose (right, left) & xyz + quaternion, world or head-camera frame & 14 \\
\quad Arm joint position / velocity & 7 per arm & 14 each \\
\quad Hand joint position / velocity & 22 per hand & 44 each \\
\midrule
\multicolumn{3}{l}{\textit{Actions (one 58-dimensional control representation)}} \\
\quad Arm component & joint targets or end-effector pose in world/head-camera frame & 14 \\
\quad Hand joint targets & 22 per hand & 44 \\
\midrule
\multicolumn{3}{l}{\textit{Video streams (H.264, 25\,fps)}} \\
\quad Head / left-wrist / right-wrist RGB & & $360{\times}480$ \\
\quad Head / left-wrist / right-wrist depth & 3-channel & $360{\times}480$ \\
\bottomrule
\end{tabular}
\end{table}

\paragraph{Quality control.} Collection proceeds as teleoperated
sessions with an in-loop review interface: each episode is recorded,
replayed, and accepted or rejected by the operator. Accepted episodes
are then re-validated automatically by replaying them from their
serialized initial states and re-checking the task's goal predicates;
only episodes that re-achieve full subtask completion are exported. The
released corpus is therefore success-only by construction.

\paragraph{Normalization statistics.} Per-task and per-mixture
normalization statistics (mean/std and 1st/99th percentiles of every
action and observation channel) are distributed with the dataset and
used by all baselines (Appendix~\ref{app:training}).

\section{Per-Model Training Configurations}
\label{app:training}

All seven baselines are fine-tuned from their official pretrained
checkpoints on the identical all-category BiDex-100 mixture (58-DoF
absolute joint targets; head + left-wrist + right-wrist color streams)
using the checkpoint selected for the main evaluation: 60k steps for
$\pi_{0.5}$ and LingBot-VLA-v2, 120k for OpenDM-DM05, and 400k for
GR00T-N1.7, InternVLA-A1.5, Galaxea G0.5, and Xiaomi XR1.
Table~\ref{tab:hyper} summarizes the optimizer settings; model-specific
notes follow.

\begin{table*}[t]
\caption{Training hyperparameters and selected checkpoints of the seven evaluated VLAs (cosine-family schedules, grad-clip 1.0). GBS = global batch size.}
\label{tab:hyper}
\centering
\footnotesize
\setlength{\tabcolsep}{4pt}
\begin{tabular}{lllllccccc}
\toprule
\textbf{Model} & \textbf{Train} & \textbf{Optimizer} & \textbf{Peak LR} & \textbf{Warmup} & \textbf{Min LR} & \textbf{GBS} & \textbf{Dtype} & \textbf{Chunk} & \textbf{Frozen modules} \\
\midrule
$\pi_{0.5}$      & 60k  & AdamW (optax) & 5e-5  & 100   & 5e-5  & 256 & bf16 & 50 & none (EMA 0.999) \\
GR00T-N1.7       & 400k & AdamW         & 1e-4  & 1500  & $\sim$0 & 256 & bf16 & 16 & VLM + vision encoder \\
InternVLA-A1.5   & 400k & AdamW         & 5e-5  & 2000  & 5e-6  & 256 & bf16 & 50 & none \\
Galaxea G0.5     & 400k & AdamW         & 1e-5  & 1500  & 1e-6  & 256 & bf16 & 32 & none \\
OpenDM-DM05      & 120k & AdamW         & 2.5e-5 & 1000 & 2.5e-6 & 256 & bf16 & 50 & VLM token embeddings \\
LingBot-VLA-v2   & 60k  & Muon (+AdamW) & 1e-4  & 0     & 5e-5  & 256 & fp32 & 50 & none \\
Xiaomi XR1       & 400k & FusedAdam     & 2e-5  & 500   & 5e-6  & 128 & bf16 & 30 & VLM input embeddings \\
\bottomrule
\end{tabular}
\end{table*}

\paragraph{$\pi_{0.5}$ (openpi).} PaliGemma-2B VLM with a 300M-parameter
flow-matching action expert; JAX/pmap training in bf16. Uses \emph{relative}
(delta) 58-DoF actions; action horizon 50 with 10 denoising steps at
inference (single training noise step, $t\sim\mathrm{Beta}(1.5,1)$). The
pretrained 32-DoF action projections are resized to 58 DoF by re-initializing
mismatched layers. Images resized to $224{\times}224$.

\paragraph{GR00T-N1.7.} Cosmos-Reason2-2B VLM (Qwen3-VL architecture) with a
16-layer Alternate-VL DiT flow-matching expert (hidden 1024). The VLM and
vision tower are frozen; only the DiT, state/action projectors, and the VLLN
adapter train (DeepSpeed ZeRO-2). Absolute 58-DoF actions; chunk 16, 4
inference denoising steps; color-jitter augmentation; min/max percentile
normalization.

\paragraph{InternVLA-A1.5.} Qwen3.5-2B with a 1B-parameter action expert;
flow-matching MSE weighted $\times 10$. The pretrained 32-DoF action
projection is trimmed and rebuilt at 58 DoF. Chunk 50, 10 denoising steps.

\paragraph{Galaxea G0.5.} Qwen3.5-2B (mrope + FLA linear attention) with a
flow-matching action expert (1024-wide, adaLN). Absolute 58-DoF actions; chunk
32, single-step FM Euler at inference. The base 27-DoF action projections are
partially copied and zero-padded to 58 DoF; the RVQ action tokenizer is
bypassed in flow-matching mode. Learning rate set to $1/8$ of the model's
default (1e-5) with a warmup-constant-cosine schedule.

\paragraph{OpenDM-DM05.} Gemma3-4B-IT + SigLIP-SO400M ($448{\times}448$, 256
visual tokens per image) with an independent Gemma3-1B action expert
(rectified flow). Absolute actions padded from 32 to 64 slots (58 active);
chunk 50, 10 denoising steps. VLM token embeddings frozen; eager attention;
Liger kernels; gradient checkpointing.

\paragraph{LingBot-VLA-v2.} Qwen3-VL-4B with a 36-layer Mixture-of-Experts
flow-matching expert (32 experts, top-4) trained with an L1-FM loss. Muon on
2D+ weights (routed-expert LR $\times\sqrt{32/4}$) with AdamW for 1D/embeddings;
full fp32 training. Absolute 58-DoF actions (55-DoF pretrained projection
resized); chunk 50, 10 denoising steps; alignment/depth auxiliary heads
disabled.

\paragraph{Xiaomi XR1.} Qwen3-VL-4B with a 36-layer 60-slot DiT expert
(AdaLN, cross-attention to VLM key-values); the 58 active channels occupy the
right-aligned slots. FusedAdam with a LambdaLR cosine schedule; chunk 30, 5
denoising steps; training repeats each noise sample 4$\times$. VLM input
embeddings frozen; vision encoder and language FFNs train with gradient
checkpointing; DeepSpeed.

\section{Evaluation Protocol}
\label{app:eval}

\paragraph{Protocol.} Each (model, task, domain) triple is evaluated over 2
fixed seeds $\times$ 25 episodes, using ID seeds 0 and 1 and OOD seeds 2 and
3. Object placements are sampled from the applicable task regions with a seed
that is a deterministic function of the task, seed index, and episode index.
The ID protocol additionally restores a deterministically selected working
robot pose, while OOD uses the default robot reset pose. This procedure makes
every rollout reproducible within its domain. Episodes terminate on full success, on timeout
(explicit task \texttt{max\_steps}, otherwise $1.2\times$ the P90 demonstration length), or, in OOD scenes with
distractors, on grasping a distractor object. The vectorized harness
co-locates policy inference and simulation with a single shared renderer per
worker.

\paragraph{Metrics.} Per episode we record strict success, the progress score
(as defined in the main paper), the per-subtask latch times, and the
time-to-first-failure. Aggregation reports, per capability axis and domain:
mean strict success (\emph{SR}) followed by mean progress score (\emph{Score}); the
overall number is the equal-weight mean over all 212 task-domain means. Per-task breakdowns and the
full rollout videos are retained for audit, and the leaderboard requires no
human labeling.

\section{Task Catalog}
\label{app:tasks}

This section documents every one of the 106 task specifications, grouped by
capability axis. Each axis opens with a short description of the capability it
stresses, followed by one entry per task. Each entry shows a representative
head-camera image of the task scene, the task identifier, the number of
scored subtasks $K$, the number of released demonstrations, the language
instruction, and the full scoring criteria, namely the subtask names together with
their integer score weights, which sum to 100 per task. Ordered subtask
chains are joined by ``$\to$'' and unordered subtasks by ``$\cdot$''. A
subtask marked $^{\dagger}$ carries \emph{must-hold} predicates that must
still be true at episode end; $^{\ddagger}$ marks \emph{latch-violation}
predicates that permanently fail the episode if the forbidden state ever
occurs (Appendix~\ref{app:formalism}). Ten Instruction-Following shelf-transfer
switch and combination specifications contain 30 demonstrations each, while
the other ten Instruction-Following specifications contain 100 demonstrations
each.

\subsection{Generalization (40 specifications, 40 with released demonstrations)}
\label{app:tasks:generalization}

Single- and few-stage skills, including lifting, placement, insertion, threading, and cooking, over a diverse object library spanning food items, tableware, containers, and tools. The ten \texttt{kitchen\_*} tasks execute on the Astribot~S1 humanoid morphology and the remaining 30 use the dual FR3v2 platform; all 40 are included in the reported Generalization results. In-domain resets already randomize object poses, so the axis measures transfer across object geometry and embodiment.

\par\vspace{3pt}\noindent
{\renewcommand{\arraystretch}{1.0}%
\begin{tabular}{@{}>{\columncolor{catgen}}p{2.2pt}@{\hspace{5pt}}m{2.1cm}@{\hspace{6pt}}p{\dimexpr\linewidth-2.2pt-5pt-2.1cm-6pt\relax}@{}}
& \raisebox{-5pt}{\includegraphics[width=2.1cm]{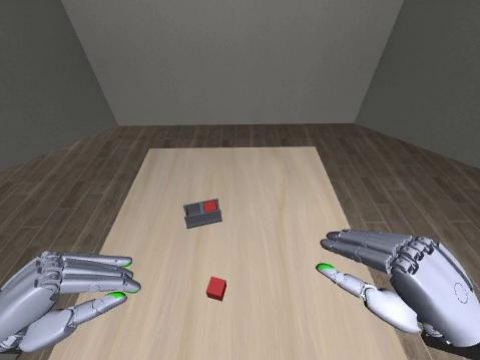}} &
\textbf{1.~\texttt{insert\_\allowbreak two\_\allowbreak cubes\_\allowbreak into\_\allowbreak slot}}\hfill {\footnotesize $K{=}2$ subtasks $\cdot$ 100 demos}\newline
{\small\emph{Instruction:} Insert two cubes into the slot.}\newline
{\footnotesize\textbf{Scoring:} Insert cube A into the slot \textbf{(50)}$^{\dagger}$ $\cdot$ Insert cube B into the slot \textbf{(50)}$^{\dagger}$.}
\end{tabular}}

\par\vspace{3pt}\noindent
{\renewcommand{\arraystretch}{1.0}%
\begin{tabular}{@{}>{\columncolor{catgen}}p{2.2pt}@{\hspace{5pt}}m{2.1cm}@{\hspace{6pt}}p{\dimexpr\linewidth-2.2pt-5pt-2.1cm-6pt\relax}@{}}
& \raisebox{-5pt}{\includegraphics[width=2.1cm]{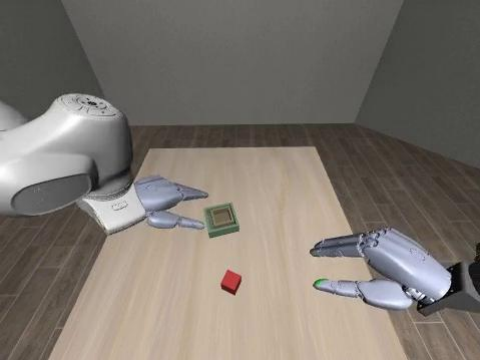}} &
\textbf{2.~\texttt{nest\_\allowbreak medium\_\allowbreak box\_\allowbreak in\_\allowbreak large\_\allowbreak box}}\hfill {\footnotesize $K{=}4$ subtasks $\cdot$ 100 demos}\newline
{\small\emph{Instruction:} Place the medium box inside the large box, then place the cube inside the medium box.}\newline
{\footnotesize\textbf{Scoring:} Pick up the medium box \textbf{(10)} $\cdot$ Nest the medium box inside the large box \textbf{(40)}$^{\dagger}$ $\cdot$ Pick up the cube \textbf{(10)} $\cdot$ Place the cube inside the medium box \textbf{(40)}$^{\dagger}$.}
\end{tabular}}

\par\vspace{3pt}\noindent
{\renewcommand{\arraystretch}{1.0}%
\begin{tabular}{@{}>{\columncolor{catgen}}p{2.2pt}@{\hspace{5pt}}m{2.1cm}@{\hspace{6pt}}p{\dimexpr\linewidth-2.2pt-5pt-2.1cm-6pt\relax}@{}}
& \raisebox{-5pt}{\includegraphics[width=2.1cm]{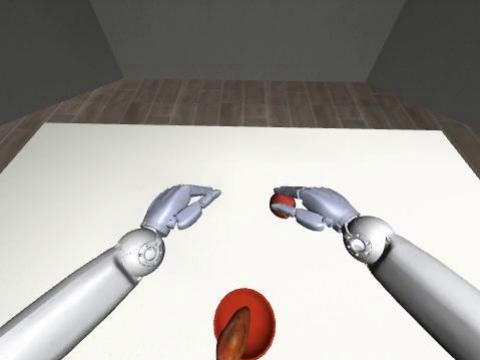}} &
\textbf{3.~\texttt{put\_\allowbreak sweet\_\allowbreak potato\_\allowbreak and\_\allowbreak onion\_\allowbreak in\_\allowbreak bowl}}\hfill {\footnotesize $K{=}3$ subtasks $\cdot$ 100 demos}\newline
{\small\emph{Instruction:} Put the sweet potato and the onion in the bowl.}\newline
{\footnotesize\textbf{Scoring:} Pick up the sweet potato \textbf{(25)} $\cdot$ Put the sweet potato in the bowl \textbf{(25)}$^{\dagger}$ $\cdot$ Put the onion in the bowl \textbf{(50)}$^{\dagger}$.}
\end{tabular}}

\par\vspace{3pt}\noindent
{\renewcommand{\arraystretch}{1.0}%
\begin{tabular}{@{}>{\columncolor{catgen}}p{2.2pt}@{\hspace{5pt}}m{2.1cm}@{\hspace{6pt}}p{\dimexpr\linewidth-2.2pt-5pt-2.1cm-6pt\relax}@{}}
& \raisebox{-5pt}{\includegraphics[width=2.1cm]{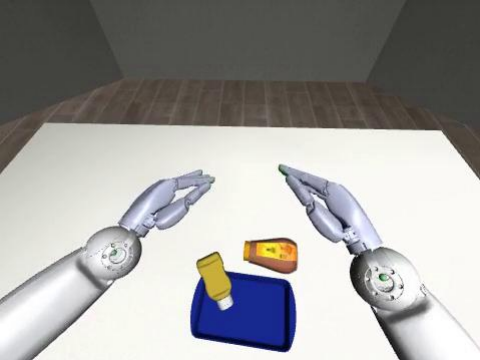}} &
\textbf{4.~\texttt{put\_\allowbreak mustard\_\allowbreak and\_\allowbreak honey\_\allowbreak bottle\_\allowbreak on\_\allowbreak tray}}\hfill {\footnotesize $K{=}3$ subtasks $\cdot$ 100 demos}\newline
{\small\emph{Instruction:} Put the mustard and the honey bottle on the tray.}\newline
{\footnotesize\textbf{Scoring:} Pick up the mustard \textbf{(25)} $\cdot$ Place the mustard on the tray \textbf{(25)}$^{\dagger}$ $\cdot$ Place the honey bottle on the tray \textbf{(50)}$^{\dagger}$.}
\end{tabular}}

\par\vspace{3pt}\noindent
{\renewcommand{\arraystretch}{1.0}%
\begin{tabular}{@{}>{\columncolor{catgen}}p{2.2pt}@{\hspace{5pt}}m{2.1cm}@{\hspace{6pt}}p{\dimexpr\linewidth-2.2pt-5pt-2.1cm-6pt\relax}@{}}
& \raisebox{-5pt}{\includegraphics[width=2.1cm]{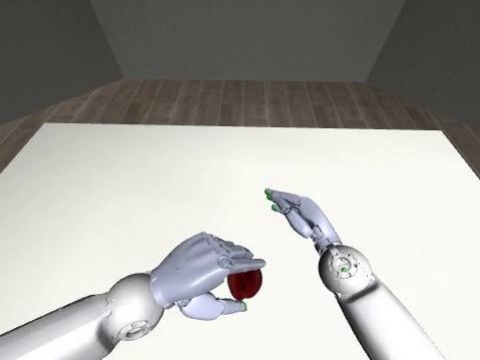}} &
\textbf{5.~\texttt{pick\_\allowbreak up\_\allowbreak peach}}\hfill {\footnotesize $K{=}1$ subtask $\cdot$ 100 demos}\newline
{\small\emph{Instruction:} Pick up the peach.}\newline
{\footnotesize\textbf{Scoring:} Pick up the peach \textbf{(100)}.}
\end{tabular}}

\par\vspace{3pt}\noindent
{\renewcommand{\arraystretch}{1.0}%
\begin{tabular}{@{}>{\columncolor{catgen}}p{2.2pt}@{\hspace{5pt}}m{2.1cm}@{\hspace{6pt}}p{\dimexpr\linewidth-2.2pt-5pt-2.1cm-6pt\relax}@{}}
& \raisebox{-5pt}{\includegraphics[width=2.1cm]{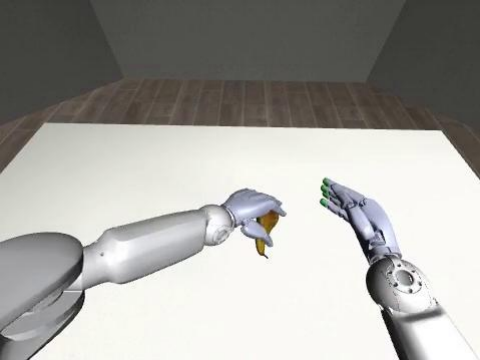}} &
\textbf{6.~\texttt{pick\_\allowbreak up\_\allowbreak pear}}\hfill {\footnotesize $K{=}1$ subtask $\cdot$ 100 demos}\newline
{\small\emph{Instruction:} Pick up the pear.}\newline
{\footnotesize\textbf{Scoring:} Pick up the pear \textbf{(100)}.}
\end{tabular}}

\par\vspace{3pt}\noindent
{\renewcommand{\arraystretch}{1.0}%
\begin{tabular}{@{}>{\columncolor{catgen}}p{2.2pt}@{\hspace{5pt}}m{2.1cm}@{\hspace{6pt}}p{\dimexpr\linewidth-2.2pt-5pt-2.1cm-6pt\relax}@{}}
& \raisebox{-5pt}{\includegraphics[width=2.1cm]{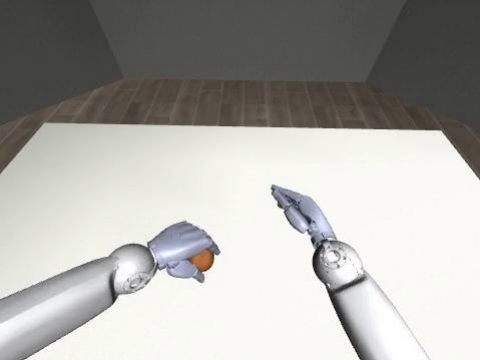}} &
\textbf{7.~\texttt{pick\_\allowbreak up\_\allowbreak tangerine}}\hfill {\footnotesize $K{=}1$ subtask $\cdot$ 100 demos}\newline
{\small\emph{Instruction:} Pick up the tangerine.}\newline
{\footnotesize\textbf{Scoring:} Pick up the tangerine \textbf{(100)}.}
\end{tabular}}

\par\vspace{3pt}\noindent
{\renewcommand{\arraystretch}{1.0}%
\begin{tabular}{@{}>{\columncolor{catgen}}p{2.2pt}@{\hspace{5pt}}m{2.1cm}@{\hspace{6pt}}p{\dimexpr\linewidth-2.2pt-5pt-2.1cm-6pt\relax}@{}}
& \raisebox{-5pt}{\includegraphics[width=2.1cm]{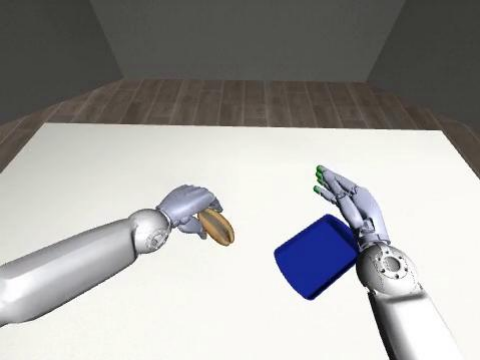}} &
\textbf{8.~\texttt{put\_\allowbreak hotdog\_\allowbreak bun\_\allowbreak on\_\allowbreak tray}}\hfill {\footnotesize $K{=}2$ subtasks $\cdot$ 100 demos}\newline
{\small\emph{Instruction:} Pick up the hotdog bun and place it on the tray.}\newline
{\footnotesize\textbf{Scoring:} Pick up the hotdog\_bun \textbf{(50)} $\cdot$ Place the hotdog\_bun on the tray \textbf{(50)}$^{\dagger}$.}
\end{tabular}}

\par\vspace{3pt}\noindent
{\renewcommand{\arraystretch}{1.0}%
\begin{tabular}{@{}>{\columncolor{catgen}}p{2.2pt}@{\hspace{5pt}}m{2.1cm}@{\hspace{6pt}}p{\dimexpr\linewidth-2.2pt-5pt-2.1cm-6pt\relax}@{}}
& \raisebox{-5pt}{\includegraphics[width=2.1cm]{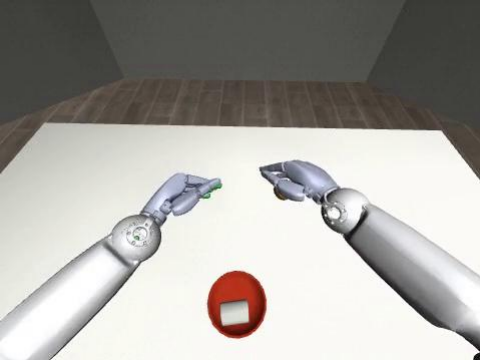}} &
\textbf{9.~\texttt{put\_\allowbreak marshmallow\_\allowbreak and\_\allowbreak cookie\_\allowbreak dough\_\allowbreak ball\_\allowbreak in\_\allowbreak bowl}}\hfill {\footnotesize $K{=}3$ subtasks $\cdot$ 100 demos}\newline
{\small\emph{Instruction:} Put the marshmallow and the cookie dough ball in the bowl.}\newline
{\footnotesize\textbf{Scoring:} Pick up the marshmallow \textbf{(25)} $\cdot$ Put the marshmallow in the bowl \textbf{(25)}$^{\dagger}$ $\cdot$ Put the cookie dough ball in the bowl \textbf{(50)}$^{\dagger}$.}
\end{tabular}}

\par\vspace{3pt}\noindent
{\renewcommand{\arraystretch}{1.0}%
\begin{tabular}{@{}>{\columncolor{catgen}}p{2.2pt}@{\hspace{5pt}}m{2.1cm}@{\hspace{6pt}}p{\dimexpr\linewidth-2.2pt-5pt-2.1cm-6pt\relax}@{}}
& \raisebox{-5pt}{\includegraphics[width=2.1cm]{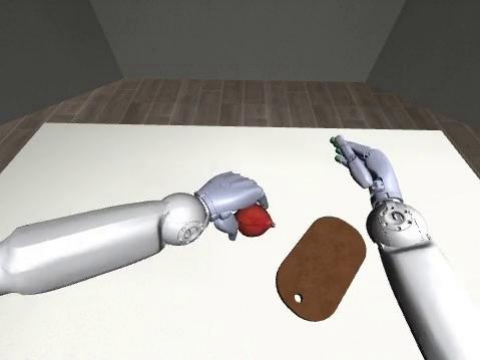}} &
\textbf{10.~\texttt{put\_\allowbreak onion\_\allowbreak on\_\allowbreak cutting\_\allowbreak board}}\hfill {\footnotesize $K{=}2$ subtasks $\cdot$ 100 demos}\newline
{\small\emph{Instruction:} Pick up the onion and place it on the cutting board.}\newline
{\footnotesize\textbf{Scoring:} Pick up the onion \textbf{(50)} $\cdot$ Place the onion on the cutting\_board \textbf{(50)}$^{\dagger}$.}
\end{tabular}}

\par\vspace{3pt}\noindent
{\renewcommand{\arraystretch}{1.0}%
\begin{tabular}{@{}>{\columncolor{catgen}}p{2.2pt}@{\hspace{5pt}}m{2.1cm}@{\hspace{6pt}}p{\dimexpr\linewidth-2.2pt-5pt-2.1cm-6pt\relax}@{}}
& \raisebox{-5pt}{\includegraphics[width=2.1cm]{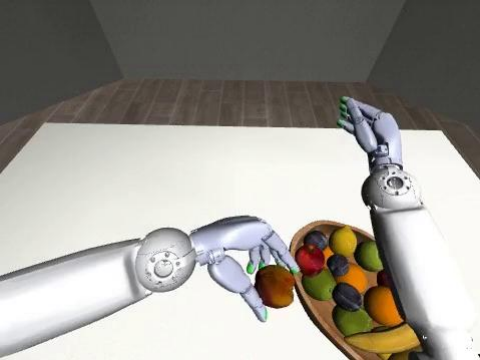}} &
\textbf{11.~\texttt{put\_\allowbreak peach\_\allowbreak in\_\allowbreak fruit\_\allowbreak bowl}}\hfill {\footnotesize $K{=}2$ subtasks $\cdot$ 100 demos}\newline
{\small\emph{Instruction:} Pick up the peach and put it in the fruit bowl.}\newline
{\footnotesize\textbf{Scoring:} Pick up the peach \textbf{(50)} $\cdot$ Place the peach in the fruit\_bowl \textbf{(50)}$^{\dagger}$.}
\end{tabular}}

\par\vspace{3pt}\noindent
{\renewcommand{\arraystretch}{1.0}%
\begin{tabular}{@{}>{\columncolor{catgen}}p{2.2pt}@{\hspace{5pt}}m{2.1cm}@{\hspace{6pt}}p{\dimexpr\linewidth-2.2pt-5pt-2.1cm-6pt\relax}@{}}
& \raisebox{-5pt}{\includegraphics[width=2.1cm]{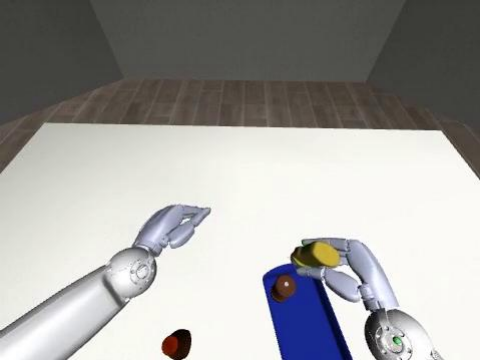}} &
\textbf{12.~\texttt{place\_\allowbreak condiments\_\allowbreak on\_\allowbreak tray}}\hfill {\footnotesize $K{=}4$ subtasks $\cdot$ 100 demos}\newline
{\small\emph{Instruction:} Place the salt and pepper shaker, mustard, and bbq sauce on the tray.}\newline
{\footnotesize\textbf{Scoring:} Pick up the salt and pepper shaker \textbf{(20)} $\cdot$ Place the salt and pepper shaker on the tray \textbf{(20)}$^{\dagger}$ $\cdot$ Place the mustard on the tray \textbf{(30)}$^{\dagger}$ $\cdot$ Place the bbq sauce on the tray \textbf{(30)}$^{\dagger}$.}
\end{tabular}}

\par\vspace{3pt}\noindent
{\renewcommand{\arraystretch}{1.0}%
\begin{tabular}{@{}>{\columncolor{catgen}}p{2.2pt}@{\hspace{5pt}}m{2.1cm}@{\hspace{6pt}}p{\dimexpr\linewidth-2.2pt-5pt-2.1cm-6pt\relax}@{}}
& \raisebox{-5pt}{\includegraphics[width=2.1cm]{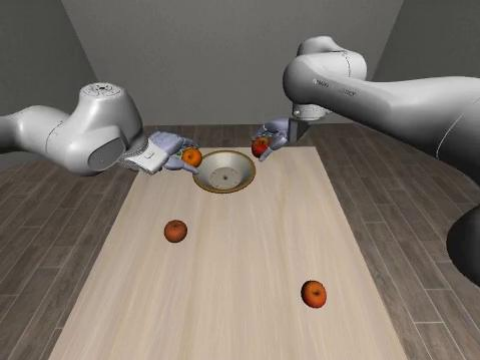}} &
\textbf{13.~\texttt{put\_\allowbreak apple\_\allowbreak and\_\allowbreak orange\_\allowbreak in\_\allowbreak bowl}}\hfill {\footnotesize $K{=}6$ subtasks $\cdot$ 100 demos}\newline
{\small\emph{Instruction:} Put one apple and one orange into the bowl, and leave one apple and one orange untouched on the table.}\newline
{\footnotesize\textbf{Scoring:} Pick up the first apple \textbf{(20)} $\cdot$ Place the first apple into the bowl \textbf{(30)}$^{\dagger}$ $\cdot$ Pick up the second orange \textbf{(20)} $\cdot$ Place the second orange into the bowl \textbf{(30)}$^{\dagger}$ $\cdot$ Leave the first orange untouched on the table \textbf{(0)}$^{\ddagger}$ $\cdot$ Leave the second apple untouched on the table \textbf{(0)}$^{\ddagger}$.}
\end{tabular}}

\par\vspace{3pt}\noindent
{\renewcommand{\arraystretch}{1.0}%
\begin{tabular}{@{}>{\columncolor{catgen}}p{2.2pt}@{\hspace{5pt}}m{2.1cm}@{\hspace{6pt}}p{\dimexpr\linewidth-2.2pt-5pt-2.1cm-6pt\relax}@{}}
& \raisebox{-5pt}{\includegraphics[width=2.1cm]{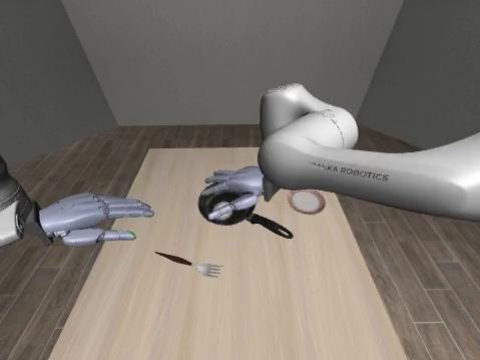}} &
\textbf{14.~\texttt{cook\_\allowbreak egg\_\allowbreak in\_\allowbreak frypan}}\hfill {\footnotesize $K{=}5$ subtasks $\cdot$ 100 demos}\newline
{\small\emph{Instruction:} Pick up the frypan, then put the egg onto the frypan, and finally place the fork on the plate.}\newline
{\footnotesize\textbf{Scoring:} Pick up the frypan \textbf{(20)} $\cdot$ Pick up the egg \textbf{(15)} $\cdot$ Put the egg on the frypan \textbf{(15)}$^{\dagger}$ $\cdot$ Pick up the fork \textbf{(30)} $\cdot$ Place the fork on the plate \textbf{(20)}$^{\dagger}$.}
\end{tabular}}

\par\vspace{3pt}\noindent
{\renewcommand{\arraystretch}{1.0}%
\begin{tabular}{@{}>{\columncolor{catgen}}p{2.2pt}@{\hspace{5pt}}m{2.1cm}@{\hspace{6pt}}p{\dimexpr\linewidth-2.2pt-5pt-2.1cm-6pt\relax}@{}}
& \raisebox{-5pt}{\includegraphics[width=2.1cm]{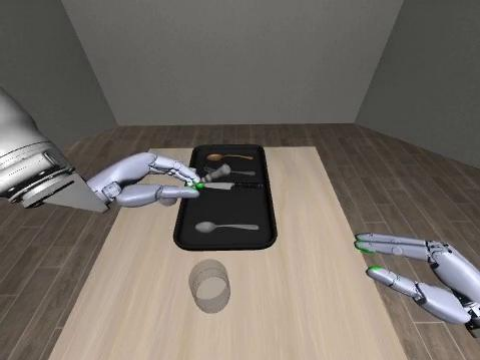}} &
\textbf{15.~\texttt{pick\_\allowbreak up\_\allowbreak spoon\_\allowbreak from\_\allowbreak tray\_\allowbreak into\_\allowbreak cup}}\hfill {\footnotesize $K{=}2$ subtasks $\cdot$ 100 demos}\newline
{\small\emph{Instruction:} From the tray of mixed utensils, pick up the one target spoon and drop it into the cup.}\newline
{\footnotesize\textbf{Scoring:} Pick up the target utensil and drop it into the cup \textbf{(100)}$^{\dagger}$ $\cdot$ Another utensil is not picked up \textbf{(0)}$^{\ddagger}$.}
\end{tabular}}

\par\vspace{3pt}\noindent
{\renewcommand{\arraystretch}{1.0}%
\begin{tabular}{@{}>{\columncolor{catgen}}p{2.2pt}@{\hspace{5pt}}m{2.1cm}@{\hspace{6pt}}p{\dimexpr\linewidth-2.2pt-5pt-2.1cm-6pt\relax}@{}}
& \raisebox{-5pt}{\includegraphics[width=2.1cm]{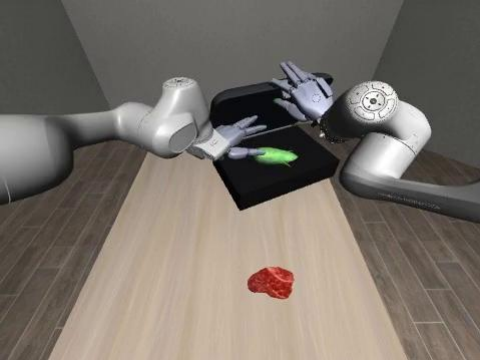}} &
\textbf{16.~\texttt{grill\_\allowbreak steak\_\allowbreak and\_\allowbreak corn}}\hfill {\footnotesize $K{=}6$ subtasks $\cdot$ 100 demos}\newline
{\small\emph{Instruction:} Open the grill lid, put the steak and the corn on the grill, then close the grill lid.}\newline
{\footnotesize\textbf{Scoring:} Open the grill lid \textbf{(20)} $\to$ Pick up the steak \textbf{(20)} $\to$ Put the steak on the grill \textbf{(15)}$^{\dagger}$ $\to$ Pick up the corn \textbf{(15)} $\to$ Put the corn on the grill \textbf{(15)}$^{\dagger}$ $\to$ Close the grill lid \textbf{(15)}$^{\dagger}$.}
\end{tabular}}

\par\vspace{3pt}\noindent
{\renewcommand{\arraystretch}{1.0}%
\begin{tabular}{@{}>{\columncolor{catgen}}p{2.2pt}@{\hspace{5pt}}m{2.1cm}@{\hspace{6pt}}p{\dimexpr\linewidth-2.2pt-5pt-2.1cm-6pt\relax}@{}}
& \raisebox{-5pt}{\includegraphics[width=2.1cm]{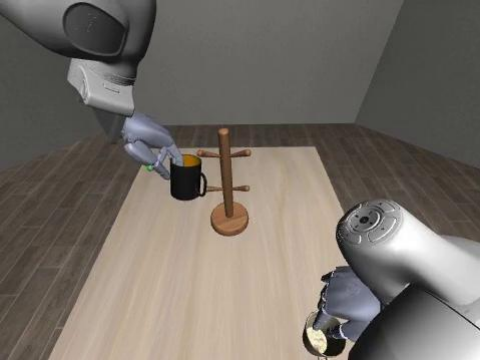}} &
\textbf{17.~\texttt{hang\_\allowbreak mugs\_\allowbreak on\_\allowbreak tree}}\hfill {\footnotesize $K{=}4$ subtasks $\cdot$ 100 demos}\newline
{\small\emph{Instruction:} Pick up each of the two mugs and hang them both on the wooden mug tree.}\newline
{\footnotesize\textbf{Scoring:} Pick up the first mug \textbf{(10)} $\cdot$ Hang the first mug on the mug tree \textbf{(40)}$^{\dagger}$ $\cdot$ Pick up the second mug \textbf{(10)} $\cdot$ Hang the second mug on the mug tree \textbf{(40)}$^{\dagger}$.}
\end{tabular}}

\par\vspace{3pt}\noindent
{\renewcommand{\arraystretch}{1.0}%
\begin{tabular}{@{}>{\columncolor{catgen}}p{2.2pt}@{\hspace{5pt}}m{2.1cm}@{\hspace{6pt}}p{\dimexpr\linewidth-2.2pt-5pt-2.1cm-6pt\relax}@{}}
& \raisebox{-5pt}{\includegraphics[width=2.1cm]{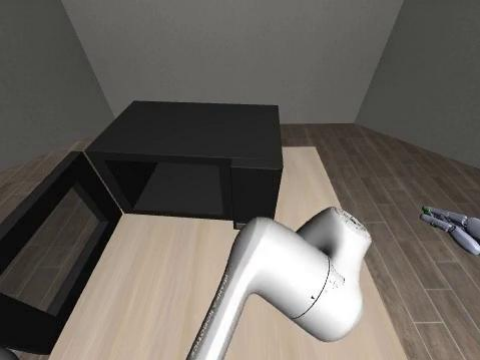}} &
\textbf{18.~\texttt{heat\_\allowbreak sandwich\_\allowbreak bread\_\allowbreak in\_\allowbreak microwave}}\hfill {\footnotesize $K{=}4$ subtasks $\cdot$ 100 demos}\newline
{\small\emph{Instruction:} Open the microwave, put the sandwich bread inside, and close the microwave.}\newline
{\footnotesize\textbf{Scoring:} Open the microwave door \textbf{(30)} $\to$ Pick up the sandwich bread \textbf{(20)} $\to$ Put the sandwich bread inside the microwave \textbf{(20)} $\to$ Close the microwave door \textbf{(30)}.}
\end{tabular}}

\par\vspace{3pt}\noindent
{\renewcommand{\arraystretch}{1.0}%
\begin{tabular}{@{}>{\columncolor{catgen}}p{2.2pt}@{\hspace{5pt}}m{2.1cm}@{\hspace{6pt}}p{\dimexpr\linewidth-2.2pt-5pt-2.1cm-6pt\relax}@{}}
& \raisebox{-5pt}{\includegraphics[width=2.1cm]{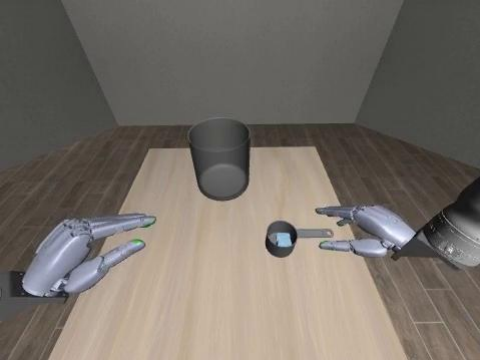}} &
\textbf{19.~\texttt{put\_\allowbreak measuring\_\allowbreak cup\_\allowbreak and\_\allowbreak ice\_\allowbreak into\_\allowbreak glass}}\hfill {\footnotesize $K{=}3$ subtasks $\cdot$ 100 demos}\newline
{\small\emph{Instruction:} Place the ice cube in the measuring cup, then place the measuring cup and ice into the glass.}\newline
{\footnotesize\textbf{Scoring:} Pick up the measuring cup \textbf{(25)} $\to$ Pick up an ice cube \textbf{(25)} $\to$ Place the measuring cup and ice into the glass \textbf{(50)}$^{\dagger}$.}
\end{tabular}}

\par\vspace{3pt}\noindent
{\renewcommand{\arraystretch}{1.0}%
\begin{tabular}{@{}>{\columncolor{catgen}}p{2.2pt}@{\hspace{5pt}}m{2.1cm}@{\hspace{6pt}}p{\dimexpr\linewidth-2.2pt-5pt-2.1cm-6pt\relax}@{}}
& \raisebox{-5pt}{\includegraphics[width=2.1cm]{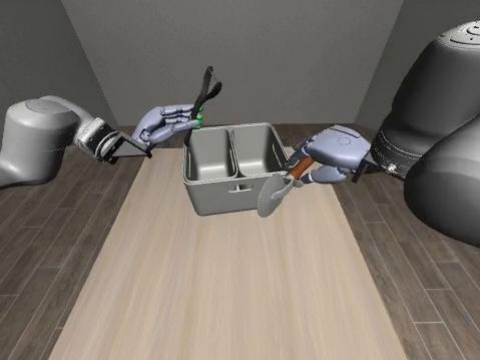}} &
\textbf{20.~\texttt{put\_\allowbreak peeler\_\allowbreak and\_\allowbreak pizza\_\allowbreak cutter\_\allowbreak in\_\allowbreak drainer}}\hfill {\footnotesize $K{=}4$ subtasks $\cdot$ 100 demos}\newline
{\small\emph{Instruction:} Pick up the peeler and the pizza cutter and put them in the drainer.}\newline
{\footnotesize\textbf{Scoring:} Pick up the pizza cutter \textbf{(20)} $\cdot$ Put the pizza cutter in the drainer \textbf{(30)}$^{\dagger}$ $\cdot$ Pick up the peeler \textbf{(20)} $\cdot$ Put the peeler in the drainer \textbf{(30)}$^{\dagger}$.}
\end{tabular}}

\par\vspace{3pt}\noindent
{\renewcommand{\arraystretch}{1.0}%
\begin{tabular}{@{}>{\columncolor{catgen}}p{2.2pt}@{\hspace{5pt}}m{2.1cm}@{\hspace{6pt}}p{\dimexpr\linewidth-2.2pt-5pt-2.1cm-6pt\relax}@{}}
& \raisebox{-5pt}{\includegraphics[width=2.1cm]{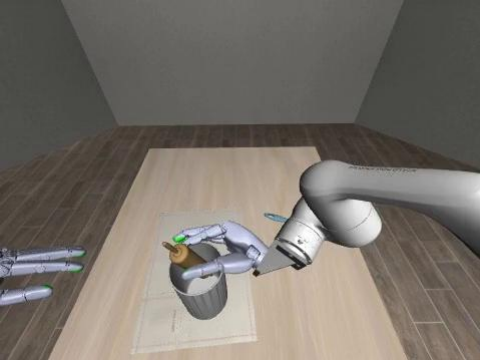}} &
\textbf{21.~\texttt{insert\_\allowbreak spatula\_\allowbreak and\_\allowbreak rolling\_\allowbreak pin\_\allowbreak into\_\allowbreak glass\_\allowbreak cup}}\hfill {\footnotesize $K{=}6$ subtasks $\cdot$ 100 demos}\newline
{\small\emph{Instruction:} Pick up the glass cup, then insert the spatula into the cup, insert the rolling pin into the cup, and place the cup on the placemat.}\newline
{\footnotesize\textbf{Scoring:} Pick up the glass cup \textbf{(15)} $\cdot$ Place the glass cup on the placemat \textbf{(15)}$^{\dagger}$ $\cdot$ Pick up the spatula \textbf{(15)} $\cdot$ Insert the spatula into the glass cup \textbf{(20)}$^{\dagger}$ $\cdot$ Pick up the rolling pin \textbf{(15)} $\cdot$ Insert the rolling pin into the glass cup \textbf{(20)}$^{\dagger}$.}
\end{tabular}}

\par\vspace{3pt}\noindent
{\renewcommand{\arraystretch}{1.0}%
\begin{tabular}{@{}>{\columncolor{catgen}}p{2.2pt}@{\hspace{5pt}}m{2.1cm}@{\hspace{6pt}}p{\dimexpr\linewidth-2.2pt-5pt-2.1cm-6pt\relax}@{}}
& \raisebox{-5pt}{\includegraphics[width=2.1cm]{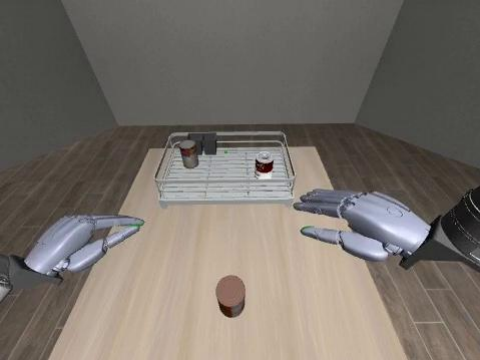}} &
\textbf{22.~\texttt{load\_\allowbreak three\_\allowbreak canned\_\allowbreak food\_\allowbreak onto\_\allowbreak dish\_\allowbreak rack}}\hfill {\footnotesize $K{=}6$ subtasks $\cdot$ 100 demos}\newline
{\small\emph{Instruction:} Pick up the three cans of food and set each onto a tier of the wooden dish rack to dry.}\newline
{\footnotesize\textbf{Scoring:} Pick up the first can \textbf{(10)} $\cdot$ Set the first can onto the dish rack \textbf{(20)}$^{\dagger}$ $\cdot$ Pick up the second can \textbf{(10)} $\cdot$ Set the second can onto the dish rack \textbf{(25)}$^{\dagger}$ $\cdot$ Pick up the third can \textbf{(10)} $\cdot$ Set the third can onto the dish rack \textbf{(25)}$^{\dagger}$.}
\end{tabular}}

\par\vspace{3pt}\noindent
{\renewcommand{\arraystretch}{1.0}%
\begin{tabular}{@{}>{\columncolor{catgen}}p{2.2pt}@{\hspace{5pt}}m{2.1cm}@{\hspace{6pt}}p{\dimexpr\linewidth-2.2pt-5pt-2.1cm-6pt\relax}@{}}
& \raisebox{-5pt}{\includegraphics[width=2.1cm]{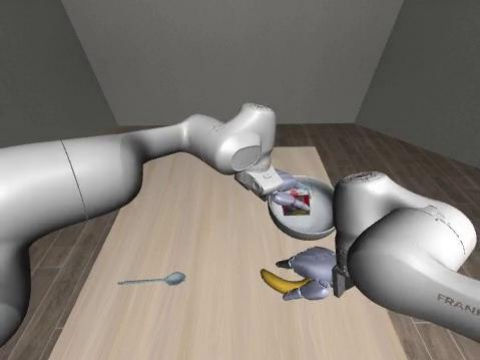}} &
\textbf{23.~\texttt{make\_\allowbreak cereal\_\allowbreak in\_\allowbreak bowl}}\hfill {\footnotesize $K{=}6$ subtasks $\cdot$ 100 demos}\newline
{\small\emph{Instruction:} Pick up the cereal box and put it into the red bowl, then add the banana, and place the spoon inside the bowl.}\newline
{\footnotesize\textbf{Scoring:} Pick up the cereal box \textbf{(15)} $\cdot$ Put the cereal into the red bowl \textbf{(15)}$^{\dagger}$ $\cdot$ Pick up the banana \textbf{(15)} $\cdot$ Put the banana into the red bowl \textbf{(20)}$^{\dagger}$ $\cdot$ Pick up the spoon \textbf{(15)} $\cdot$ Put the spoon inside the red bowl \textbf{(20)}$^{\dagger}$.}
\end{tabular}}

\par\vspace{3pt}\noindent
{\renewcommand{\arraystretch}{1.0}%
\begin{tabular}{@{}>{\columncolor{catgen}}p{2.2pt}@{\hspace{5pt}}m{2.1cm}@{\hspace{6pt}}p{\dimexpr\linewidth-2.2pt-5pt-2.1cm-6pt\relax}@{}}
& \raisebox{-5pt}{\includegraphics[width=2.1cm]{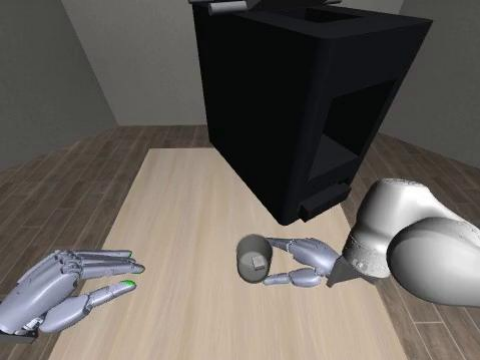}} &
\textbf{24.~\texttt{make\_\allowbreak espresso}}\hfill {\footnotesize $K{=}4$ subtasks $\cdot$ 100 demos}\newline
{\small\emph{Instruction:} Put the sugar cube into the cup, then place the cup in the espresso machine's front slot.}\newline
{\footnotesize\textbf{Scoring:} Pick up the sugar cube \textbf{(25)} $\cdot$ Drop the sugar cube into the cup \textbf{(25)}$^{\dagger}$ $\cdot$ Pick up the cup \textbf{(25)} $\cdot$ Place the cup in the espresso machine front slot \textbf{(25)}$^{\dagger}$.}
\end{tabular}}

\par\vspace{3pt}\noindent
{\renewcommand{\arraystretch}{1.0}%
\begin{tabular}{@{}>{\columncolor{catgen}}p{2.2pt}@{\hspace{5pt}}m{2.1cm}@{\hspace{6pt}}p{\dimexpr\linewidth-2.2pt-5pt-2.1cm-6pt\relax}@{}}
& \raisebox{-5pt}{\includegraphics[width=2.1cm]{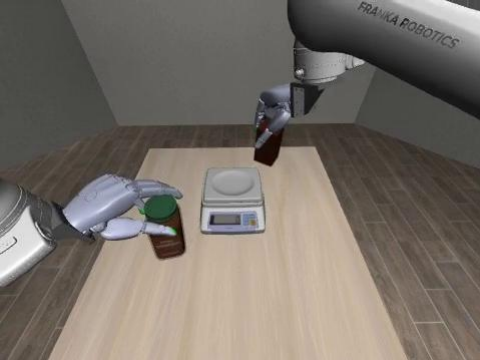}} &
\textbf{25.~\texttt{measure\_\allowbreak two\_\allowbreak paprika\_\allowbreak shakers\_\allowbreak on\_\allowbreak digital\_\allowbreak scale}}\hfill {\footnotesize $K{=}6$ subtasks $\cdot$ 100 demos}\newline
{\small\emph{Instruction:} Take the paprika shakers one at a time, put each on the digital scale, and then remove it.}\newline
{\footnotesize\textbf{Scoring:} Pick up the first paprika shaker \textbf{(10)} $\to$ Place the first paprika shaker on the digital scale \textbf{(20)} $\to$ Pick up the second paprika shaker \textbf{(20)} $\to$ Place the second paprika shaker on the digital scale \textbf{(10)} $\to$ Take the first paprika shaker off the digital scale \textbf{(20)}$^{\dagger}$ $\to$ Take the second paprika shaker off the digital scale \textbf{(20)}$^{\dagger}$.}
\end{tabular}}

\par\vspace{3pt}\noindent
{\renewcommand{\arraystretch}{1.0}%
\begin{tabular}{@{}>{\columncolor{catgen}}p{2.2pt}@{\hspace{5pt}}m{2.1cm}@{\hspace{6pt}}p{\dimexpr\linewidth-2.2pt-5pt-2.1cm-6pt\relax}@{}}
& \raisebox{-5pt}{\includegraphics[width=2.1cm]{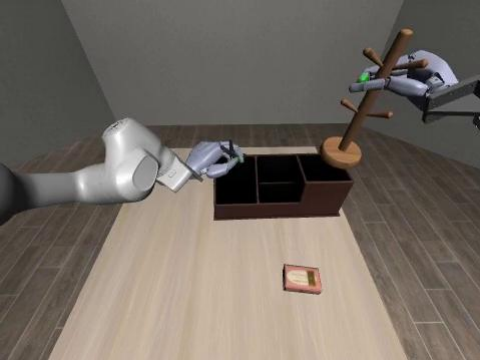}} &
\textbf{26.~\texttt{organize\_\allowbreak items\_\allowbreak in\_\allowbreak desk\_\allowbreak caddy}}\hfill {\footnotesize $K{=}6$ subtasks $\cdot$ 100 demos}\newline
{\small\emph{Instruction:} Pick up the scissors and put them into the desk caddy, then place the soap into the desk caddy, and put the mug tree into the caddy.}\newline
{\footnotesize\textbf{Scoring:} Pick up the scissors \textbf{(15)} $\cdot$ Put the scissors into the desk caddy \textbf{(20)}$^{\dagger}$ $\cdot$ Pick up the soap \textbf{(15)} $\cdot$ Put the soap into the desk caddy \textbf{(15)}$^{\dagger}$ $\cdot$ Pick up the mug tree \textbf{(15)} $\cdot$ Place the mug tree into the desk caddy \textbf{(20)}$^{\dagger}$.}
\end{tabular}}

\par\vspace{3pt}\noindent
{\renewcommand{\arraystretch}{1.0}%
\begin{tabular}{@{}>{\columncolor{catgen}}p{2.2pt}@{\hspace{5pt}}m{2.1cm}@{\hspace{6pt}}p{\dimexpr\linewidth-2.2pt-5pt-2.1cm-6pt\relax}@{}}
& \raisebox{-5pt}{\includegraphics[width=2.1cm]{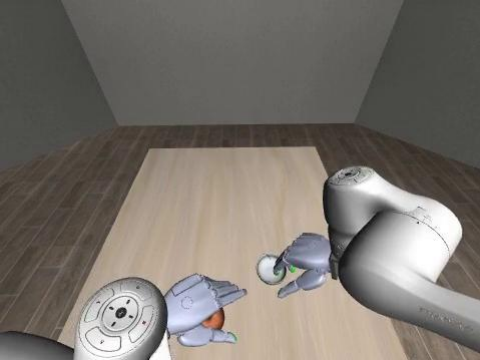}} &
\textbf{27.~\texttt{put\_\allowbreak apple\_\allowbreak in\_\allowbreak white\_\allowbreak bowl}}\hfill {\footnotesize $K{=}2$ subtasks $\cdot$ 100 demos}\newline
{\small\emph{Instruction:} Pick up the apple and put it inside the white bowl.}\newline
{\footnotesize\textbf{Scoring:} Pick up the apple \textbf{(50)} $\cdot$ Put the apple inside the white bowl \textbf{(50)}$^{\dagger}$.}
\end{tabular}}

\par\vspace{3pt}\noindent
{\renewcommand{\arraystretch}{1.0}%
\begin{tabular}{@{}>{\columncolor{catgen}}p{2.2pt}@{\hspace{5pt}}m{2.1cm}@{\hspace{6pt}}p{\dimexpr\linewidth-2.2pt-5pt-2.1cm-6pt\relax}@{}}
& \raisebox{-5pt}{\includegraphics[width=2.1cm]{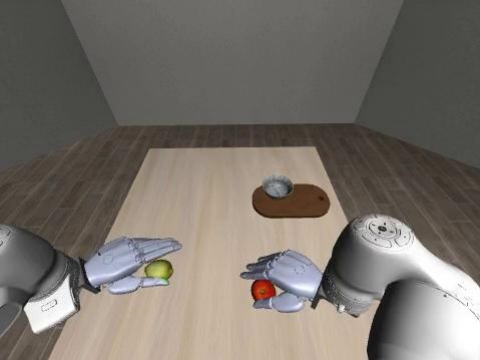}} &
\textbf{28.~\texttt{prepare\_\allowbreak salad\_\allowbreak in\_\allowbreak bowl}}\hfill {\footnotesize $K{=}6$ subtasks $\cdot$ 100 demos}\newline
{\small\emph{Instruction:} Place the bowl on the cutting board, then put the tomato and the lemon inside the bowl.}\newline
{\footnotesize\textbf{Scoring:} Pick up the black bowl \textbf{(15)} $\cdot$ Place the black bowl on the cutting board \textbf{(15)}$^{\dagger}$ $\cdot$ Pick up the tomato \textbf{(15)} $\cdot$ Put the tomato inside the bowl \textbf{(20)}$^{\dagger}$ $\cdot$ Pick up the lemon \textbf{(15)} $\cdot$ Put the lemon inside the bowl \textbf{(20)}$^{\dagger}$.}
\end{tabular}}

\par\vspace{3pt}\noindent
{\renewcommand{\arraystretch}{1.0}%
\begin{tabular}{@{}>{\columncolor{catgen}}p{2.2pt}@{\hspace{5pt}}m{2.1cm}@{\hspace{6pt}}p{\dimexpr\linewidth-2.2pt-5pt-2.1cm-6pt\relax}@{}}
& \raisebox{-5pt}{\includegraphics[width=2.1cm]{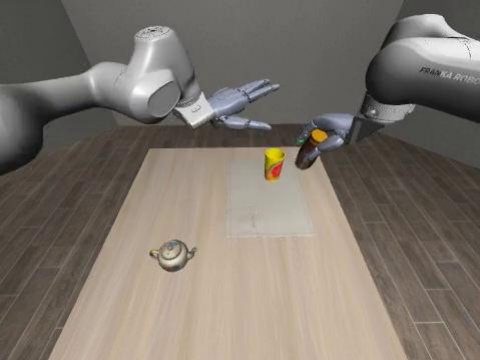}} &
\textbf{29.~\texttt{prepare\_\allowbreak tea\_\allowbreak with\_\allowbreak cup}}\hfill {\footnotesize $K{=}6$ subtasks $\cdot$ 100 demos}\newline
{\small\emph{Instruction:} Pick up the cup and place it on the placemat, then put the honey bottle next to the cup, and place the teapot on the placemat.}\newline
{\footnotesize\textbf{Scoring:} Pick up the cup \textbf{(15)} $\cdot$ Place the cup on the placemat \textbf{(15)}$^{\dagger}$ $\cdot$ Pick up the honey bottle \textbf{(15)} $\cdot$ Place the honey bottle on the placemat \textbf{(20)}$^{\dagger}$ $\cdot$ Pick up the teapot \textbf{(15)} $\cdot$ Place the teapot on the placemat \textbf{(20)}$^{\dagger}$.}
\end{tabular}}

\par\vspace{3pt}\noindent
{\renewcommand{\arraystretch}{1.0}%
\begin{tabular}{@{}>{\columncolor{catgen}}p{2.2pt}@{\hspace{5pt}}m{2.1cm}@{\hspace{6pt}}p{\dimexpr\linewidth-2.2pt-5pt-2.1cm-6pt\relax}@{}}
& \raisebox{-5pt}{\includegraphics[width=2.1cm]{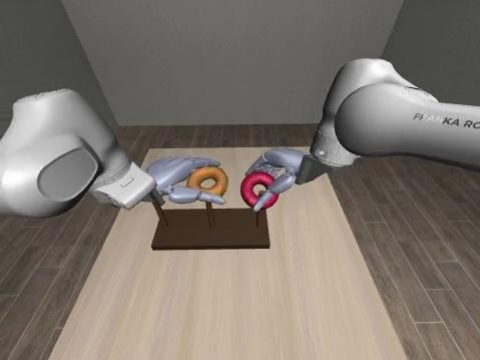}} &
\textbf{30.~\texttt{rethread\_\allowbreak bagel\_\allowbreak and\_\allowbreak donut\_\allowbreak onto\_\allowbreak middle\_\allowbreak rod}}\hfill {\footnotesize $K{=}4$ subtasks $\cdot$ 100 demos}\newline
{\small\emph{Instruction:} Move the bagel from the left rod to the middle rod, then the donut from the right rod to the middle rod.}\newline
{\footnotesize\textbf{Scoring:} Left arm lifts the bagel off the left rod \textbf{(25)} $\to$ Left arm re-threads the bagel onto the middle rod \textbf{(25)}$^{\dagger}$ $\to$ Right arm lifts the donut off the right rod \textbf{(25)} $\to$ Right arm re-threads the donut onto the middle rod \textbf{(25)}$^{\dagger}$.}
\end{tabular}}

\par\vspace{3pt}\noindent
{\renewcommand{\arraystretch}{1.0}%
\begin{tabular}{@{}>{\columncolor{catgen}}p{2.2pt}@{\hspace{5pt}}m{2.1cm}@{\hspace{6pt}}p{\dimexpr\linewidth-2.2pt-5pt-2.1cm-6pt\relax}@{}}
& \raisebox{-5pt}{\includegraphics[width=2.1cm]{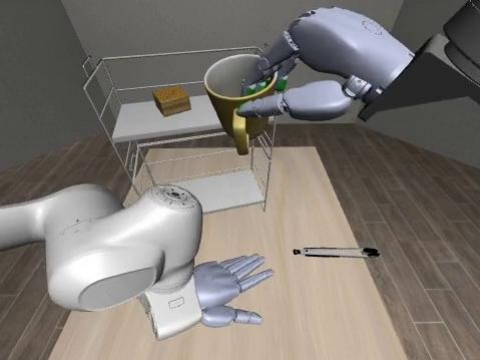}} &
\textbf{31.~\texttt{place\_\allowbreak cake\_\allowbreak tongs\_\allowbreak and\_\allowbreak mug\_\allowbreak on\_\allowbreak tiered\_\allowbreak shelf}}\hfill {\footnotesize $K{=}6$ subtasks $\cdot$ 100 demos}\newline
{\small\emph{Instruction:} Place the cake, the tongs, and the mug on the tiered shelf.}\newline
{\footnotesize\textbf{Scoring:} Pick up the cake \textbf{(15)} $\cdot$ Place the cake on the tiered shelf \textbf{(15)}$^{\dagger}$ $\cdot$ Pick up the mug \textbf{(15)} $\cdot$ Place the mug on the tiered shelf \textbf{(20)}$^{\dagger}$ $\cdot$ Pick up the tongs \textbf{(15)} $\cdot$ Place the tongs on the tiered shelf \textbf{(20)}$^{\dagger}$.}
\end{tabular}}

\par\vspace{3pt}\noindent
{\renewcommand{\arraystretch}{1.0}%
\begin{tabular}{@{}>{\columncolor{catgen}}p{2.2pt}@{\hspace{5pt}}m{2.1cm}@{\hspace{6pt}}p{\dimexpr\linewidth-2.2pt-5pt-2.1cm-6pt\relax}@{}}
& \raisebox{-5pt}{\includegraphics[width=2.1cm]{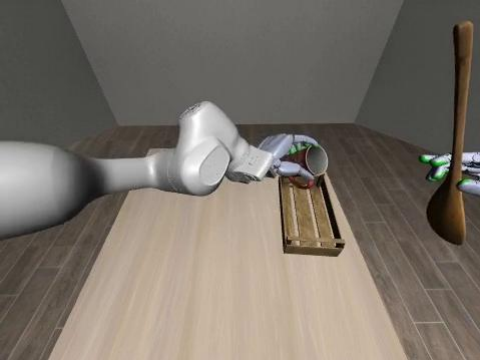}} &
\textbf{32.~\texttt{place\_\allowbreak coffee\_\allowbreak mug\_\allowbreak and\_\allowbreak spoon\_\allowbreak on\_\allowbreak tray}}\hfill {\footnotesize $K{=}4$ subtasks $\cdot$ 100 demos}\newline
{\small\emph{Instruction:} Place the red coffee mug and the wooden spoon on the wooden tray.}\newline
{\footnotesize\textbf{Scoring:} Pick up the red coffee mug \textbf{(25)} $\cdot$ Place the red coffee mug on the wooden tray \textbf{(25)}$^{\dagger}$ $\cdot$ Pick up the wooden spoon \textbf{(25)} $\cdot$ Put the wooden spoon on the wooden tray \textbf{(25)}$^{\dagger}$.}
\end{tabular}}

\par\vspace{3pt}\noindent
{\renewcommand{\arraystretch}{1.0}%
\begin{tabular}{@{}>{\columncolor{catgen}}p{2.2pt}@{\hspace{5pt}}m{2.1cm}@{\hspace{6pt}}p{\dimexpr\linewidth-2.2pt-5pt-2.1cm-6pt\relax}@{}}
& \raisebox{-5pt}{\includegraphics[width=2.1cm]{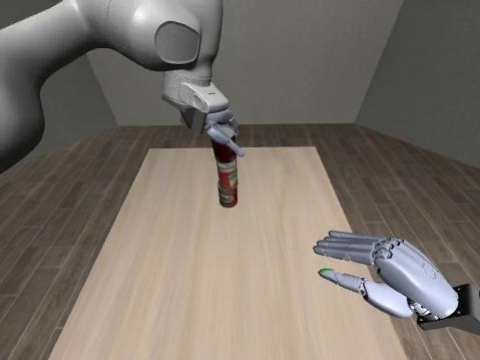}} &
\textbf{33.~\texttt{stack\_\allowbreak three\_\allowbreak canned\_\allowbreak food}}\hfill {\footnotesize $K{=}4$ subtasks $\cdot$ 100 demos}\newline
{\small\emph{Instruction:} Stack the three cans on top of each other.}\newline
{\footnotesize\textbf{Scoring:} Pick up the middle can \textbf{(10)} $\cdot$ Stack the middle can on the base can \textbf{(40)}$^{\dagger}$ $\cdot$ Pick up the top can \textbf{(10)} $\cdot$ Stack the top can on the middle can \textbf{(40)}$^{\dagger}$.}
\end{tabular}}

\par\vspace{3pt}\noindent
{\renewcommand{\arraystretch}{1.0}%
\begin{tabular}{@{}>{\columncolor{catgen}}p{2.2pt}@{\hspace{5pt}}m{2.1cm}@{\hspace{6pt}}p{\dimexpr\linewidth-2.2pt-5pt-2.1cm-6pt\relax}@{}}
& \raisebox{-5pt}{\includegraphics[width=2.1cm]{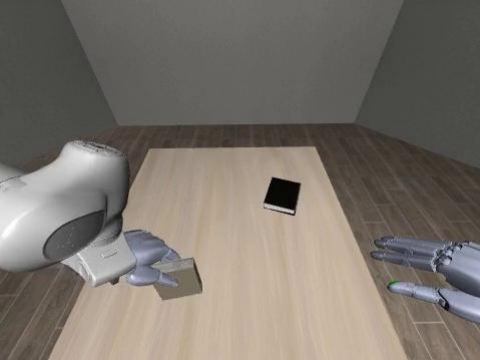}} &
\textbf{34.~\texttt{stack\_\allowbreak yellow\_\allowbreak book\_\allowbreak on\_\allowbreak black\_\allowbreak book}}\hfill {\footnotesize $K{=}2$ subtasks $\cdot$ 100 demos}\newline
{\small\emph{Instruction:} Pick up the yellow book and stack it on top of the black book.}\newline
{\footnotesize\textbf{Scoring:} Pick up the yellow book \textbf{(40)} $\cdot$ Stack the yellow book on the black book \textbf{(60)}$^{\dagger}$.}
\end{tabular}}

\par\vspace{3pt}\noindent
{\renewcommand{\arraystretch}{1.0}%
\begin{tabular}{@{}>{\columncolor{catgen}}p{2.2pt}@{\hspace{5pt}}m{2.1cm}@{\hspace{6pt}}p{\dimexpr\linewidth-2.2pt-5pt-2.1cm-6pt\relax}@{}}
& \raisebox{-5pt}{\includegraphics[width=2.1cm]{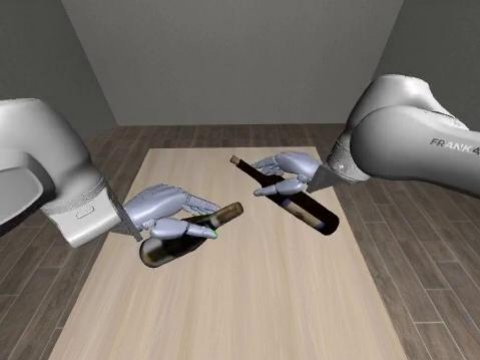}} &
\textbf{35.~\texttt{stand\_\allowbreak two\_\allowbreak wine\_\allowbreak bottles\_\allowbreak upright}}\hfill {\footnotesize $K{=}4$ subtasks $\cdot$ 100 demos}\newline
{\small\emph{Instruction:} Pick up the two wine bottles lying on their sides and stand them upright on the table.}\newline
{\footnotesize\textbf{Scoring:} Pick up the first wine bottle \textbf{(25)} $\to$ Stand the first wine bottle upright \textbf{(25)}$^{\dagger}$ $\to$ Pick up the second wine bottle \textbf{(25)} $\to$ Stand the second wine bottle upright \textbf{(25)}$^{\dagger}$.}
\end{tabular}}

\par\vspace{3pt}\noindent
{\renewcommand{\arraystretch}{1.0}%
\begin{tabular}{@{}>{\columncolor{catgen}}p{2.2pt}@{\hspace{5pt}}m{2.1cm}@{\hspace{6pt}}p{\dimexpr\linewidth-2.2pt-5pt-2.1cm-6pt\relax}@{}}
& \raisebox{-5pt}{\includegraphics[width=2.1cm]{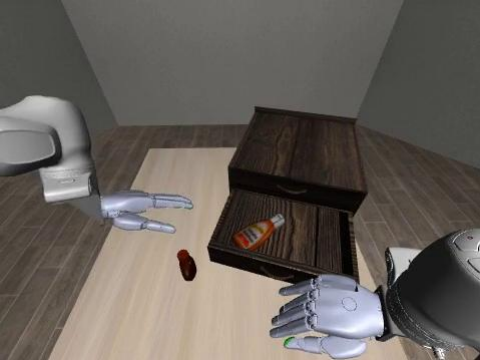}} &
\textbf{36.~\texttt{store\_\allowbreak condiments\_\allowbreak in\_\allowbreak short\_\allowbreak cabinet}}\hfill {\footnotesize $K{=}6$ subtasks $\cdot$ 100 demos}\newline
{\small\emph{Instruction:} Open the short cabinet drawer, put the ketchup inside, put the BBQ sauce inside, then close the drawer.}\newline
{\footnotesize\textbf{Scoring:} Open the short cabinet drawer \textbf{(15)} $\to$ Pick up the ketchup \textbf{(15)} $\to$ Put the ketchup inside the short cabinet \textbf{(20)} $\to$ Pick up the BBQ sauce \textbf{(15)} $\to$ Put the BBQ sauce inside the short cabinet \textbf{(20)} $\to$ Close the short cabinet drawer \textbf{(15)}.}
\end{tabular}}

\par\vspace{3pt}\noindent
{\renewcommand{\arraystretch}{1.0}%
\begin{tabular}{@{}>{\columncolor{catgen}}p{2.2pt}@{\hspace{5pt}}m{2.1cm}@{\hspace{6pt}}p{\dimexpr\linewidth-2.2pt-5pt-2.1cm-6pt\relax}@{}}
& \raisebox{-5pt}{\includegraphics[width=2.1cm]{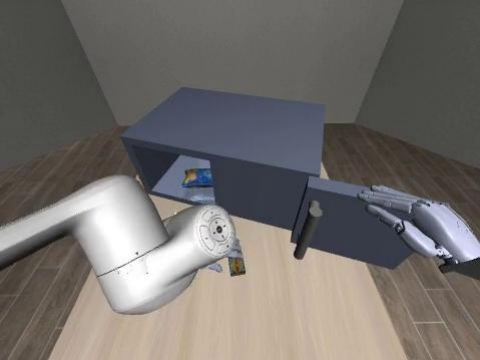}} &
\textbf{37.~\texttt{store\_\allowbreak snacks\_\allowbreak in\_\allowbreak slide\_\allowbreak cabinet}}\hfill {\footnotesize $K{=}6$ subtasks $\cdot$ 100 demos}\newline
{\small\emph{Instruction:} Open the slide cabinet, put the popcorn inside, put the chips inside, then close the slide cabinet.}\newline
{\footnotesize\textbf{Scoring:} Open the slide cabinet door \textbf{(15)} $\to$ Pick up the popcorn \textbf{(15)} $\to$ Put the popcorn inside the slide cabinet \textbf{(20)}$^{\dagger}$ $\to$ Pick up the chips \textbf{(15)} $\to$ Put the chips inside the slide cabinet \textbf{(20)}$^{\dagger}$ $\to$ Close the slide cabinet door \textbf{(15)}.}
\end{tabular}}

\par\vspace{3pt}\noindent
{\renewcommand{\arraystretch}{1.0}%
\begin{tabular}{@{}>{\columncolor{catgen}}p{2.2pt}@{\hspace{5pt}}m{2.1cm}@{\hspace{6pt}}p{\dimexpr\linewidth-2.2pt-5pt-2.1cm-6pt\relax}@{}}
& \raisebox{-5pt}{\includegraphics[width=2.1cm]{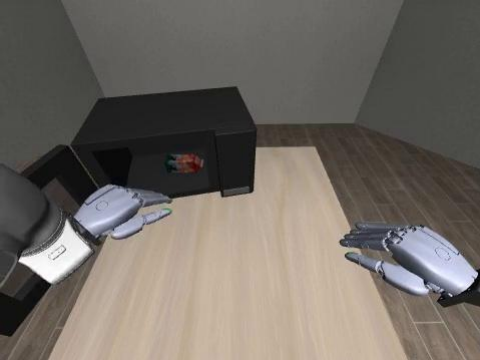}} &
\textbf{38.~\texttt{take\_\allowbreak milk\_\allowbreak from\_\allowbreak microwave}}\hfill {\footnotesize $K{=}3$ subtasks $\cdot$ 100 demos}\newline
{\small\emph{Instruction:} Open the microwave, take out the milk, and place it on the table.}\newline
{\footnotesize\textbf{Scoring:} Open the microwave door \textbf{(40)} $\cdot$ Pick up the milk \textbf{(30)} $\cdot$ Place the milk on the table \textbf{(30)}.}
\end{tabular}}

\par\vspace{3pt}\noindent
{\renewcommand{\arraystretch}{1.0}%
\begin{tabular}{@{}>{\columncolor{catgen}}p{2.2pt}@{\hspace{5pt}}m{2.1cm}@{\hspace{6pt}}p{\dimexpr\linewidth-2.2pt-5pt-2.1cm-6pt\relax}@{}}
& \raisebox{-5pt}{\includegraphics[width=2.1cm]{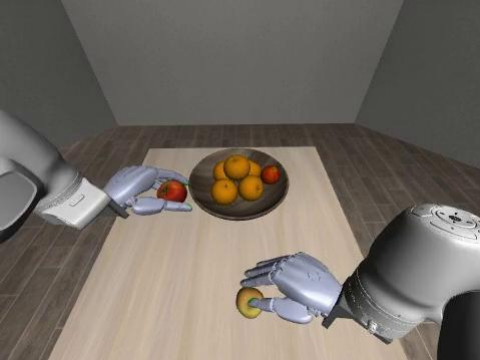}} &
\textbf{39.~\texttt{put\_\allowbreak three\_\allowbreak apples\_\allowbreak into\_\allowbreak fruit\_\allowbreak bowl}}\hfill {\footnotesize $K{=}1$ subtask $\cdot$ 100 demos}\newline
{\small\emph{Instruction:} Pick up the three apples and place them into the fruit bowl.}\newline
{\footnotesize\textbf{Scoring:} Place the item into the container \textbf{(100)}$^{\dagger}$.}
\end{tabular}}

\par\vspace{3pt}\noindent
{\renewcommand{\arraystretch}{1.0}%
\begin{tabular}{@{}>{\columncolor{catgen}}p{2.2pt}@{\hspace{5pt}}m{2.1cm}@{\hspace{6pt}}p{\dimexpr\linewidth-2.2pt-5pt-2.1cm-6pt\relax}@{}}
& \raisebox{-5pt}{\includegraphics[width=2.1cm]{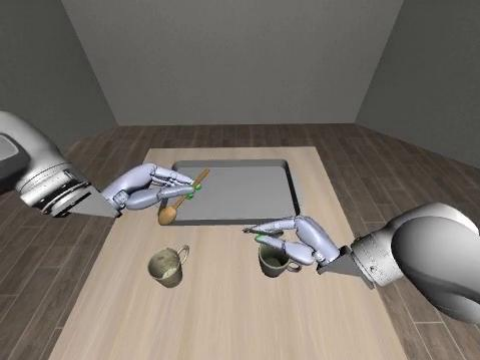}} &
\textbf{40.~\texttt{drop\_\allowbreak two\_\allowbreak spoons\_\allowbreak from\_\allowbreak tray\_\allowbreak into\_\allowbreak two\_\allowbreak cups}}\hfill {\footnotesize $K{=}2$ subtasks $\cdot$ 100 demos}\newline
{\small\emph{Instruction:} With each arm pick up a spoon from the tray and drop it, one each, into the two cups.}\newline
{\footnotesize\textbf{Scoring:} Drop the first spoon into a cup \textbf{(50)}$^{\dagger}$ $\to$ Drop the second spoon into a cup \textbf{(50)}$^{\dagger}$.}
\end{tabular}}

\subsection{Instruction-Following (20 specifications, 20 with released demonstrations)}
\label{app:tasks:instruction_following}

Verbal-command fidelity, tested three ways: \emph{count} variants share one scene but demand different object counts, \emph{combo} tasks conjoin two instructions, and \emph{switch} tasks replace the instruction mid-episode through a language schedule, so the policy must serve the current command rather than a remembered one. For switch tasks the two instruction phases are listed as Phase~1 and Phase~2; grasping a wrong object terminates the episode.

\par\vspace{3pt}\noindent
{\renewcommand{\arraystretch}{1.0}%
\begin{tabular}{@{}>{\columncolor{catif}}p{2.2pt}@{\hspace{5pt}}m{2.1cm}@{\hspace{6pt}}p{\dimexpr\linewidth-2.2pt-5pt-2.1cm-6pt\relax}@{}}
& \raisebox{-5pt}{\includegraphics[width=2.1cm]{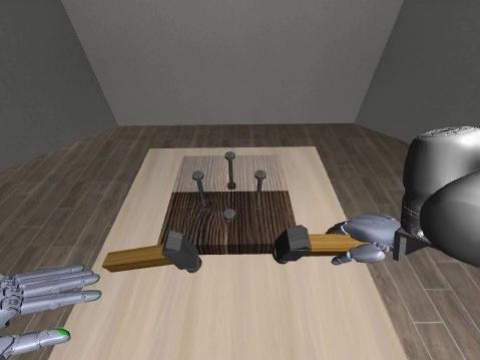}} &
\textbf{1.~\texttt{hammer\_\allowbreak one\_\allowbreak nail\_\allowbreak each\_\allowbreak hand}}\hfill {\footnotesize $K{=}4$ subtasks $\cdot$ 100 demos}\newline
{\small\emph{Instruction:} With the left arm use the left hammer to drive the up nail into the board, and with the right arm use the right hammer to drive the down nail.}\newline
{\footnotesize\textbf{Scoring:} Left arm drives the up nail into the board \textbf{(25)} $\cdot$ Right arm drives the down nail into the board \textbf{(25)} $\cdot$ Extra left nail is not driven into the board \textbf{(25)}$^{\dagger}$ $\cdot$ Extra right nail is not driven into the board \textbf{(25)}$^{\dagger}$.}
\end{tabular}}

\par\vspace{3pt}\noindent
{\renewcommand{\arraystretch}{1.0}%
\begin{tabular}{@{}>{\columncolor{catif}}p{2.2pt}@{\hspace{5pt}}m{2.1cm}@{\hspace{6pt}}p{\dimexpr\linewidth-2.2pt-5pt-2.1cm-6pt\relax}@{}}
& \raisebox{-5pt}{\includegraphics[width=2.1cm]{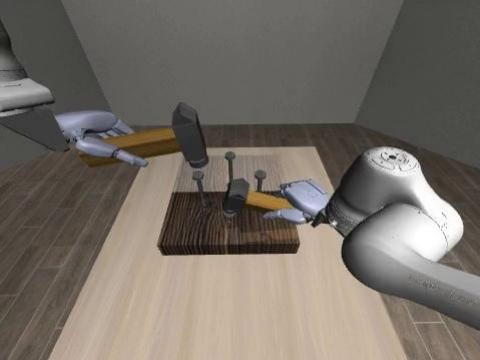}} &
\textbf{2.~\texttt{hammer\_\allowbreak two\_\allowbreak nails\_\allowbreak each\_\allowbreak hand}}\hfill {\footnotesize $K{=}4$ subtasks $\cdot$ 100 demos}\newline
{\small\emph{Instruction:} With the left arm use the left hammer to drive the two left nails into the board, and with the right arm use the right hammer to drive the two right nails.}\newline
{\footnotesize\textbf{Scoring:} Left arm drives the first left nail into the board \textbf{(25)} $\cdot$ Left arm drives the second left nail into the board \textbf{(25)} $\cdot$ Right arm drives the first right nail into the board \textbf{(25)} $\cdot$ Right arm drives the second right nail into the board \textbf{(25)}.}
\end{tabular}}

\par\vspace{3pt}\noindent
{\renewcommand{\arraystretch}{1.0}%
\begin{tabular}{@{}>{\columncolor{catif}}p{2.2pt}@{\hspace{5pt}}m{2.1cm}@{\hspace{6pt}}p{\dimexpr\linewidth-2.2pt-5pt-2.1cm-6pt\relax}@{}}
& \raisebox{-5pt}{\includegraphics[width=2.1cm]{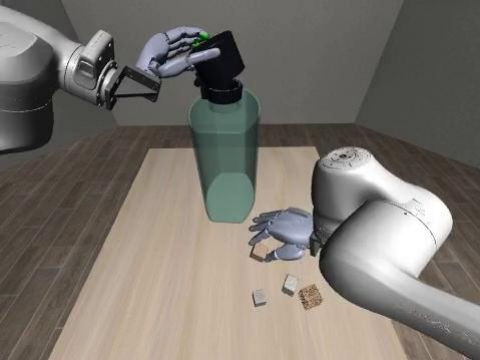}} &
\textbf{3.~\texttt{open\_\allowbreak bottle\_\allowbreak and\_\allowbreak drop\_\allowbreak one\_\allowbreak sugar\_\allowbreak cube\_\allowbreak in\_\allowbreak bottle}}\hfill {\footnotesize $K{=}9$ subtasks $\cdot$ 100 demos}\newline
{\small\emph{Instruction:} With the left arm flip open the bottle's articulated cap, and with the right arm pick up the first sugar cube and drop it into the bottle.}\newline
{\footnotesize\textbf{Scoring:} Left arm flips open the bottle's articulated cap \textbf{(10)} $\cdot$ Right arm picks up the first sugar cube \textbf{(10)} $\cdot$ Right arm drops the first sugar cube into the bottle \textbf{(20)} $\cdot$ Extra sugar cube 1 never enters the bottle \textbf{(10)}$^{\ddagger}$ $\cdot$ Extra sugar cube 2 never enters the bottle \textbf{(10)}$^{\ddagger}$ $\cdot$ Extra sugar cube 3 never enters the bottle \textbf{(10)}$^{\ddagger}$ $\cdot$ Extra sugar cube 1 is never picked up \textbf{(10)}$^{\ddagger}$ $\cdot$ Extra sugar cube 2 is never picked up \textbf{(10)}$^{\ddagger}$ $\cdot$ Extra sugar cube 3 is never picked up \textbf{(10)}$^{\ddagger}$.}
\end{tabular}}

\par\vspace{3pt}\noindent
{\renewcommand{\arraystretch}{1.0}%
\begin{tabular}{@{}>{\columncolor{catif}}p{2.2pt}@{\hspace{5pt}}m{2.1cm}@{\hspace{6pt}}p{\dimexpr\linewidth-2.2pt-5pt-2.1cm-6pt\relax}@{}}
& \raisebox{-5pt}{\includegraphics[width=2.1cm]{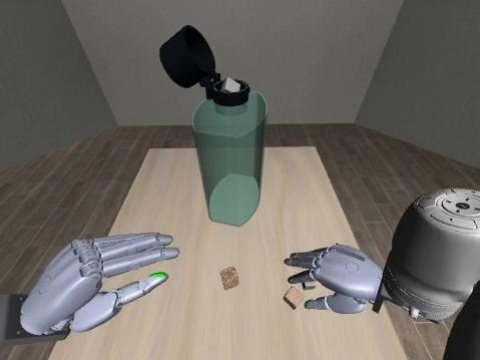}} &
\textbf{4.~\texttt{open\_\allowbreak bottle\_\allowbreak and\_\allowbreak drop\_\allowbreak two\_\allowbreak sugar\_\allowbreak cubes\_\allowbreak in\_\allowbreak bottle}}\hfill {\footnotesize $K{=}9$ subtasks $\cdot$ 100 demos}\newline
{\small\emph{Instruction:} With the left arm flip open the bottle's articulated cap, and with the right arm pick up the first and second sugar cubes and drop them into the bottle.}\newline
{\footnotesize\textbf{Scoring:} Left arm flips open the bottle's articulated cap \textbf{(10)} $\cdot$ Right arm picks up the first sugar cube \textbf{(10)} $\cdot$ Right arm drops the first sugar cube into the bottle \textbf{(15)} $\cdot$ Right arm picks up the second sugar cube \textbf{(10)} $\cdot$ Right arm drops the second sugar cube into the bottle \textbf{(15)} $\cdot$ Extra sugar cube 1 never enters the bottle \textbf{(10)}$^{\ddagger}$ $\cdot$ Extra sugar cube 2 never enters the bottle \textbf{(10)}$^{\ddagger}$ $\cdot$ Extra sugar cube 1 is never picked up \textbf{(10)}$^{\ddagger}$ $\cdot$ Extra sugar cube 2 is never picked up \textbf{(10)}$^{\ddagger}$.}
\end{tabular}}

\par\vspace{3pt}\noindent
{\renewcommand{\arraystretch}{1.0}%
\begin{tabular}{@{}>{\columncolor{catif}}p{2.2pt}@{\hspace{5pt}}m{2.1cm}@{\hspace{6pt}}p{\dimexpr\linewidth-2.2pt-5pt-2.1cm-6pt\relax}@{}}
& \raisebox{-5pt}{\includegraphics[width=2.1cm]{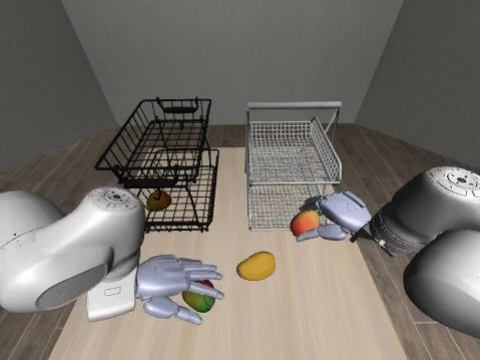}} &
\textbf{5.~\texttt{pack\_\allowbreak four\_\allowbreak mangoes\_\allowbreak into\_\allowbreak baskets}}\hfill {\footnotesize $K{=}8$ subtasks $\cdot$ 100 demos}\newline
{\small\emph{Instruction:} With the left arm pick up the two leftmost mangoes and place them into the left basket, and with the right arm pick up the two rightmost mangoes and place them into the right basket.}\newline
{\footnotesize\textbf{Scoring:} Left arm picks up the first (leftmost) mango \textbf{(10)} $\cdot$ Left arm places the first mango into the left basket \textbf{(20)}$^{\dagger}$ $\cdot$ Left arm picks up the second mango \textbf{(10)} $\cdot$ Left arm places the second mango into the left basket \textbf{(10)}$^{\dagger}$ $\cdot$ Right arm picks up the fourth (rightmost) mango \textbf{(10)} $\cdot$ Right arm places the fourth mango into the right basket \textbf{(20)}$^{\dagger}$ $\cdot$ Right arm picks up the third mango \textbf{(10)} $\cdot$ Right arm places the third mango into the right basket \textbf{(10)}$^{\dagger}$.}
\end{tabular}}

\par\vspace{3pt}\noindent
{\renewcommand{\arraystretch}{1.0}%
\begin{tabular}{@{}>{\columncolor{catif}}p{2.2pt}@{\hspace{5pt}}m{2.1cm}@{\hspace{6pt}}p{\dimexpr\linewidth-2.2pt-5pt-2.1cm-6pt\relax}@{}}
& \raisebox{-5pt}{\includegraphics[width=2.1cm]{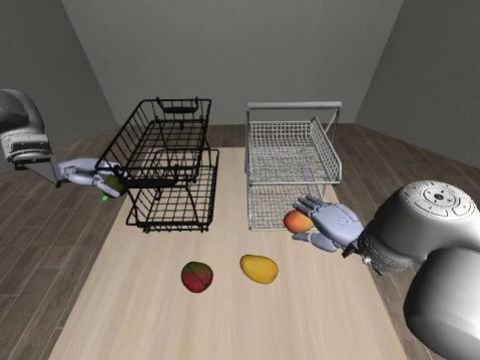}} &
\textbf{6.~\texttt{pack\_\allowbreak two\_\allowbreak mangoes\_\allowbreak into\_\allowbreak two\_\allowbreak baskets}}\hfill {\footnotesize $K{=}10$ subtasks $\cdot$ 100 demos}\newline
{\small\emph{Instruction:} With the left arm pick up the leftmost mango and place it into the left basket, and with the right arm pick up the rightmost mango and place it into the right basket.}\newline
{\footnotesize\textbf{Scoring:} Left arm picks up the first (leftmost) mango \textbf{(10)} $\cdot$ Left arm places the leftmost mango into the left basket \textbf{(10)}$^{\dagger}$ $\cdot$ Right arm picks up the fourth (rightmost) mango \textbf{(10)} $\cdot$ Right arm places the rightmost mango into the right basket \textbf{(10)}$^{\dagger}$ $\cdot$ Extra mango\_1 stays out of the left basket \textbf{(10)}$^{\ddagger}$ $\cdot$ Extra mango\_1 stays out of the right basket \textbf{(10)}$^{\ddagger}$ $\cdot$ Extra mango\_2 stays out of the left basket \textbf{(10)}$^{\ddagger}$ $\cdot$ Extra mango\_2 stays out of the right basket \textbf{(10)}$^{\ddagger}$ $\cdot$ Extra mango\_1 is not picked up \textbf{(10)}$^{\ddagger}$ $\cdot$ Extra mango\_2 is not picked up \textbf{(10)}$^{\ddagger}$.}
\end{tabular}}

\par\vspace{3pt}\noindent
{\renewcommand{\arraystretch}{1.0}%
\begin{tabular}{@{}>{\columncolor{catif}}p{2.2pt}@{\hspace{5pt}}m{2.1cm}@{\hspace{6pt}}p{\dimexpr\linewidth-2.2pt-5pt-2.1cm-6pt\relax}@{}}
& \raisebox{-5pt}{\includegraphics[width=2.1cm]{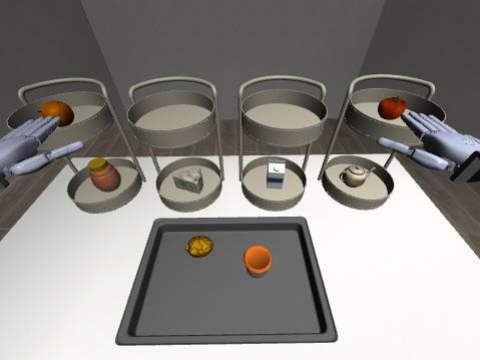}} &
\textbf{7.~\texttt{put\_\allowbreak jam\_\allowbreak and\_\allowbreak lemon\_\allowbreak from\_\allowbreak tray\_\allowbreak onto\_\allowbreak shelf}}\hfill {\footnotesize $K{=}2$ subtasks $\cdot$ 30 demos}\newline
{\small\emph{Instruction:} Take the jam and the lemon out of the tray and place them back onto the shelf.}\newline
{\footnotesize\textbf{Scoring:} Put the jam back onto the shelf \textbf{(50)}$^{\dagger}$ $\cdot$ Put the lemon back onto the shelf \textbf{(50)}$^{\dagger}$.}
\end{tabular}}

\par\vspace{3pt}\noindent
{\renewcommand{\arraystretch}{1.0}%
\begin{tabular}{@{}>{\columncolor{catif}}p{2.2pt}@{\hspace{5pt}}m{2.1cm}@{\hspace{6pt}}p{\dimexpr\linewidth-2.2pt-5pt-2.1cm-6pt\relax}@{}}
& \raisebox{-5pt}{\includegraphics[width=2.1cm]{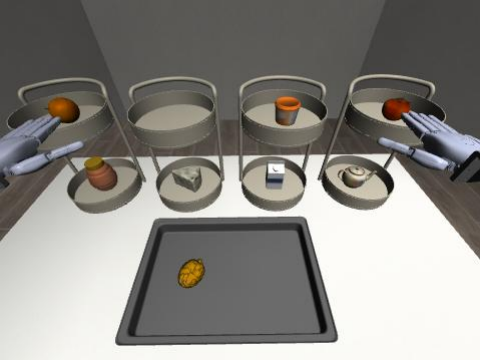}} &
\textbf{8.~\texttt{put\_\allowbreak lemon\_\allowbreak on\_\allowbreak shelf\_\allowbreak then\_\allowbreak take\_\allowbreak jam\_\allowbreak into\_\allowbreak tray}}\hfill {\footnotesize $K{=}2$ subtasks $\cdot$ 30 demos}\newline
{\small\emph{Instruction} (switches after control step 782)\emph{:} \emph{Phase 1:} Take the lemon out of the tray and place it onto the shelf. \emph{Phase 2:} Leave the lemon. Pick up the jam from the shelf and put it into the tray instead.}\newline
{\footnotesize\textbf{Scoring:} Put the lemon onto the shelf \textbf{(50)} $\to$ Put the jam into the tray \textbf{(50)}$^{\dagger}$.}
\end{tabular}}

\par\vspace{3pt}\noindent
{\renewcommand{\arraystretch}{1.0}%
\begin{tabular}{@{}>{\columncolor{catif}}p{2.2pt}@{\hspace{5pt}}m{2.1cm}@{\hspace{6pt}}p{\dimexpr\linewidth-2.2pt-5pt-2.1cm-6pt\relax}@{}}
& \raisebox{-5pt}{\includegraphics[width=2.1cm]{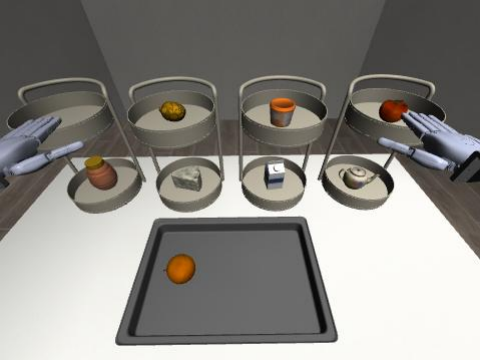}} &
\textbf{9.~\texttt{put\_\allowbreak orange\_\allowbreak on\_\allowbreak shelf\_\allowbreak then\_\allowbreak take\_\allowbreak back\_\allowbreak into\_\allowbreak tray}}\hfill {\footnotesize $K{=}2$ subtasks $\cdot$ 30 demos}\newline
{\small\emph{Instruction} (switches after control step 750)\emph{:} \emph{Phase 1:} Take the orange out of the tray and place it onto the shelf. \emph{Phase 2:} Put the orange back into the tray.}\newline
{\footnotesize\textbf{Scoring:} Put the orange onto the shelf \textbf{(50)} $\to$ Put the orange back into the tray \textbf{(50)}$^{\dagger}$.}
\end{tabular}}

\par\vspace{3pt}\noindent
{\renewcommand{\arraystretch}{1.0}%
\begin{tabular}{@{}>{\columncolor{catif}}p{2.2pt}@{\hspace{5pt}}m{2.1cm}@{\hspace{6pt}}p{\dimexpr\linewidth-2.2pt-5pt-2.1cm-6pt\relax}@{}}
& \raisebox{-5pt}{\includegraphics[width=2.1cm]{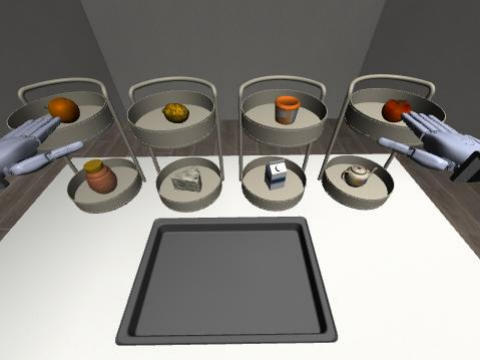}} &
\textbf{10.~\texttt{take\_\allowbreak jam\_\allowbreak into\_\allowbreak tray\_\allowbreak then\_\allowbreak put\_\allowbreak back\_\allowbreak on\_\allowbreak shelf}}\hfill {\footnotesize $K{=}2$ subtasks $\cdot$ 30 demos}\newline
{\small\emph{Instruction} (switches after control step 674)\emph{:} \emph{Phase 1:} Pick up the jam from the shelf and put it into the tray. \emph{Phase 2:} Put the jam back onto the shelf.}\newline
{\footnotesize\textbf{Scoring:} Put the jam into the tray \textbf{(50)} $\to$ Put the jam back onto the shelf \textbf{(50)}$^{\dagger}$.}
\end{tabular}}

\par\vspace{3pt}\noindent
{\renewcommand{\arraystretch}{1.0}%
\begin{tabular}{@{}>{\columncolor{catif}}p{2.2pt}@{\hspace{5pt}}m{2.1cm}@{\hspace{6pt}}p{\dimexpr\linewidth-2.2pt-5pt-2.1cm-6pt\relax}@{}}
& \raisebox{-5pt}{\includegraphics[width=2.1cm]{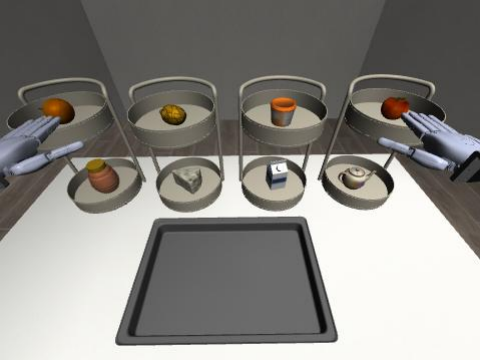}} &
\textbf{11.~\texttt{take\_\allowbreak milk\_\allowbreak and\_\allowbreak lemon\_\allowbreak from\_\allowbreak shelf\_\allowbreak into\_\allowbreak tray}}\hfill {\footnotesize $K{=}2$ subtasks $\cdot$ 30 demos}\newline
{\small\emph{Instruction} (switches after control step 492)\emph{:} \emph{Phase 1:} Pick up the milk carton from the shelf and put it into the tray. \emph{Phase 2:} Leave the milk. Pick up the lemon from the shelf and put it into the tray instead.}\newline
{\footnotesize\textbf{Scoring:} Put the milk carton into the tray \textbf{(50)} $\to$ Put the lemon into the tray \textbf{(50)}$^{\dagger}$.}
\end{tabular}}

\par\vspace{3pt}\noindent
{\renewcommand{\arraystretch}{1.0}%
\begin{tabular}{@{}>{\columncolor{catif}}p{2.2pt}@{\hspace{5pt}}m{2.1cm}@{\hspace{6pt}}p{\dimexpr\linewidth-2.2pt-5pt-2.1cm-6pt\relax}@{}}
& \raisebox{-5pt}{\includegraphics[width=2.1cm]{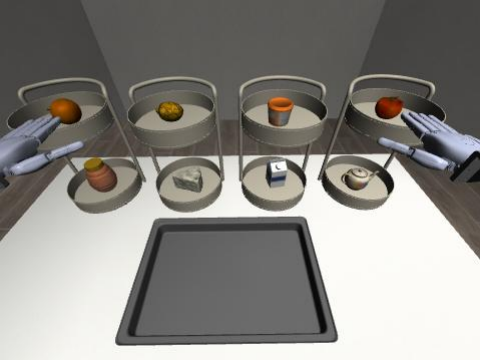}} &
\textbf{12.~\texttt{take\_\allowbreak teapot\_\allowbreak into\_\allowbreak tray\_\allowbreak then\_\allowbreak put\_\allowbreak back\_\allowbreak on\_\allowbreak shelf}}\hfill {\footnotesize $K{=}2$ subtasks $\cdot$ 30 demos}\newline
{\small\emph{Instruction} (switches after control step 536)\emph{:} \emph{Phase 1:} Pick up the teapot from the shelf and put it into the tray. \emph{Phase 2:} Put the teapot back onto the shelf.}\newline
{\footnotesize\textbf{Scoring:} Put the teapot into the tray \textbf{(50)} $\to$ Put the teapot back onto the shelf \textbf{(50)}$^{\dagger}$.}
\end{tabular}}

\par\vspace{3pt}\noindent
{\renewcommand{\arraystretch}{1.0}%
\begin{tabular}{@{}>{\columncolor{catif}}p{2.2pt}@{\hspace{5pt}}m{2.1cm}@{\hspace{6pt}}p{\dimexpr\linewidth-2.2pt-5pt-2.1cm-6pt\relax}@{}}
& \raisebox{-5pt}{\includegraphics[width=2.1cm]{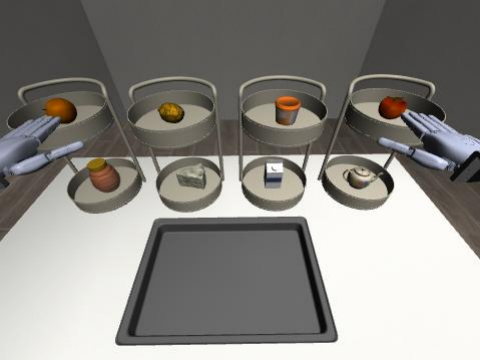}} &
\textbf{13.~\texttt{take\_\allowbreak jam\_\allowbreak and\_\allowbreak lemon\_\allowbreak from\_\allowbreak shelf\_\allowbreak into\_\allowbreak tray}}\hfill {\footnotesize $K{=}2$ subtasks $\cdot$ 30 demos}\newline
{\small\emph{Instruction:} Pick up the jam and the lemon from the shelf and put them into the tray.}\newline
{\footnotesize\textbf{Scoring:} Put the jam into the tray \textbf{(50)}$^{\dagger}$ $\cdot$ Put the lemon into the tray \textbf{(50)}$^{\dagger}$.}
\end{tabular}}

\par\vspace{3pt}\noindent
{\renewcommand{\arraystretch}{1.0}%
\begin{tabular}{@{}>{\columncolor{catif}}p{2.2pt}@{\hspace{5pt}}m{2.1cm}@{\hspace{6pt}}p{\dimexpr\linewidth-2.2pt-5pt-2.1cm-6pt\relax}@{}}
& \raisebox{-5pt}{\includegraphics[width=2.1cm]{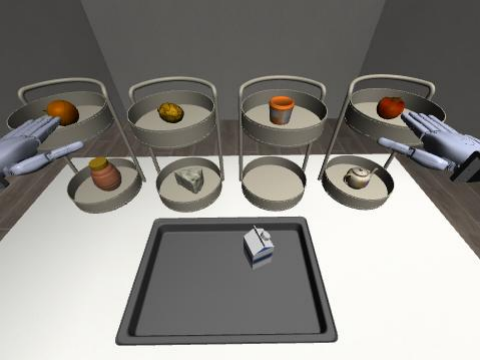}} &
\textbf{14.~\texttt{take\_\allowbreak jam\_\allowbreak into\_\allowbreak tray\_\allowbreak and\_\allowbreak put\_\allowbreak milk\_\allowbreak on\_\allowbreak shelf}}\hfill {\footnotesize $K{=}2$ subtasks $\cdot$ 30 demos}\newline
{\small\emph{Instruction:} Pick up the jam from the shelf and put it into the tray, and take the milk carton out of the tray and place it back onto the shelf.}\newline
{\footnotesize\textbf{Scoring:} Put the jam into the tray \textbf{(50)}$^{\dagger}$ $\cdot$ Put the milk carton back onto the shelf \textbf{(50)}$^{\dagger}$.}
\end{tabular}}

\par\vspace{3pt}\noindent
{\renewcommand{\arraystretch}{1.0}%
\begin{tabular}{@{}>{\columncolor{catif}}p{2.2pt}@{\hspace{5pt}}m{2.1cm}@{\hspace{6pt}}p{\dimexpr\linewidth-2.2pt-5pt-2.1cm-6pt\relax}@{}}
& \raisebox{-5pt}{\includegraphics[width=2.1cm]{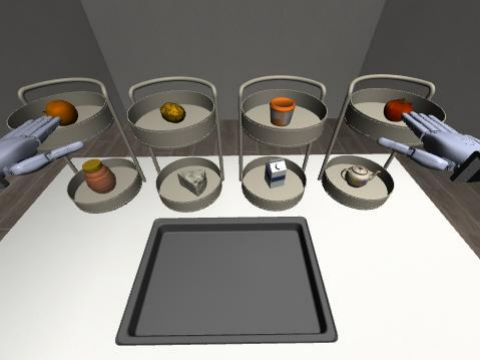}} &
\textbf{15.~\texttt{take\_\allowbreak milk\_\allowbreak and\_\allowbreak orange\_\allowbreak from\_\allowbreak shelf\_\allowbreak into\_\allowbreak tray}}\hfill {\footnotesize $K{=}2$ subtasks $\cdot$ 30 demos}\newline
{\small\emph{Instruction:} Pick up the milk carton and the orange from the shelf and put them into the tray.}\newline
{\footnotesize\textbf{Scoring:} Put the milk carton into the tray \textbf{(50)}$^{\dagger}$ $\cdot$ Put the orange into the tray \textbf{(50)}$^{\dagger}$.}
\end{tabular}}

\par\vspace{3pt}\noindent
{\renewcommand{\arraystretch}{1.0}%
\begin{tabular}{@{}>{\columncolor{catif}}p{2.2pt}@{\hspace{5pt}}m{2.1cm}@{\hspace{6pt}}p{\dimexpr\linewidth-2.2pt-5pt-2.1cm-6pt\relax}@{}}
& \raisebox{-5pt}{\includegraphics[width=2.1cm]{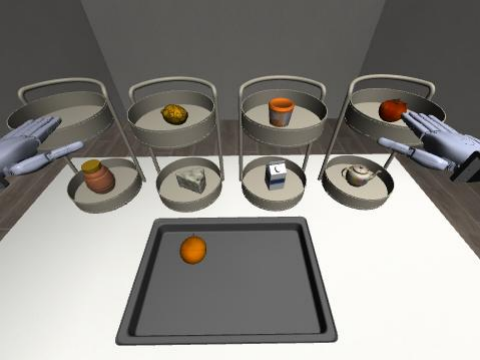}} &
\textbf{16.~\texttt{take\_\allowbreak teapot\_\allowbreak into\_\allowbreak tray\_\allowbreak and\_\allowbreak put\_\allowbreak orange\_\allowbreak on\_\allowbreak shelf}}\hfill {\footnotesize $K{=}2$ subtasks $\cdot$ 30 demos}\newline
{\small\emph{Instruction:} Pick up the teapot from the shelf and put it into the tray, and take the orange out of the tray and place it onto the shelf.}\newline
{\footnotesize\textbf{Scoring:} Put the teapot into the tray \textbf{(50)}$^{\dagger}$ $\cdot$ Put the orange onto the shelf \textbf{(50)}$^{\dagger}$.}
\end{tabular}}

\par\vspace{3pt}\noindent
{\renewcommand{\arraystretch}{1.0}%
\begin{tabular}{@{}>{\columncolor{catif}}p{2.2pt}@{\hspace{5pt}}m{2.1cm}@{\hspace{6pt}}p{\dimexpr\linewidth-2.2pt-5pt-2.1cm-6pt\relax}@{}}
& \raisebox{-5pt}{\includegraphics[width=2.1cm]{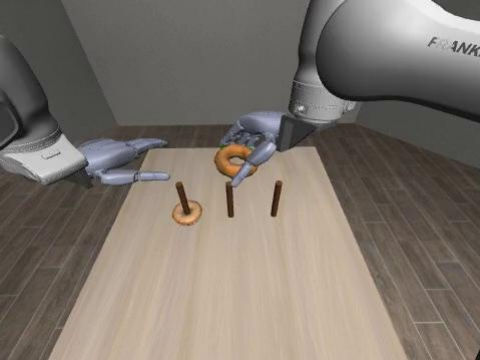}} &
\textbf{17.~\texttt{thread\_\allowbreak bagel\_\allowbreak and\_\allowbreak donut\_\allowbreak onto\_\allowbreak middle\_\allowbreak and\_\allowbreak side\_\allowbreak rods}}\hfill {\footnotesize $K{=}4$ subtasks $\cdot$ 100 demos}\newline
{\small\emph{Instruction:} Thread the bagel onto the middle rod and the donut onto the left rod.}\newline
{\footnotesize\textbf{Scoring:} Left arm picks up the donut \textbf{(10)} $\cdot$ Left arm threads the donut onto the left rod \textbf{(40)}$^{\dagger}$ $\cdot$ Right arm picks up the bagel \textbf{(10)} $\cdot$ Right arm threads the bagel onto the middle rod \textbf{(40)}$^{\dagger}$.}
\end{tabular}}

\par\vspace{3pt}\noindent
{\renewcommand{\arraystretch}{1.0}%
\begin{tabular}{@{}>{\columncolor{catif}}p{2.2pt}@{\hspace{5pt}}m{2.1cm}@{\hspace{6pt}}p{\dimexpr\linewidth-2.2pt-5pt-2.1cm-6pt\relax}@{}}
& \raisebox{-5pt}{\includegraphics[width=2.1cm]{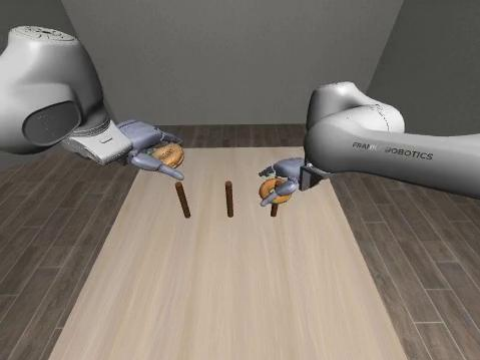}} &
\textbf{18.~\texttt{thread\_\allowbreak bagel\_\allowbreak and\_\allowbreak donut\_\allowbreak onto\_\allowbreak side\_\allowbreak rods}}\hfill {\footnotesize $K{=}4$ subtasks $\cdot$ 100 demos}\newline
{\small\emph{Instruction:} Thread the bagel onto the right rod and the donut onto the left rod.}\newline
{\footnotesize\textbf{Scoring:} Right arm picks up the donut \textbf{(10)} $\cdot$ Right arm threads the donut onto the left rod \textbf{(40)}$^{\dagger}$ $\cdot$ Left arm picks up the bagel \textbf{(10)} $\cdot$ Left arm threads the bagel onto the right rod \textbf{(40)}$^{\dagger}$.}
\end{tabular}}

\par\vspace{3pt}\noindent
{\renewcommand{\arraystretch}{1.0}%
\begin{tabular}{@{}>{\columncolor{catif}}p{2.2pt}@{\hspace{5pt}}m{2.1cm}@{\hspace{6pt}}p{\dimexpr\linewidth-2.2pt-5pt-2.1cm-6pt\relax}@{}}
& \raisebox{-5pt}{\includegraphics[width=2.1cm]{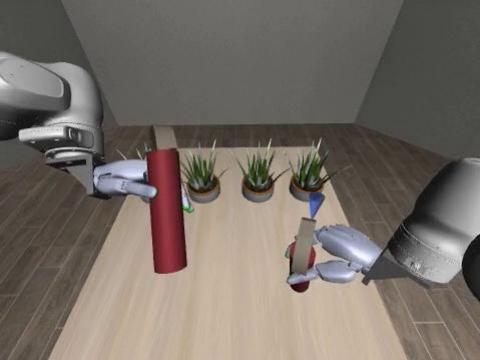}} &
\textbf{19.~\texttt{water\_\allowbreak one\_\allowbreak plant\_\allowbreak each\_\allowbreak hand}}\hfill {\footnotesize $K{=}4$ subtasks $\cdot$ 100 demos}\newline
{\small\emph{Instruction:} Use the left watering can to water the leftmost plant and the right watering can to water the rightmost plant.}\newline
{\footnotesize\textbf{Scoring:} Right arm waters the rightmost plant \textbf{(25)} $\cdot$ Left arm waters the leftmost plant \textbf{(25)} $\cdot$ Extra plant\_1 is not watered \textbf{(25)}$^{\ddagger}$ $\cdot$ Extra plant\_2 is not watered \textbf{(25)}$^{\ddagger}$.}
\end{tabular}}

\par\vspace{3pt}\noindent
{\renewcommand{\arraystretch}{1.0}%
\begin{tabular}{@{}>{\columncolor{catif}}p{2.2pt}@{\hspace{5pt}}m{2.1cm}@{\hspace{6pt}}p{\dimexpr\linewidth-2.2pt-5pt-2.1cm-6pt\relax}@{}}
& \raisebox{-5pt}{\includegraphics[width=2.1cm]{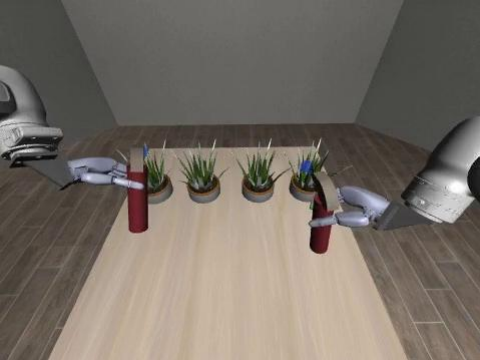}} &
\textbf{20.~\texttt{water\_\allowbreak two\_\allowbreak plants\_\allowbreak each\_\allowbreak hand}}\hfill {\footnotesize $K{=}4$ subtasks $\cdot$ 100 demos}\newline
{\small\emph{Instruction:} Use the left watering can to water the two left plants and the right watering can to water the two right plants.}\newline
{\footnotesize\textbf{Scoring:} Right arm waters the first right plant \textbf{(25)} $\cdot$ Right arm waters the second right plant \textbf{(25)} $\cdot$ Left arm waters the first left plant \textbf{(25)} $\cdot$ Left arm waters the second left plant \textbf{(25)}.}
\end{tabular}}

\subsection{Limited-Demonstrations (26 specifications, 26 with released demonstrations)}
\label{app:tasks:limited_demonstrations}

Data-scarce skill acquisition: every task ships with exactly 30 demonstrations. The axis includes ten articulated-mechanism tasks, such as opening a microwave, cash register, or rice cooker, which require force-aware actuation of hinged, sliding, and rotary joints. This regime makes 44-DoF hand coordination decisive.

\par\vspace{3pt}\noindent
{\renewcommand{\arraystretch}{1.0}%
\begin{tabular}{@{}>{\columncolor{catld}}p{2.2pt}@{\hspace{5pt}}m{2.1cm}@{\hspace{6pt}}p{\dimexpr\linewidth-2.2pt-5pt-2.1cm-6pt\relax}@{}}
& \raisebox{-5pt}{\includegraphics[width=2.1cm]{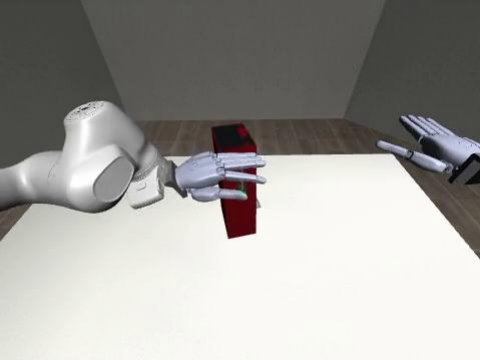}} &
\textbf{1.~\texttt{pull\_\allowbreak down\_\allowbreak slot\_\allowbreak machine\_\allowbreak lever}}\hfill {\footnotesize $K{=}1$ subtask $\cdot$ 30 demos}\newline
{\small\emph{Instruction:} Pull down the lever of the slot machine.}\newline
{\footnotesize\textbf{Scoring:} Pull down the lever of the slot machine \textbf{(100)}.}
\end{tabular}}

\par\vspace{3pt}\noindent
{\renewcommand{\arraystretch}{1.0}%
\begin{tabular}{@{}>{\columncolor{catld}}p{2.2pt}@{\hspace{5pt}}m{2.1cm}@{\hspace{6pt}}p{\dimexpr\linewidth-2.2pt-5pt-2.1cm-6pt\relax}@{}}
& \raisebox{-5pt}{\includegraphics[width=2.1cm]{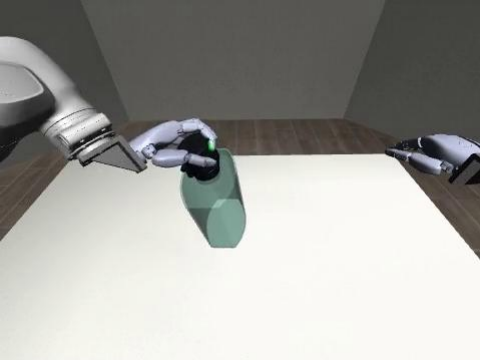}} &
\textbf{2.~\texttt{flip\_\allowbreak open\_\allowbreak bottle\_\allowbreak cap}}\hfill {\footnotesize $K{=}1$ subtask $\cdot$ 30 demos}\newline
{\small\emph{Instruction:} Flip open the bottle's articulated cap.}\newline
{\footnotesize\textbf{Scoring:} Flip open the bottle's articulated cap \textbf{(100)}.}
\end{tabular}}

\par\vspace{3pt}\noindent
{\renewcommand{\arraystretch}{1.0}%
\begin{tabular}{@{}>{\columncolor{catld}}p{2.2pt}@{\hspace{5pt}}m{2.1cm}@{\hspace{6pt}}p{\dimexpr\linewidth-2.2pt-5pt-2.1cm-6pt\relax}@{}}
& \raisebox{-5pt}{\includegraphics[width=2.1cm]{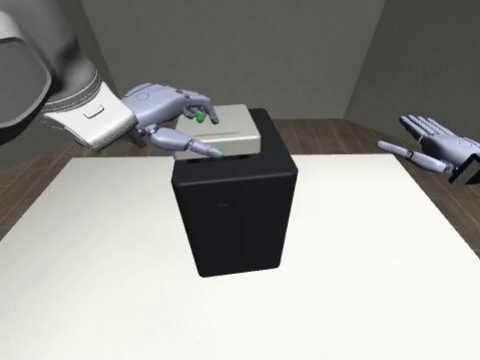}} &
\textbf{3.~\texttt{open\_\allowbreak bread\_\allowbreak maker\_\allowbreak lid}}\hfill {\footnotesize $K{=}1$ subtask $\cdot$ 30 demos}\newline
{\small\emph{Instruction:} Open the lid of the bread maker.}\newline
{\footnotesize\textbf{Scoring:} Open the lid of the bread maker \textbf{(100)}.}
\end{tabular}}

\par\vspace{3pt}\noindent
{\renewcommand{\arraystretch}{1.0}%
\begin{tabular}{@{}>{\columncolor{catld}}p{2.2pt}@{\hspace{5pt}}m{2.1cm}@{\hspace{6pt}}p{\dimexpr\linewidth-2.2pt-5pt-2.1cm-6pt\relax}@{}}
& \raisebox{-5pt}{\includegraphics[width=2.1cm]{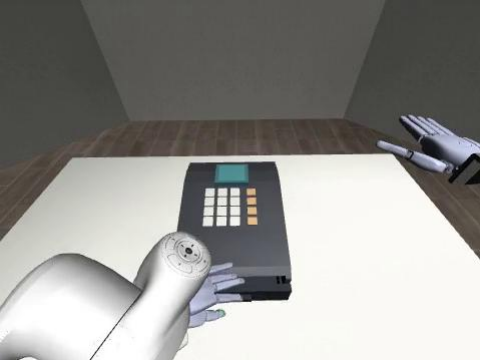}} &
\textbf{4.~\texttt{pull\_\allowbreak open\_\allowbreak cash\_\allowbreak register\_\allowbreak drawer}}\hfill {\footnotesize $K{=}1$ subtask $\cdot$ 30 demos}\newline
{\small\emph{Instruction:} Pull open the cash register drawer.}\newline
{\footnotesize\textbf{Scoring:} Pull open the cash register drawer \textbf{(100)}.}
\end{tabular}}

\par\vspace{3pt}\noindent
{\renewcommand{\arraystretch}{1.0}%
\begin{tabular}{@{}>{\columncolor{catld}}p{2.2pt}@{\hspace{5pt}}m{2.1cm}@{\hspace{6pt}}p{\dimexpr\linewidth-2.2pt-5pt-2.1cm-6pt\relax}@{}}
& \raisebox{-5pt}{\includegraphics[width=2.1cm]{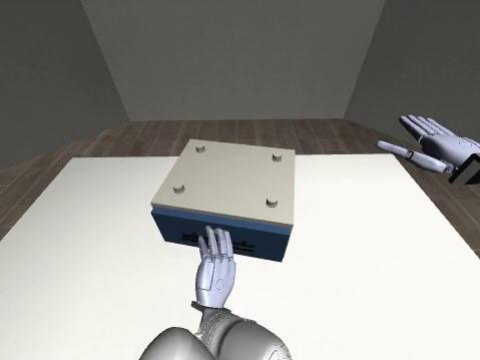}} &
\textbf{5.~\texttt{pull\_\allowbreak open\_\allowbreak cabinet\_\allowbreak drawer}}\hfill {\footnotesize $K{=}1$ subtask $\cdot$ 30 demos}\newline
{\small\emph{Instruction:} Pull open the drawer of the cabinet.}\newline
{\footnotesize\textbf{Scoring:} Pull open the drawer of the cabinet \textbf{(100)}.}
\end{tabular}}

\par\vspace{3pt}\noindent
{\renewcommand{\arraystretch}{1.0}%
\begin{tabular}{@{}>{\columncolor{catld}}p{2.2pt}@{\hspace{5pt}}m{2.1cm}@{\hspace{6pt}}p{\dimexpr\linewidth-2.2pt-5pt-2.1cm-6pt\relax}@{}}
& \raisebox{-5pt}{\includegraphics[width=2.1cm]{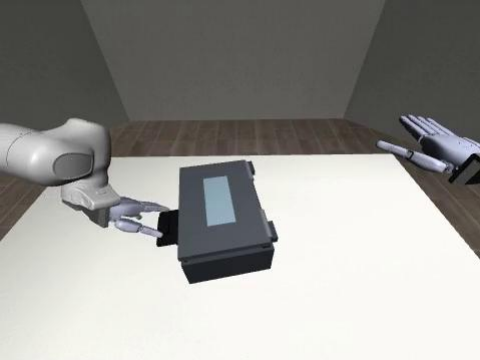}} &
\textbf{6.~\texttt{open\_\allowbreak flatbed\_\allowbreak scanner\_\allowbreak lid}}\hfill {\footnotesize $K{=}1$ subtask $\cdot$ 30 demos}\newline
{\small\emph{Instruction:} Open the lid of the flatbed scanner.}\newline
{\footnotesize\textbf{Scoring:} Open the lid of the flatbed scanner \textbf{(100)}.}
\end{tabular}}

\par\vspace{3pt}\noindent
{\renewcommand{\arraystretch}{1.0}%
\begin{tabular}{@{}>{\columncolor{catld}}p{2.2pt}@{\hspace{5pt}}m{2.1cm}@{\hspace{6pt}}p{\dimexpr\linewidth-2.2pt-5pt-2.1cm-6pt\relax}@{}}
& \raisebox{-5pt}{\includegraphics[width=2.1cm]{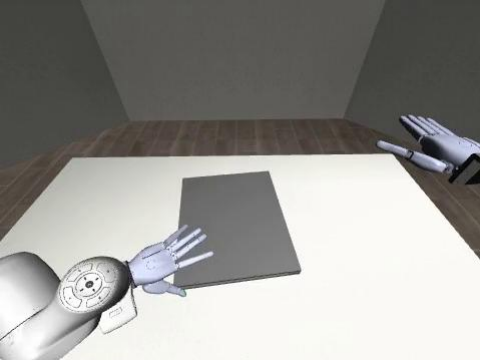}} &
\textbf{7.~\texttt{raise\_\allowbreak laptop\_\allowbreak stand\_\allowbreak platform}}\hfill {\footnotesize $K{=}1$ subtask $\cdot$ 30 demos}\newline
{\small\emph{Instruction:} Raise the platform of the laptop stand.}\newline
{\footnotesize\textbf{Scoring:} Raise the platform of the laptop stand \textbf{(100)}.}
\end{tabular}}

\par\vspace{3pt}\noindent
{\renewcommand{\arraystretch}{1.0}%
\begin{tabular}{@{}>{\columncolor{catld}}p{2.2pt}@{\hspace{5pt}}m{2.1cm}@{\hspace{6pt}}p{\dimexpr\linewidth-2.2pt-5pt-2.1cm-6pt\relax}@{}}
& \raisebox{-5pt}{\includegraphics[width=2.1cm]{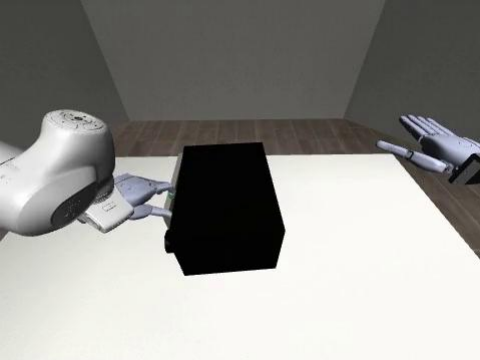}} &
\textbf{8.~\texttt{open\_\allowbreak microwave\_\allowbreak oven\_\allowbreak door}}\hfill {\footnotesize $K{=}1$ subtask $\cdot$ 30 demos}\newline
{\small\emph{Instruction:} Open the microwave oven door.}\newline
{\footnotesize\textbf{Scoring:} Open the microwave oven door \textbf{(100)}.}
\end{tabular}}

\par\vspace{3pt}\noindent
{\renewcommand{\arraystretch}{1.0}%
\begin{tabular}{@{}>{\columncolor{catld}}p{2.2pt}@{\hspace{5pt}}m{2.1cm}@{\hspace{6pt}}p{\dimexpr\linewidth-2.2pt-5pt-2.1cm-6pt\relax}@{}}
& \raisebox{-5pt}{\includegraphics[width=2.1cm]{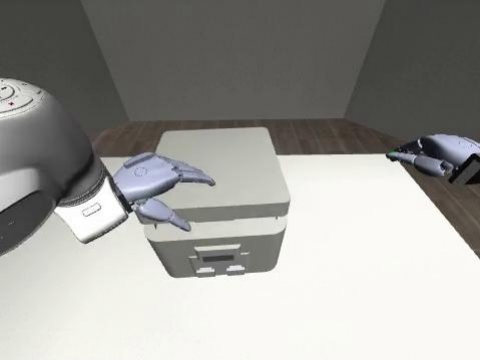}} &
\textbf{9.~\texttt{open\_\allowbreak rice\_\allowbreak cooker\_\allowbreak lid}}\hfill {\footnotesize $K{=}1$ subtask $\cdot$ 30 demos}\newline
{\small\emph{Instruction:} Open the lid of the rice cooker.}\newline
{\footnotesize\textbf{Scoring:} Open the lid of the rice cooker \textbf{(100)}.}
\end{tabular}}

\par\vspace{3pt}\noindent
{\renewcommand{\arraystretch}{1.0}%
\begin{tabular}{@{}>{\columncolor{catld}}p{2.2pt}@{\hspace{5pt}}m{2.1cm}@{\hspace{6pt}}p{\dimexpr\linewidth-2.2pt-5pt-2.1cm-6pt\relax}@{}}
& \raisebox{-5pt}{\includegraphics[width=2.1cm]{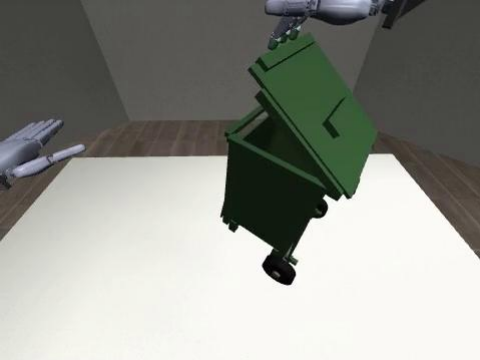}} &
\textbf{10.~\texttt{open\_\allowbreak wheelie\_\allowbreak bin\_\allowbreak lid}}\hfill {\footnotesize $K{=}1$ subtask $\cdot$ 30 demos}\newline
{\small\emph{Instruction:} Open the lid of the wheelie bin.}\newline
{\footnotesize\textbf{Scoring:} Open the lid of the wheelie bin \textbf{(100)}.}
\end{tabular}}

\par\vspace{3pt}\noindent
{\renewcommand{\arraystretch}{1.0}%
\begin{tabular}{@{}>{\columncolor{catld}}p{2.2pt}@{\hspace{5pt}}m{2.1cm}@{\hspace{6pt}}p{\dimexpr\linewidth-2.2pt-5pt-2.1cm-6pt\relax}@{}}
& \raisebox{-5pt}{\includegraphics[width=2.1cm]{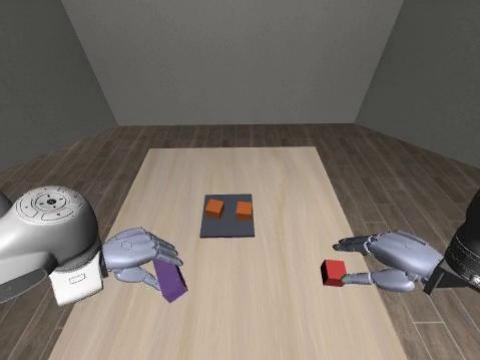}} &
\textbf{11.~\texttt{assemble\_\allowbreak arch\_\allowbreak bridge}}\hfill {\footnotesize $K{=}6$ subtasks $\cdot$ 30 demos}\newline
{\small\emph{Instruction:} Stand two piers on the base, lay the deck across them, then set the capstone cube on the middle of the deck.}\newline
{\footnotesize\textbf{Scoring:} Stand pier A on the base \textbf{(10)}$^{\dagger}$ $\cdot$ Stand pier B on the base \textbf{(10)}$^{\dagger}$ $\cdot$ Pick up the deck \textbf{(20)} $\cdot$ Lay the deck across the two piers \textbf{(20)}$^{\dagger}$ $\cdot$ Pick up the capstone \textbf{(20)} $\cdot$ Set the capstone on the deck centre \textbf{(20)}$^{\dagger}$.}
\end{tabular}}

\par\vspace{3pt}\noindent
{\renewcommand{\arraystretch}{1.0}%
\begin{tabular}{@{}>{\columncolor{catld}}p{2.2pt}@{\hspace{5pt}}m{2.1cm}@{\hspace{6pt}}p{\dimexpr\linewidth-2.2pt-5pt-2.1cm-6pt\relax}@{}}
& \raisebox{-5pt}{\includegraphics[width=2.1cm]{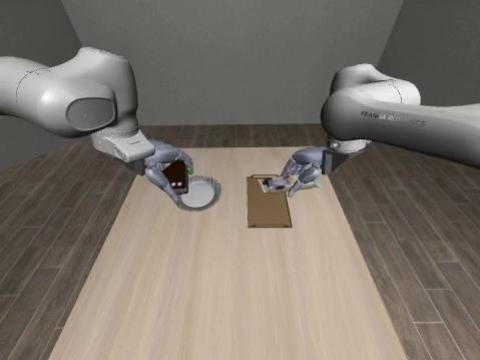}} &
\textbf{12.~\texttt{arrange\_\allowbreak bar\_\allowbreak on\_\allowbreak tray\_\allowbreak and\_\allowbreak cake\_\allowbreak on\_\allowbreak plate}}\hfill {\footnotesize $K{=}4$ subtasks $\cdot$ 30 demos}\newline
{\small\emph{Instruction:} Place the food bar on the tray, then place the cake on the plate.}\newline
{\footnotesize\textbf{Scoring:} Pick up the food bar \textbf{(20)} $\cdot$ Place the food bar on the tray \textbf{(30)}$^{\dagger}$ $\cdot$ Pick up the cake \textbf{(20)} $\cdot$ Place the cake on the plate \textbf{(30)}$^{\dagger}$.}
\end{tabular}}

\par\vspace{3pt}\noindent
{\renewcommand{\arraystretch}{1.0}%
\begin{tabular}{@{}>{\columncolor{catld}}p{2.2pt}@{\hspace{5pt}}m{2.1cm}@{\hspace{6pt}}p{\dimexpr\linewidth-2.2pt-5pt-2.1cm-6pt\relax}@{}}
& \raisebox{-5pt}{\includegraphics[width=2.1cm]{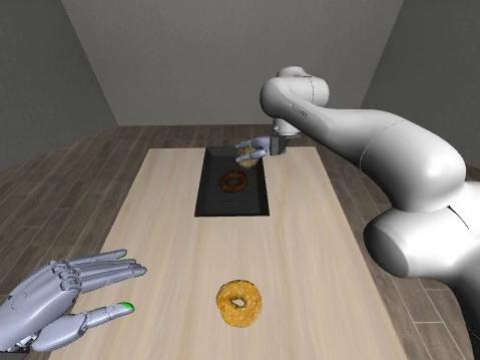}} &
\textbf{13.~\texttt{arrange\_\allowbreak three\_\allowbreak bagels\_\allowbreak in\_\allowbreak oven\_\allowbreak tray}}\hfill {\footnotesize $K{=}6$ subtasks $\cdot$ 30 demos}\newline
{\small\emph{Instruction:} Pick up the three bagels one at a time and lay each flat in the oven tray so it rests within the tray.}\newline
{\footnotesize\textbf{Scoring:} Pick up the first bagel \textbf{(10)} $\cdot$ Lay the first bagel in the oven tray \textbf{(20)}$^{\dagger}$ $\cdot$ Pick up the second bagel \textbf{(10)} $\cdot$ Lay the second bagel in the oven tray \textbf{(25)}$^{\dagger}$ $\cdot$ Pick up the third bagel \textbf{(10)} $\cdot$ Lay the third bagel in the oven tray \textbf{(25)}$^{\dagger}$.}
\end{tabular}}

\par\vspace{3pt}\noindent
{\renewcommand{\arraystretch}{1.0}%
\begin{tabular}{@{}>{\columncolor{catld}}p{2.2pt}@{\hspace{5pt}}m{2.1cm}@{\hspace{6pt}}p{\dimexpr\linewidth-2.2pt-5pt-2.1cm-6pt\relax}@{}}
& \raisebox{-5pt}{\includegraphics[width=2.1cm]{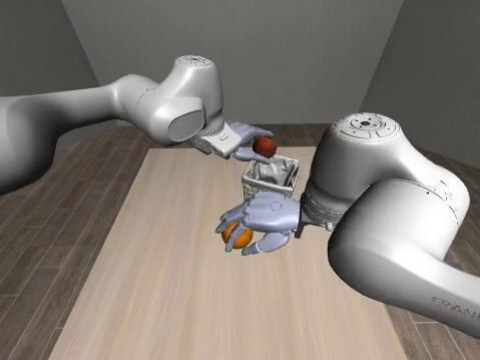}} &
\textbf{14.~\texttt{put\_\allowbreak apple\_\allowbreak and\_\allowbreak orange\_\allowbreak into\_\allowbreak basket}}\hfill {\footnotesize $K{=}4$ subtasks $\cdot$ 30 demos}\newline
{\small\emph{Instruction:} Pick up the apple and the orange and put them both inside the basket.}\newline
{\footnotesize\textbf{Scoring:} Pick up the apple \textbf{(25)} $\cdot$ Put the apple inside the basket \textbf{(25)}$^{\dagger}$ $\cdot$ Pick up the orange \textbf{(25)} $\cdot$ Put the orange inside the basket \textbf{(25)}$^{\dagger}$.}
\end{tabular}}

\par\vspace{3pt}\noindent
{\renewcommand{\arraystretch}{1.0}%
\begin{tabular}{@{}>{\columncolor{catld}}p{2.2pt}@{\hspace{5pt}}m{2.1cm}@{\hspace{6pt}}p{\dimexpr\linewidth-2.2pt-5pt-2.1cm-6pt\relax}@{}}
& \raisebox{-5pt}{\includegraphics[width=2.1cm]{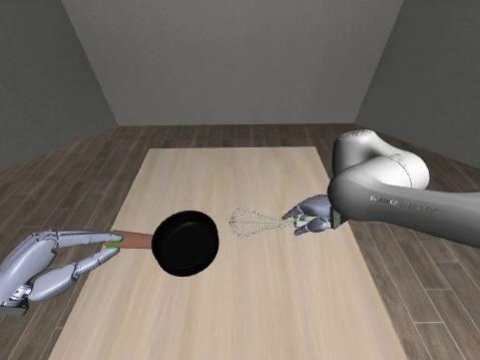}} &
\textbf{15.~\texttt{pick\_\allowbreak up\_\allowbreak whisk}}\hfill {\footnotesize $K{=}1$ subtask $\cdot$ 30 demos}\newline
{\small\emph{Instruction:} Pick up the whisk.}\newline
{\footnotesize\textbf{Scoring:} Pick up the whisk \textbf{(100)}$^{\dagger}$.}
\end{tabular}}

\par\vspace{3pt}\noindent
{\renewcommand{\arraystretch}{1.0}%
\begin{tabular}{@{}>{\columncolor{catld}}p{2.2pt}@{\hspace{5pt}}m{2.1cm}@{\hspace{6pt}}p{\dimexpr\linewidth-2.2pt-5pt-2.1cm-6pt\relax}@{}}
& \raisebox{-5pt}{\includegraphics[width=2.1cm]{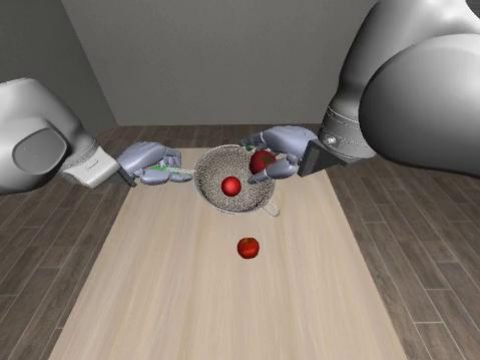}} &
\textbf{16.~\texttt{drain\_\allowbreak three\_\allowbreak tomatoes\_\allowbreak in\_\allowbreak strainer}}\hfill {\footnotesize $K{=}7$ subtasks $\cdot$ 30 demos}\newline
{\small\emph{Instruction:} Pick up the three tomatoes and place them into the strainer so they rest in it.}\newline
{\footnotesize\textbf{Scoring:} Pick up the strainer \textbf{(10)}$^{\dagger}$ $\cdot$ Pick up the first tomato \textbf{(10)} $\cdot$ Place the first tomato into the strainer \textbf{(20)}$^{\dagger}$ $\cdot$ Pick up the second tomato \textbf{(10)} $\cdot$ Place the second tomato into the strainer \textbf{(20)}$^{\dagger}$ $\cdot$ Pick up the third tomato \textbf{(10)} $\cdot$ Place the third tomato into the strainer \textbf{(20)}$^{\dagger}$.}
\end{tabular}}

\par\vspace{3pt}\noindent
{\renewcommand{\arraystretch}{1.0}%
\begin{tabular}{@{}>{\columncolor{catld}}p{2.2pt}@{\hspace{5pt}}m{2.1cm}@{\hspace{6pt}}p{\dimexpr\linewidth-2.2pt-5pt-2.1cm-6pt\relax}@{}}
& \raisebox{-5pt}{\includegraphics[width=2.1cm]{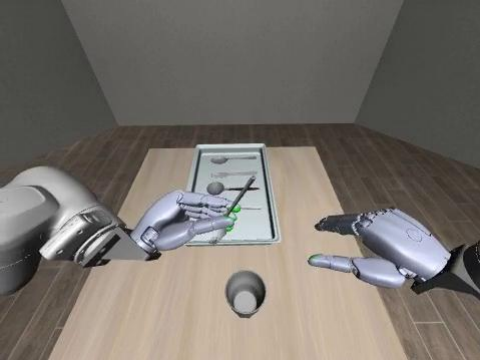}} &
\textbf{17.~\texttt{pick\_\allowbreak up\_\allowbreak fork\_\allowbreak from\_\allowbreak tray\_\allowbreak into\_\allowbreak cup}}\hfill {\footnotesize $K{=}2$ subtasks $\cdot$ 30 demos}\newline
{\small\emph{Instruction:} From the tray of mixed utensils, pick up the one fork and drop it into the cup.}\newline
{\footnotesize\textbf{Scoring:} Pick up the target utensil and drop it into the cup \textbf{(100)}$^{\dagger}$ $\cdot$ Another utensil is not picked up \textbf{(0)}$^{\ddagger}$.}
\end{tabular}}

\par\vspace{3pt}\noindent
{\renewcommand{\arraystretch}{1.0}%
\begin{tabular}{@{}>{\columncolor{catld}}p{2.2pt}@{\hspace{5pt}}m{2.1cm}@{\hspace{6pt}}p{\dimexpr\linewidth-2.2pt-5pt-2.1cm-6pt\relax}@{}}
& \raisebox{-5pt}{\includegraphics[width=2.1cm]{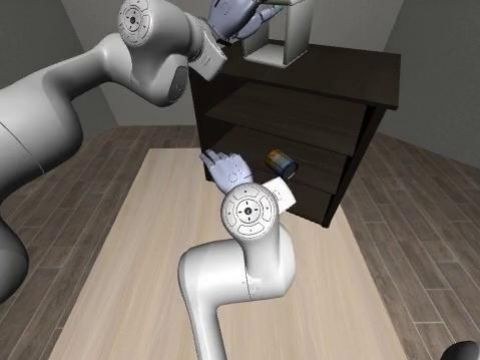}} &
\textbf{18.~\texttt{insert\_\allowbreak items\_\allowbreak into\_\allowbreak shelf}}\hfill {\footnotesize $K{=}6$ subtasks $\cdot$ 30 demos}\newline
{\small\emph{Instruction:} Pick up the alphabet soup and the BBQ sauce and insert them into the wooden shelf, then place the white storage box on the shelf.}\newline
{\footnotesize\textbf{Scoring:} Pick up the alphabet soup \textbf{(15)} $\cdot$ Insert the alphabet soup into the shelf \textbf{(20)}$^{\dagger}$ $\cdot$ Pick up the BBQ sauce \textbf{(15)} $\cdot$ Insert the BBQ sauce into the shelf \textbf{(20)}$^{\dagger}$ $\cdot$ Pick up the storage box \textbf{(15)} $\cdot$ Place the storage box on the shelf \textbf{(15)}$^{\dagger}$.}
\end{tabular}}

\par\vspace{3pt}\noindent
{\renewcommand{\arraystretch}{1.0}%
\begin{tabular}{@{}>{\columncolor{catld}}p{2.2pt}@{\hspace{5pt}}m{2.1cm}@{\hspace{6pt}}p{\dimexpr\linewidth-2.2pt-5pt-2.1cm-6pt\relax}@{}}
& \raisebox{-5pt}{\includegraphics[width=2.1cm]{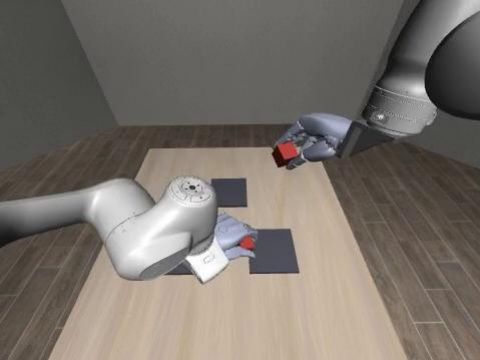}} &
\textbf{19.~\texttt{rotate\_\allowbreak two\_\allowbreak cubes\_\allowbreak around\_\allowbreak plates}}\hfill {\footnotesize $K{=}5$ subtasks $\cdot$ 30 demos}\newline
{\small\emph{Instruction:} Move the two cubes clockwise around the three triangle plates one move at a time, always moving the cube before the empty plate onto it, until each returns to its start.}\newline
{\footnotesize\textbf{Scoring:} Right arm moves the cube before the empty plate onto it \textbf{(50)} $\to$ Left arm moves the cube before the empty plate onto it \textbf{(30)} $\to$ Left arm moves the last cube back to its start \textbf{(20)}$^{\dagger}$ $\to$ Second cube restored on its own plate \textbf{(0)}$^{\dagger}$ $\to$ The plate that started empty is empty again \textbf{(0)}$^{\dagger}$.}
\end{tabular}}

\par\vspace{3pt}\noindent
{\renewcommand{\arraystretch}{1.0}%
\begin{tabular}{@{}>{\columncolor{catld}}p{2.2pt}@{\hspace{5pt}}m{2.1cm}@{\hspace{6pt}}p{\dimexpr\linewidth-2.2pt-5pt-2.1cm-6pt\relax}@{}}
& \raisebox{-5pt}{\includegraphics[width=2.1cm]{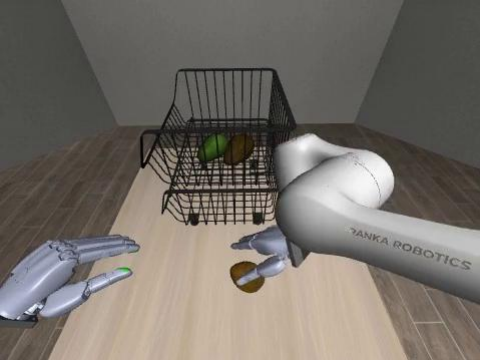}} &
\textbf{20.~\texttt{pack\_\allowbreak three\_\allowbreak avocados\_\allowbreak into\_\allowbreak tiered\_\allowbreak basket}}\hfill {\footnotesize $K{=}6$ subtasks $\cdot$ 30 demos}\newline
{\small\emph{Instruction:} Pick up the three avocados one at a time and drop them into the tiered basket so they rest in it.}\newline
{\footnotesize\textbf{Scoring:} Pick up the first avocado \textbf{(10)} $\cdot$ Place the first avocado into the basket \textbf{(20)}$^{\dagger}$ $\cdot$ Pick up the second avocado \textbf{(10)} $\cdot$ Place the second avocado into the basket \textbf{(25)}$^{\dagger}$ $\cdot$ Pick up the third avocado \textbf{(10)} $\cdot$ Place the third avocado into the basket \textbf{(25)}$^{\dagger}$.}
\end{tabular}}

\par\vspace{3pt}\noindent
{\renewcommand{\arraystretch}{1.0}%
\begin{tabular}{@{}>{\columncolor{catld}}p{2.2pt}@{\hspace{5pt}}m{2.1cm}@{\hspace{6pt}}p{\dimexpr\linewidth-2.2pt-5pt-2.1cm-6pt\relax}@{}}
& \raisebox{-5pt}{\includegraphics[width=2.1cm]{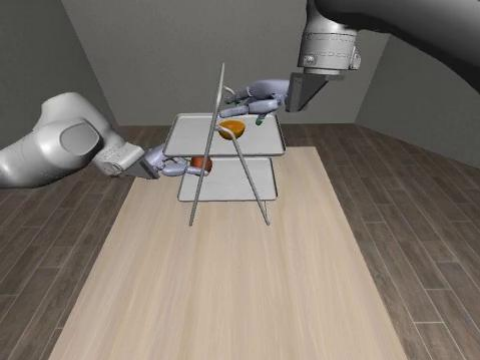}} &
\textbf{21.~\texttt{place\_\allowbreak two\_\allowbreak oranges\_\allowbreak on\_\allowbreak shelf\_\allowbreak tiers}}\hfill {\footnotesize $K{=}2$ subtasks $\cdot$ 30 demos}\newline
{\small\emph{Instruction:} Pick up the two oranges and place one on the lower tier of the shelf and the other on the upper tier.}\newline
{\footnotesize\textbf{Scoring:} Place one orange on the upper tier of the shelf \textbf{(40)}$^{\dagger}$ $\cdot$ Place the other orange on the lower tier of the shelf \textbf{(60)}$^{\dagger}$.}
\end{tabular}}

\par\vspace{3pt}\noindent
{\renewcommand{\arraystretch}{1.0}%
\begin{tabular}{@{}>{\columncolor{catld}}p{2.2pt}@{\hspace{5pt}}m{2.1cm}@{\hspace{6pt}}p{\dimexpr\linewidth-2.2pt-5pt-2.1cm-6pt\relax}@{}}
& \raisebox{-5pt}{\includegraphics[width=2.1cm]{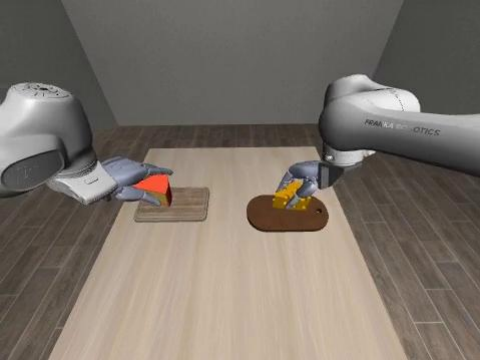}} &
\textbf{22.~\texttt{serve\_\allowbreak cheese\_\allowbreak on\_\allowbreak two\_\allowbreak cutting\_\allowbreak boards}}\hfill {\footnotesize $K{=}4$ subtasks $\cdot$ 30 demos}\newline
{\small\emph{Instruction:} Pick up a cheese block and set it flat on its cutting board, one cheese per board.}\newline
{\footnotesize\textbf{Scoring:} Pick up the first cheese block \textbf{(25)} $\cdot$ Set the first cheese block on its cutting board \textbf{(25)}$^{\dagger}$ $\cdot$ Pick up the second cheese block \textbf{(25)} $\cdot$ Set the second cheese block on its cutting board \textbf{(25)}$^{\dagger}$.}
\end{tabular}}

\par\vspace{3pt}\noindent
{\renewcommand{\arraystretch}{1.0}%
\begin{tabular}{@{}>{\columncolor{catld}}p{2.2pt}@{\hspace{5pt}}m{2.1cm}@{\hspace{6pt}}p{\dimexpr\linewidth-2.2pt-5pt-2.1cm-6pt\relax}@{}}
& \raisebox{-5pt}{\includegraphics[width=2.1cm]{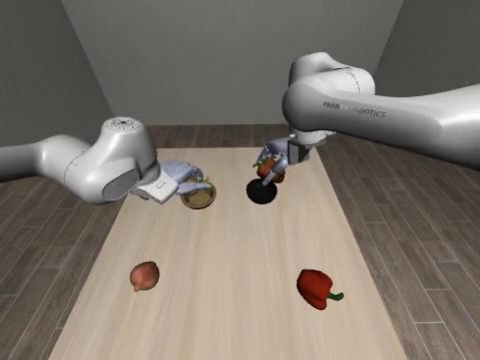}} &
\textbf{23.~\texttt{sort\_\allowbreak pepper\_\allowbreak and\_\allowbreak onion\_\allowbreak into\_\allowbreak two\_\allowbreak bowls}}\hfill {\footnotesize $K{=}4$ subtasks $\cdot$ 30 demos}\newline
{\small\emph{Instruction:} Put one bell pepper into one bowl and one onion into the other, keeping each kind separate.}\newline
{\footnotesize\textbf{Scoring:} Pick up the first bell pepper \textbf{(25)} $\cdot$ Place the first bell pepper into its bowl \textbf{(25)}$^{\dagger}$ $\cdot$ Pick up the first onion \textbf{(25)} $\cdot$ Place the first onion into its bowl \textbf{(25)}$^{\dagger}$.}
\end{tabular}}

\par\vspace{3pt}\noindent
{\renewcommand{\arraystretch}{1.0}%
\begin{tabular}{@{}>{\columncolor{catld}}p{2.2pt}@{\hspace{5pt}}m{2.1cm}@{\hspace{6pt}}p{\dimexpr\linewidth-2.2pt-5pt-2.1cm-6pt\relax}@{}}
& \raisebox{-5pt}{\includegraphics[width=2.1cm]{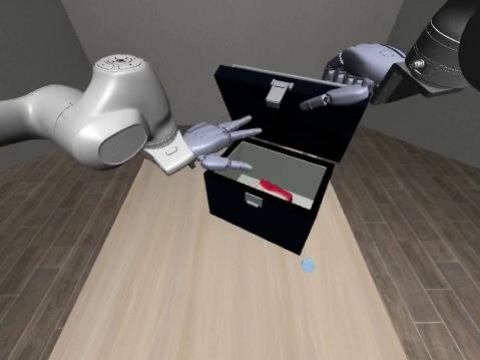}} &
\textbf{24.~\texttt{store\_\allowbreak frozen\_\allowbreak food\_\allowbreak in\_\allowbreak freezer}}\hfill {\footnotesize $K{=}6$ subtasks $\cdot$ 30 demos}\newline
{\small\emph{Instruction:} Open the chest freezer lid, put the ice cubes and boxed food inside, then close the freezer lid.}\newline
{\footnotesize\textbf{Scoring:} Open the chest freezer lid \textbf{(20)} $\to$ Pick up the ice cubes \textbf{(20)} $\to$ Put the ice cubes inside the freezer \textbf{(20)}$^{\dagger}$ $\to$ Pick up the boxed food \textbf{(15)} $\to$ Put the boxed food inside the freezer \textbf{(20)}$^{\dagger}$ $\to$ Close the freezer lid \textbf{(5)}.}
\end{tabular}}

\par\vspace{3pt}\noindent
{\renewcommand{\arraystretch}{1.0}%
\begin{tabular}{@{}>{\columncolor{catld}}p{2.2pt}@{\hspace{5pt}}m{2.1cm}@{\hspace{6pt}}p{\dimexpr\linewidth-2.2pt-5pt-2.1cm-6pt\relax}@{}}
& \raisebox{-5pt}{\includegraphics[width=2.1cm]{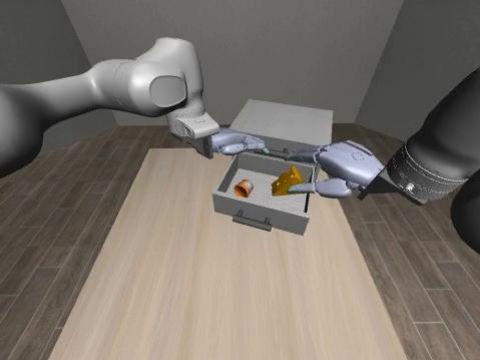}} &
\textbf{25.~\texttt{store\_\allowbreak items\_\allowbreak in\_\allowbreak white\_\allowbreak cabinet}}\hfill {\footnotesize $K{=}8$ subtasks $\cdot$ 30 demos}\newline
{\small\emph{Instruction:} Open the white cabinet middle drawer, put the cheese and the jam inside, then close the middle drawer and open the top drawer.}\newline
{\footnotesize\textbf{Scoring:} Open the white cabinet middle drawer \textbf{(20)} $\to$ Pick up the cheese \textbf{(10)} $\to$ Put the cheese inside the white cabinet \textbf{(15)}$^{\dagger}$ $\to$ Pick up the jam \textbf{(10)} $\to$ Put the jam inside the white cabinet \textbf{(15)}$^{\dagger}$ $\to$ Close the white cabinet middle drawer \textbf{(15)} $\to$ Open the white cabinet top drawer \textbf{(15)} $\to$ Close the white cabinet top drawer \textbf{(0)}.}
\end{tabular}}

\par\vspace{3pt}\noindent
{\renewcommand{\arraystretch}{1.0}%
\begin{tabular}{@{}>{\columncolor{catld}}p{2.2pt}@{\hspace{5pt}}m{2.1cm}@{\hspace{6pt}}p{\dimexpr\linewidth-2.2pt-5pt-2.1cm-6pt\relax}@{}}
& \raisebox{-5pt}{\includegraphics[width=2.1cm]{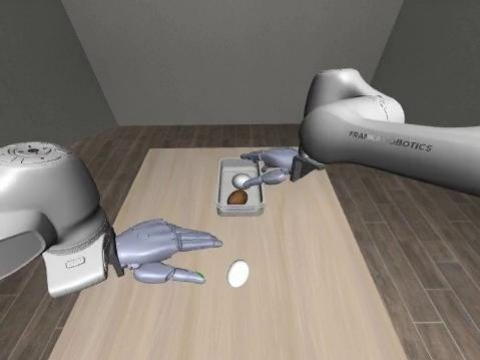}} &
\textbf{26.~\texttt{put\_\allowbreak three\_\allowbreak eggs\_\allowbreak into\_\allowbreak tupperware}}\hfill {\footnotesize $K{=}1$ subtask $\cdot$ 30 demos}\newline
{\small\emph{Instruction:} Pick up the three eggs one at a time and place them into the tupperware.}\newline
{\footnotesize\textbf{Scoring:} Place the item into the container \textbf{(100)}$^{\dagger}$.}
\end{tabular}}

\subsection{Long-Horizon (10 specifications, 10 with released demonstrations)}
\label{app:tasks:long_horizon}

Multi-stage chores, including barbecue grilling, fridge restocking, dessert plating, and recycling sorting, with 4 to 9 scored subtasks and 6.5 objects per scene on average. Their sampled demonstrations average 48.3\,s. Success requires sustaining a plan far beyond the policy's action chunk.

\par\vspace{3pt}\noindent
{\renewcommand{\arraystretch}{1.0}%
\begin{tabular}{@{}>{\columncolor{catlh}}p{2.2pt}@{\hspace{5pt}}m{2.1cm}@{\hspace{6pt}}p{\dimexpr\linewidth-2.2pt-5pt-2.1cm-6pt\relax}@{}}
& \raisebox{-5pt}{\includegraphics[width=2.1cm]{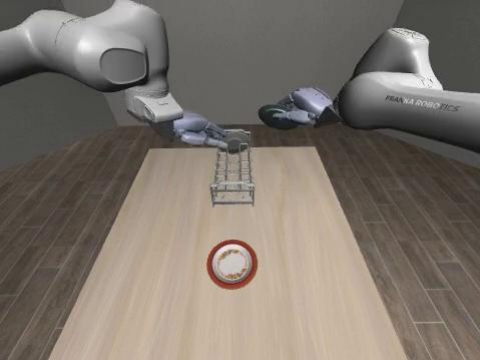}} &
\textbf{1.~\texttt{insert\_\allowbreak three\_\allowbreak plates\_\allowbreak into\_\allowbreak dish\_\allowbreak rack}}\hfill {\footnotesize $K{=}6$ subtasks $\cdot$ 100 demos}\newline
{\small\emph{Instruction:} Pick up the three plates and insert each into the dish rack to dry.}\newline
{\footnotesize\textbf{Scoring:} Pick up the first plate \textbf{(10)} $\cdot$ Insert the first plate into the dish rack \textbf{(20)}$^{\dagger}$ $\cdot$ Pick up the second plate \textbf{(10)} $\cdot$ Insert the second plate into the dish rack \textbf{(25)}$^{\dagger}$ $\cdot$ Pick up the third plate \textbf{(10)} $\cdot$ Insert the third plate into the dish rack \textbf{(25)}$^{\dagger}$.}
\end{tabular}}

\par\vspace{3pt}\noindent
{\renewcommand{\arraystretch}{1.0}%
\begin{tabular}{@{}>{\columncolor{catlh}}p{2.2pt}@{\hspace{5pt}}m{2.1cm}@{\hspace{6pt}}p{\dimexpr\linewidth-2.2pt-5pt-2.1cm-6pt\relax}@{}}
& \raisebox{-5pt}{\includegraphics[width=2.1cm]{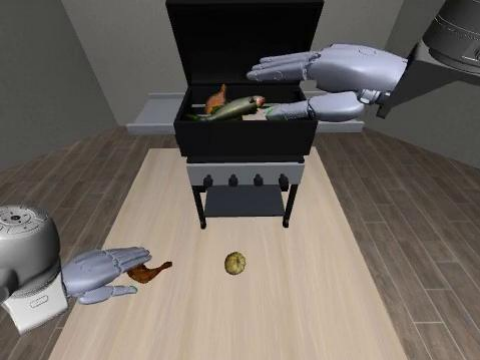}} &
\textbf{2.~\texttt{grill\_\allowbreak drumsticks\_\allowbreak fish\_\allowbreak and\_\allowbreak squash}}\hfill {\footnotesize $K{=}7$ subtasks $\cdot$ 100 demos}\newline
{\small\emph{Instruction:} Open the grill lid, place the two drumsticks, the two fish, and the squash on the grill, then close the lid.}\newline
{\footnotesize\textbf{Scoring:} Left arm opens the grill lid \textbf{(15)} $\to$ Left arm lays the first drumstick on the grill \textbf{(15)}$^{\dagger}$ $\to$ Left arm lays the second drumstick on the grill \textbf{(15)}$^{\dagger}$ $\to$ Right arm lays the first fish on the grill \textbf{(15)}$^{\dagger}$ $\to$ Right arm lays the second fish on the grill \textbf{(15)}$^{\dagger}$ $\to$ Right arm lays the squash on the grill \textbf{(15)}$^{\dagger}$ $\to$ Left arm closes the grill lid \textbf{(10)}$^{\dagger}$.}
\end{tabular}}

\par\vspace{3pt}\noindent
{\renewcommand{\arraystretch}{1.0}%
\begin{tabular}{@{}>{\columncolor{catlh}}p{2.2pt}@{\hspace{5pt}}m{2.1cm}@{\hspace{6pt}}p{\dimexpr\linewidth-2.2pt-5pt-2.1cm-6pt\relax}@{}}
& \raisebox{-5pt}{\includegraphics[width=2.1cm]{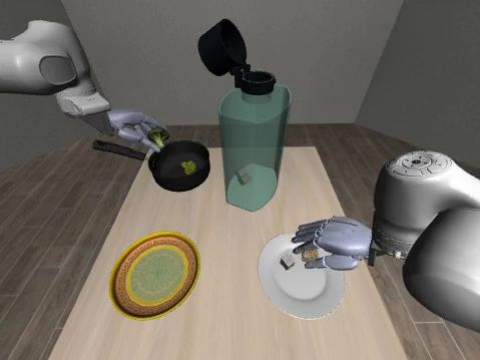}} &
\textbf{3.~\texttt{fill\_\allowbreak pot\_\allowbreak and\_\allowbreak bottle\_\allowbreak with\_\allowbreak sugar\_\allowbreak cubes\_\allowbreak and\_\allowbreak veg}}\hfill {\footnotesize $K{=}9$ subtasks $\cdot$ 100 demos}\newline
{\small\emph{Instruction:} Left arm opens the bottle's cap, drops the two vegetables and the largest sugar cube into the pot, closes the cap, and lifts it, while the right arm drops the other three sugar cubes into the bottle.}\newline
{\footnotesize\textbf{Scoring:} Left arm opens the bottle's articulated cap \textbf{(20)} $\to$ Left arm places the first vegetable into the pot \textbf{(10)}$^{\dagger}$ $\to$ Left arm places the second vegetable into the pot \textbf{(10)}$^{\dagger}$ $\to$ Left arm drops the largest sugar cube into the pot \textbf{(10)}$^{\dagger}$ $\to$ Right arm drops the second sugar cube into the bottle \textbf{(10)}$^{\dagger}$ $\to$ Right arm drops the third sugar cube into the bottle \textbf{(10)}$^{\dagger}$ $\to$ Right arm drops the fourth sugar cube into the bottle \textbf{(10)}$^{\dagger}$ $\to$ Left arm lifts the pot \textbf{(10)}$^{\dagger}$ $\to$ Left arm closes the bottle's articulated cap \textbf{(10)}$^{\dagger}$.}
\end{tabular}}

\par\vspace{3pt}\noindent
{\renewcommand{\arraystretch}{1.0}%
\begin{tabular}{@{}>{\columncolor{catlh}}p{2.2pt}@{\hspace{5pt}}m{2.1cm}@{\hspace{6pt}}p{\dimexpr\linewidth-2.2pt-5pt-2.1cm-6pt\relax}@{}}
& \raisebox{-5pt}{\includegraphics[width=2.1cm]{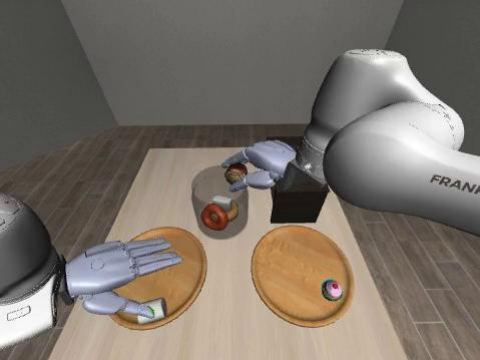}} &
\textbf{4.~\texttt{fill\_\allowbreak jar\_\allowbreak with\_\allowbreak snacks}}\hfill {\footnotesize $K{=}8$ subtasks $\cdot$ 100 demos}\newline
{\small\emph{Instruction:} Take the jar off the shelf and set it on the table, put all the snacks into the jar, and place the jar back on the shelf.}\newline
{\footnotesize\textbf{Scoring:} Take the jar off the shelf \textbf{(20)} $\to$ Place the first donut into the jar \textbf{(10)}$^{\dagger}$ $\to$ Place the second donut into the jar \textbf{(10)}$^{\dagger}$ $\to$ Drops the first marshmallow into the jar \textbf{(10)}$^{\dagger}$ $\to$ Place the first cupcake into the jar \textbf{(10)}$^{\dagger}$ $\to$ Place the second cupcake into the jar \textbf{(10)}$^{\dagger}$ $\to$ Drop the second marshmallow into the jar \textbf{(10)}$^{\dagger}$ $\to$ Place the jar back on the shelf \textbf{(20)}$^{\dagger}$.}
\end{tabular}}

\par\vspace{3pt}\noindent
{\renewcommand{\arraystretch}{1.0}%
\begin{tabular}{@{}>{\columncolor{catlh}}p{2.2pt}@{\hspace{5pt}}m{2.1cm}@{\hspace{6pt}}p{\dimexpr\linewidth-2.2pt-5pt-2.1cm-6pt\relax}@{}}
& \raisebox{-5pt}{\includegraphics[width=2.1cm]{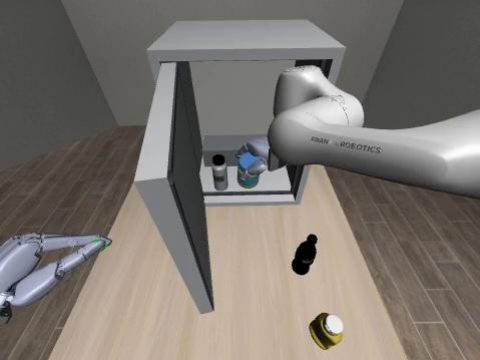}} &
\textbf{5.~\texttt{restock\_\allowbreak fridge\_\allowbreak with\_\allowbreak condiments\_\allowbreak and\_\allowbreak bottles}}\hfill {\footnotesize $K{=}7$ subtasks $\cdot$ 100 demos}\newline
{\small\emph{Instruction:} Open the fridge, put the two condiments and the two tall bottles inside, and close the door with the water bottle standing upright.}\newline
{\footnotesize\textbf{Scoring:} Left arm opens the fridge door \textbf{(25)} $\to$ Right arm puts the water bottle into the fridge \textbf{(10)}$^{\dagger}$ $\to$ Right arm puts the bottled drink into the fridge \textbf{(10)}$^{\dagger}$ $\to$ Left arm puts the mayonnaise into the fridge \textbf{(10)}$^{\dagger}$ $\to$ Left arm puts the mustard into the fridge \textbf{(10)}$^{\dagger}$ $\to$ Left arm closes the fridge door \textbf{(15)}$^{\dagger}$ $\to$ The water bottle stands upright in the fridge \textbf{(20)}$^{\dagger}$.}
\end{tabular}}

\par\vspace{3pt}\noindent
{\renewcommand{\arraystretch}{1.0}%
\begin{tabular}{@{}>{\columncolor{catlh}}p{2.2pt}@{\hspace{5pt}}m{2.1cm}@{\hspace{6pt}}p{\dimexpr\linewidth-2.2pt-5pt-2.1cm-6pt\relax}@{}}
& \raisebox{-5pt}{\includegraphics[width=2.1cm]{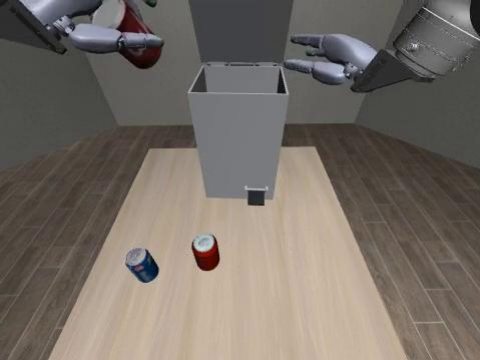}} &
\textbf{6.~\texttt{recycle\_\allowbreak five\_\allowbreak cans\_\allowbreak into\_\allowbreak bin}}\hfill {\footnotesize $K{=}7$ subtasks $\cdot$ 100 demos}\newline
{\small\emph{Instruction:} Open the pedal bin lid, toss the five cans in, then close the lid.}\newline
{\footnotesize\textbf{Scoring:} Left arm flips open the bin lid \textbf{(15)} $\to$ Left arm tosses the first can into the bin \textbf{(15)}$^{\dagger}$ $\to$ Left arm tosses the second can into the bin \textbf{(15)}$^{\dagger}$ $\to$ Left arm tosses the third can into the bin \textbf{(15)}$^{\dagger}$ $\to$ Right arm tosses the fourth can into the bin \textbf{(15)}$^{\dagger}$ $\to$ Right arm tosses the fifth can into the bin \textbf{(15)}$^{\dagger}$ $\to$ Left arm closes the bin lid \textbf{(10)}$^{\dagger}$.}
\end{tabular}}

\par\vspace{3pt}\noindent
{\renewcommand{\arraystretch}{1.0}%
\begin{tabular}{@{}>{\columncolor{catlh}}p{2.2pt}@{\hspace{5pt}}m{2.1cm}@{\hspace{6pt}}p{\dimexpr\linewidth-2.2pt-5pt-2.1cm-6pt\relax}@{}}
& \raisebox{-5pt}{\includegraphics[width=2.1cm]{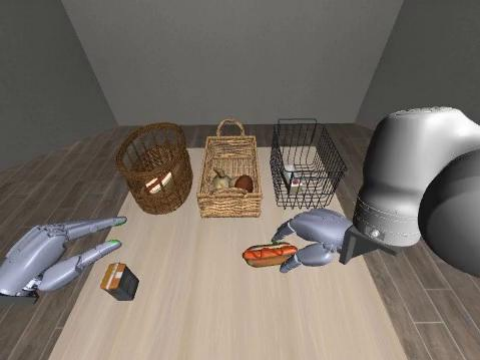}} &
\textbf{7.~\texttt{sort\_\allowbreak mixed\_\allowbreak food\_\allowbreak into\_\allowbreak boxes}}\hfill {\footnotesize $K{=}6$ subtasks $\cdot$ 100 demos}\newline
{\small\emph{Instruction:} Put the mushrooms into the middle vegetable basket, the hot dogs into the front meat basket, and the drink cartons into the back drinks basket.}\newline
{\footnotesize\textbf{Scoring:} Drop the first mushroom into the middle vegetable basket \textbf{(10)}$^{\dagger}$ $\cdot$ Drop the second mushroom into the middle vegetable basket \textbf{(10)}$^{\dagger}$ $\cdot$ Drop the first hot dog into the front meat basket \textbf{(20)}$^{\dagger}$ $\cdot$ Drop the second hot dog into the front meat basket \textbf{(20)}$^{\dagger}$ $\cdot$ Drop the first drink carton into the back drinks basket \textbf{(20)}$^{\dagger}$ $\cdot$ Drop the second drink carton into the back drinks basket \textbf{(20)}$^{\dagger}$.}
\end{tabular}}

\par\vspace{3pt}\noindent
{\renewcommand{\arraystretch}{1.0}%
\begin{tabular}{@{}>{\columncolor{catlh}}p{2.2pt}@{\hspace{5pt}}m{2.1cm}@{\hspace{6pt}}p{\dimexpr\linewidth-2.2pt-5pt-2.1cm-6pt\relax}@{}}
& \raisebox{-5pt}{\includegraphics[width=2.1cm]{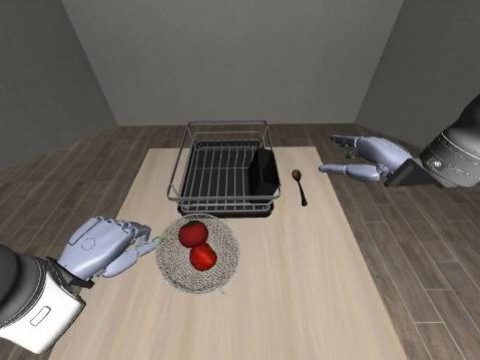}} &
\textbf{8.~\texttt{place\_\allowbreak strainer\_\allowbreak with\_\allowbreak tomatoes\_\allowbreak and\_\allowbreak spoon\_\allowbreak on\_\allowbreak dish\_\allowbreak rack}}\hfill {\footnotesize $K{=}8$ subtasks $\cdot$ 100 demos}\newline
{\small\emph{Instruction:} Pick up the strainer, put the two tomatoes into it, lift it onto the dish rack, then place the spoon onto the dish rack as well.}\newline
{\footnotesize\textbf{Scoring:} Pick up the strainer \textbf{(10)} $\cdot$ Pick up the first tomato \textbf{(10)} $\cdot$ Place the first tomato into the strainer \textbf{(10)}$^{\dagger}$ $\cdot$ Pick up the second tomato \textbf{(10)} $\cdot$ Place the second tomato into the strainer \textbf{(10)}$^{\dagger}$ $\cdot$ Place the strainer onto the dish rack \textbf{(15)}$^{\dagger}$ $\cdot$ Pick up the spoon \textbf{(15)} $\cdot$ Place the spoon onto the dish rack \textbf{(20)}$^{\dagger}$.}
\end{tabular}}

\par\vspace{3pt}\noindent
{\renewcommand{\arraystretch}{1.0}%
\begin{tabular}{@{}>{\columncolor{catlh}}p{2.2pt}@{\hspace{5pt}}m{2.1cm}@{\hspace{6pt}}p{\dimexpr\linewidth-2.2pt-5pt-2.1cm-6pt\relax}@{}}
& \raisebox{-5pt}{\includegraphics[width=2.1cm]{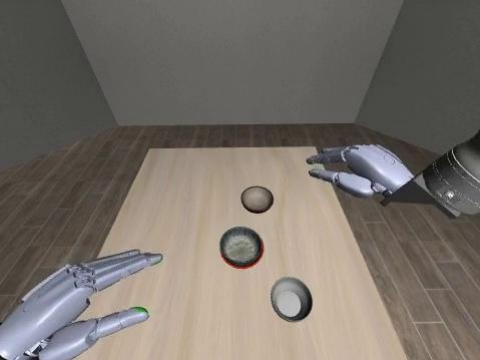}} &
\textbf{9.~\texttt{nest\_\allowbreak four\_\allowbreak bowls}}\hfill {\footnotesize $K{=}3$ subtasks $\cdot$ 100 demos}\newline
{\small\emph{Instruction:} Nest these four bowls of decreasing sizes into a stack, fitting each smaller bowl inside the larger one beneath it.}\newline
{\footnotesize\textbf{Scoring:} Nest the 12 cm bowl into the 13 cm base bowl \textbf{(20)}$^{\dagger}$ $\cdot$ Nest the 11.5 cm bowl into the 12 cm bowl \textbf{(30)}$^{\dagger}$ $\cdot$ Nest the 10 cm bowl into the 11.5 cm bowl \textbf{(50)}$^{\dagger}$.}
\end{tabular}}

\par\vspace{3pt}\noindent
{\renewcommand{\arraystretch}{1.0}%
\begin{tabular}{@{}>{\columncolor{catlh}}p{2.2pt}@{\hspace{5pt}}m{2.1cm}@{\hspace{6pt}}p{\dimexpr\linewidth-2.2pt-5pt-2.1cm-6pt\relax}@{}}
& \raisebox{-5pt}{\includegraphics[width=2.1cm]{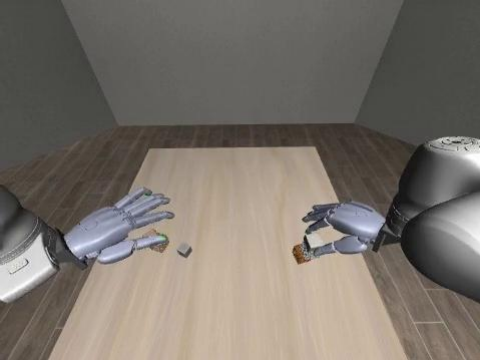}} &
\textbf{10.~\texttt{stack\_\allowbreak sugar\_\allowbreak cubes}}\hfill {\footnotesize $K{=}4$ subtasks $\cdot$ 100 demos}\newline
{\small\emph{Instruction:} With the left arm stack three sugar cubes on top of each other, and with the right arm stack three sugar cubes on top of each other.}\newline
{\footnotesize\textbf{Scoring:} Left arm stacks the middle cube on the bottom cube \textbf{(25)}$^{\dagger}$ $\cdot$ Left arm stacks the top cube on the middle cube \textbf{(25)}$^{\dagger}$ $\cdot$ Right arm stacks the middle cube on the bottom cube \textbf{(25)}$^{\dagger}$ $\cdot$ Right arm stacks the top cube on the middle cube \textbf{(25)}$^{\dagger}$.}
\end{tabular}}

\subsection{Memory (10 specifications, 10 with released demonstrations)}
\label{app:tasks:memory}

Latent-state tracking under perceptual aliasing: in \emph{color-swap cups} a cube's color is visually removed after a button sequence and must be recalled to choose the matching cup, in \emph{picnic-basket FIFO} removal order must match insertion order, and in \emph{memory cups} a reveal phase shows target contents that are later hidden. These tasks are impossible from a single frame and probe genuine temporal state tracking rather than reactive control.

\par\vspace{3pt}\noindent
{\renewcommand{\arraystretch}{1.0}%
\begin{tabular}{@{}>{\columncolor{catmem}}p{2.2pt}@{\hspace{5pt}}m{2.1cm}@{\hspace{6pt}}p{\dimexpr\linewidth-2.2pt-5pt-2.1cm-6pt\relax}@{}}
& \raisebox{-5pt}{\includegraphics[width=2.1cm]{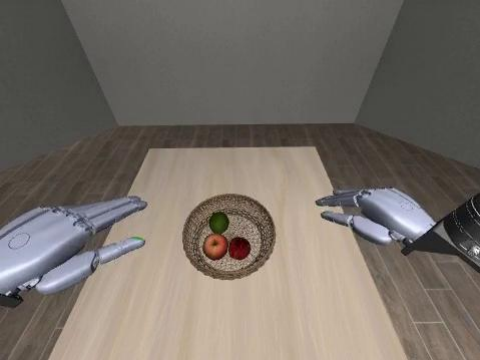}} &
\textbf{1.~\texttt{drop\_\allowbreak fruits\_\allowbreak in\_\allowbreak basket\_\allowbreak then\_\allowbreak remove\_\allowbreak in\_\allowbreak order}}\hfill {\footnotesize $K{=}2$ subtasks $\cdot$ 100 demos}\newline
{\small\emph{Instruction:} With the left arm drop the three fruits into the basket in any order, then with the right arm take them back out in the same order they went in.}\newline
{\footnotesize\textbf{Scoring:} Left arm places all three objects into the basket \textbf{(50)} $\to$ Right arm removes all three in the order they were placed \textbf{(50)}$^{\dagger}$.}
\end{tabular}}

\par\vspace{3pt}\noindent
{\renewcommand{\arraystretch}{1.0}%
\begin{tabular}{@{}>{\columncolor{catmem}}p{2.2pt}@{\hspace{5pt}}m{2.1cm}@{\hspace{6pt}}p{\dimexpr\linewidth-2.2pt-5pt-2.1cm-6pt\relax}@{}}
& \raisebox{-5pt}{\includegraphics[width=2.1cm]{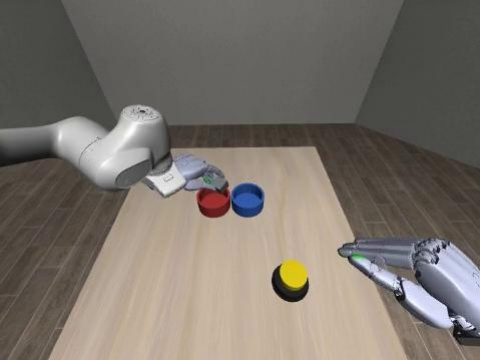}} &
\textbf{2.~\texttt{swap\_\allowbreak cup\_\allowbreak colors\_\allowbreak with\_\allowbreak cube}}\hfill {\footnotesize $K{=}4$ subtasks $\cdot$ 100 demos}\newline
{\small\emph{Instruction:} With the right arm press the yellow button three times, pick up the cube and swing it over the two cups five times to swap their colors, then place it in the matching-color cup.}\newline
{\footnotesize\textbf{Scoring:} Right arm presses the yellow button exactly three times, the cube's color disappears \textbf{(25)}$^{\dagger}$ $\to$ Pick up the cube \textbf{(5)} $\to$ Pass over the cups so the two cups swap colors five times \textbf{(25)} $\to$ Place the cube into the cup whose color matches it \textbf{(45)}$^{\dagger}$.}
\end{tabular}}

\par\vspace{3pt}\noindent
{\renewcommand{\arraystretch}{1.0}%
\begin{tabular}{@{}>{\columncolor{catmem}}p{2.2pt}@{\hspace{5pt}}m{2.1cm}@{\hspace{6pt}}p{\dimexpr\linewidth-2.2pt-5pt-2.1cm-6pt\relax}@{}}
& \raisebox{-5pt}{\includegraphics[width=2.1cm]{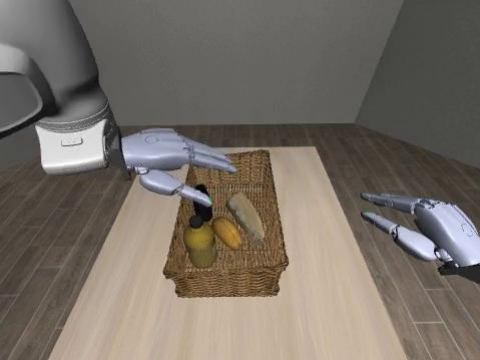}} &
\textbf{3.~\texttt{drop\_\allowbreak items\_\allowbreak into\_\allowbreak picnic\_\allowbreak basket\_\allowbreak then\_\allowbreak remove\_\allowbreak in\_\allowbreak order}}\hfill {\footnotesize $K{=}2$ subtasks $\cdot$ 100 demos}\newline
{\small\emph{Instruction:} Drop the hotdog bun, baguette, juice carton and water bottle into the picnic basket in any order, then take them out in the same order.}\newline
{\footnotesize\textbf{Scoring:} Both arms place all four items into the basket \textbf{(50)} $\to$ Both arms remove all four items in the same order they were placed \textbf{(50)}$^{\dagger}$.}
\end{tabular}}

\par\vspace{3pt}\noindent
{\renewcommand{\arraystretch}{1.0}%
\begin{tabular}{@{}>{\columncolor{catmem}}p{2.2pt}@{\hspace{5pt}}m{2.1cm}@{\hspace{6pt}}p{\dimexpr\linewidth-2.2pt-5pt-2.1cm-6pt\relax}@{}}
& \raisebox{-5pt}{\includegraphics[width=2.1cm]{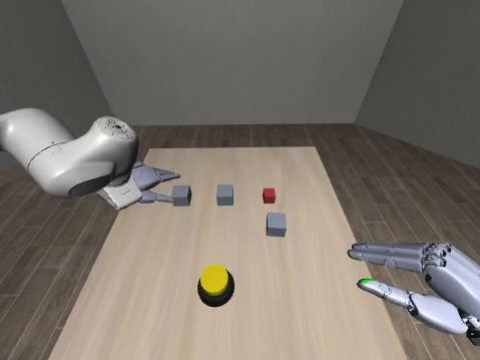}} &
\textbf{4.~\texttt{uncover\_\allowbreak cubes\_\allowbreak in\_\allowbreak order}}\hfill {\footnotesize $K{=}7$ subtasks $\cdot$ 100 demos}\newline
{\small\emph{Instruction:} Press the yellow button three times to cover the three cubes with caps, then lift the caps to uncover the red, green, and blue cubes in that order.}\newline
{\footnotesize\textbf{Scoring:} Right arm presses the yellow button, the caps appear \textbf{(10)}$^{\dagger}$ $\to$ All three cubes are covered by the caps \textbf{(0)} $\to$ Lift the cap covering the red cube first \textbf{(30)} $\to$ Then lift the cap covering the green cube \textbf{(30)} $\to$ Finally lift the cap covering the blue cube, all revealed \textbf{(30)}$^{\dagger}$ $\to$ Red cube still uncovered at the end \textbf{(0)}$^{\dagger}$ $\to$ Green cube still uncovered at the end \textbf{(0)}$^{\dagger}$.}
\end{tabular}}

\par\vspace{3pt}\noindent
{\renewcommand{\arraystretch}{1.0}%
\begin{tabular}{@{}>{\columncolor{catmem}}p{2.2pt}@{\hspace{5pt}}m{2.1cm}@{\hspace{6pt}}p{\dimexpr\linewidth-2.2pt-5pt-2.1cm-6pt\relax}@{}}
& \raisebox{-5pt}{\includegraphics[width=2.1cm]{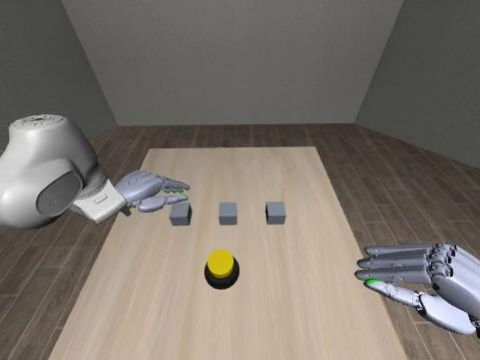}} &
\textbf{5.~\texttt{uncover\_\allowbreak red\_\allowbreak cube}}\hfill {\footnotesize $K{=}5$ subtasks $\cdot$ 100 demos}\newline
{\small\emph{Instruction:} Press the yellow button three times to lift only the cover concealing the red cube.}\newline
{\footnotesize\textbf{Scoring:} Right arm presses the yellow button exactly three times, the caps appear \textbf{(20)}$^{\dagger}$ $\to$ All three cubes are covered by the caps \textbf{(0)} $\to$ Lift only the cap covering the red cube \textbf{(80)}$^{\dagger}$ $\to$ Green cube still covered at the end \textbf{(0)}$^{\dagger}$ $\to$ Blue cube still covered at the end \textbf{(0)}$^{\dagger}$.}
\end{tabular}}

\par\vspace{3pt}\noindent
{\renewcommand{\arraystretch}{1.0}%
\begin{tabular}{@{}>{\columncolor{catmem}}p{2.2pt}@{\hspace{5pt}}m{2.1cm}@{\hspace{6pt}}p{\dimexpr\linewidth-2.2pt-5pt-2.1cm-6pt\relax}@{}}
& \raisebox{-5pt}{\includegraphics[width=2.1cm]{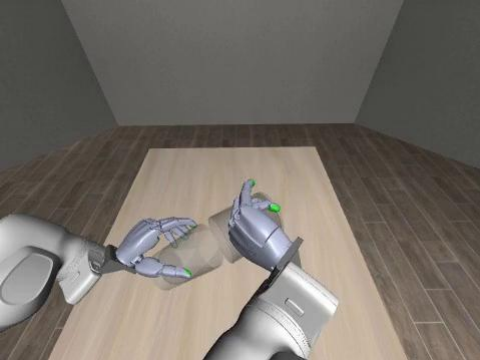}} &
\textbf{6.~\texttt{pour\_\allowbreak ice\_\allowbreak between\_\allowbreak two\_\allowbreak cups}}\hfill {\footnotesize $K{=}2$ subtasks $\cdot$ 100 demos}\newline
{\small\emph{Instruction:} Right arm pours the ice from the right cup into the left cup, then left arm pours it back into the right cup and sets both cups down.}\newline
{\footnotesize\textbf{Scoring:} Right arm pours the ice from the right cup into the left cup \textbf{(50)} $\to$ Left arm pours the ice back into the right cup and sets both cups down \textbf{(50)}$^{\dagger}$.}
\end{tabular}}

\par\vspace{3pt}\noindent
{\renewcommand{\arraystretch}{1.0}%
\begin{tabular}{@{}>{\columncolor{catmem}}p{2.2pt}@{\hspace{5pt}}m{2.1cm}@{\hspace{6pt}}p{\dimexpr\linewidth-2.2pt-5pt-2.1cm-6pt\relax}@{}}
& \raisebox{-5pt}{\includegraphics[width=2.1cm]{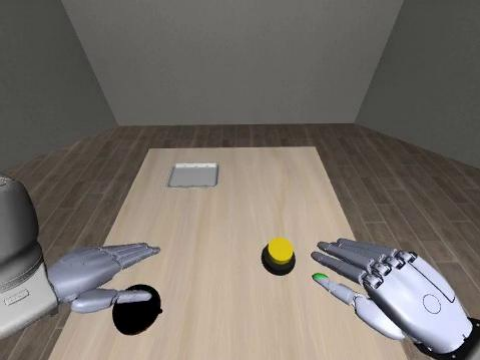}} &
\textbf{7.~\texttt{press\_\allowbreak button\_\allowbreak then\_\allowbreak grasp\_\allowbreak mug}}\hfill {\footnotesize $K{=}3$ subtasks $\cdot$ 100 demos}\newline
{\small\emph{Instruction:} With the right arm press the yellow button exactly four times, and only then, with the left arm pick up the mug and place it on the tray.}\newline
{\footnotesize\textbf{Scoring:} Right arm presses the yellow button exactly four times \textbf{(60)}$^{\dagger}$ $\to$ Left arm picks up the mug \textbf{(10)} $\to$ Left arm places the mug on the tray \textbf{(30)}$^{\dagger}$.}
\end{tabular}}

\par\vspace{3pt}\noindent
{\renewcommand{\arraystretch}{1.0}%
\begin{tabular}{@{}>{\columncolor{catmem}}p{2.2pt}@{\hspace{5pt}}m{2.1cm}@{\hspace{6pt}}p{\dimexpr\linewidth-2.2pt-5pt-2.1cm-6pt\relax}@{}}
& \raisebox{-5pt}{\includegraphics[width=2.1cm]{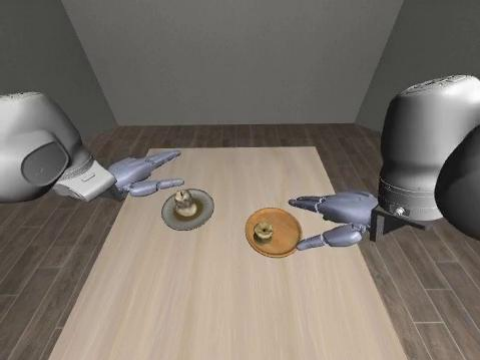}} &
\textbf{8.~\texttt{swap\_\allowbreak candles\_\allowbreak between\_\allowbreak bowls}}\hfill {\footnotesize $K{=}5$ subtasks $\cdot$ 100 demos}\newline
{\small\emph{Instruction:} Swap and restore the two candles.}\newline
{\footnotesize\textbf{Scoring:} Left arm moves the left candle to the right bowl \textbf{(20)} $\to$ Right arm moves the right candle to the left bowl \textbf{(20)} $\to$ Left arm moves the right candle back to the right bowl \textbf{(30)} $\to$ Right arm moves the left candle back to the left bowl (restored) \textbf{(30)}$^{\dagger}$ $\to$ Right candle restored to the right bowl \textbf{(0)}$^{\dagger}$.}
\end{tabular}}

\par\vspace{3pt}\noindent
{\renewcommand{\arraystretch}{1.0}%
\begin{tabular}{@{}>{\columncolor{catmem}}p{2.2pt}@{\hspace{5pt}}m{2.1cm}@{\hspace{6pt}}p{\dimexpr\linewidth-2.2pt-5pt-2.1cm-6pt\relax}@{}}
& \raisebox{-5pt}{\includegraphics[width=2.1cm]{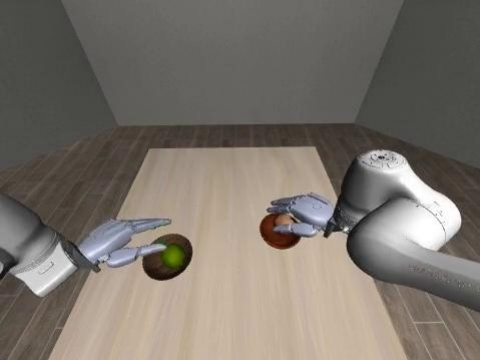}} &
\textbf{9.~\texttt{swap\_\allowbreak fruits\_\allowbreak between\_\allowbreak bowls}}\hfill {\footnotesize $K{=}4$ subtasks $\cdot$ 100 demos}\newline
{\small\emph{Instruction:} Left arm moves the peach to the right bowl, right arm moves the lime to the left, then left arm moves the lime to the right and right arm moves the peach to the left.}\newline
{\footnotesize\textbf{Scoring:} Left arm moves the peach to the right bowl \textbf{(25)} $\to$ Right arm moves the lime to the left bowl \textbf{(25)} $\to$ Left arm moves the lime back to the right bowl \textbf{(25)} $\to$ Right arm moves the peach back to the left bowl \textbf{(25)}$^{\dagger}$.}
\end{tabular}}

\par\vspace{3pt}\noindent
{\renewcommand{\arraystretch}{1.0}%
\begin{tabular}{@{}>{\columncolor{catmem}}p{2.2pt}@{\hspace{5pt}}m{2.1cm}@{\hspace{6pt}}p{\dimexpr\linewidth-2.2pt-5pt-2.1cm-6pt\relax}@{}}
& \raisebox{-5pt}{\includegraphics[width=2.1cm]{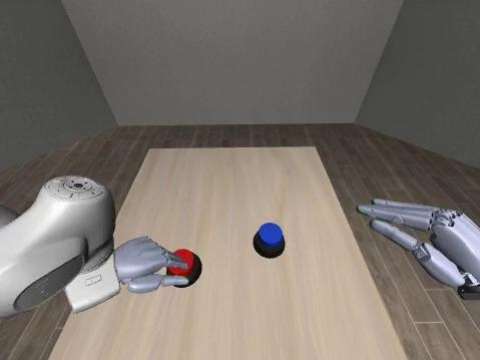}} &
\textbf{10.~\texttt{press\_\allowbreak red\_\allowbreak button\_\allowbreak then\_\allowbreak blue\_\allowbreak button}}\hfill {\footnotesize $K{=}2$ subtasks $\cdot$ 100 demos}\newline
{\small\emph{Instruction:} With the left arm press the red button exactly ten times, and only then, with the right arm press the blue button once.}\newline
{\footnotesize\textbf{Scoring:} Left arm presses the red button exactly ten times \textbf{(50)}$^{\dagger}$ $\cdot$ Right arm presses the blue button once \textbf{(50)}$^{\dagger}$.}
\end{tabular}}

\section{Discussion and Limitations}

The five-axis organization reflects a deliberate benchmark design. As VLA
training data grows, average-case skill becomes less discriminative, while
memory, horizon, instruction fidelity, data efficiency, and robustness remain
capability edges. Progress-based predicate scoring makes the heterogeneous
tasks commensurable, exposes partial progress when strict success is not
reached, and keeps evaluation fully automated. The resulting per-axis vectors
show whether short-horizon competence extends to memory and long-horizon
tasks. ID/OOD gaps measure aggregate robustness under the combined scripted
shifts, and Limited-Demonstrations isolates acquisition from 30 examples.

BiDex-100 is a rigid-body benchmark. MuJoCo does not model fluids, deformable
objects, or rich frictional contact at the fidelity of hardware, and
sim-to-real transfer of policies trained on the released demonstrations
remains to be validated on the physical platforms. The object library, while
diverse, is centered on tabletop food and kitchen scenes. The OOD shifts are
scripted and finite; learned or adversarially generated shift distributions
are a natural extension.

Within this scope, evaluating seven contemporary VLA models under one protocol
establishes reference baselines and a diagnostic framework of capability
profiles, ID/OOD robustness gaps, and low-data regimes. We release the code,
tasks, data, and evaluation tooling to support reproducible community use.

\end{document}